\documentclass[10pt,twocolumn,letterpaper]{article}

\usepackage[pagenumbers]{cvpr} %

\usepackage{microtype}

\usepackage{titletoc}

\usepackage[resetlabels]{multibib}
\newcites{supp}{Supplementary references}
\newcommand{\scite}[1]{{\renewcommand{\citenumfont}[1]{S##1}\citesupp{#1}}}

\renewcommand{\paragraph}[1]{\vspace{.4em}\noindent\textbf{#1.}}

\usepackage{xspace}

\usepackage{multirow}
\usepackage{colortbl}

\definecolor{ablrow}{gray}{0.9}
\definecolor{ablup}{RGB}{0,120,0}
\definecolor{abldown}{RGB}{190,30,30}
\newcommand{\dn}[1]{{\color{abldown}$-$#1}} %
\newcommand{\up}[1]{{\color{ablup}$+$#1}}   %
\newcommand{\ns}[1]{{\color{gray}#1}}       %

\definecolor{methodgreen}{RGB}{0,120,0}
\newcommand{\method}{TwinD\xspace}
\newcommand{\PEcoreL}{PE\textsubscript{core}L\xspace}
\newcommand{\C}[1]{\unskip\ignorespaces} %
\newcommand{\q}[1]{``#1''} %

\definecolor{cvprblue}{rgb}{0.21,0.49,0.74}
\usepackage[pagebackref,breaklinks,colorlinks,allcolors=cvprblue]{hyperref}

\def\confName{CVPR}
\def\confYear{2027}

\title{TwinD: Generalizing Deepfake Detection from Blending to Entire Face Synthesis}

\author{
Andrii Yermakov\textsuperscript{1} \quad
Jan Cech\textsuperscript{1} \quad
Mario Fritz\textsuperscript{2} \quad
Jiri Matas\textsuperscript{1} \\
\textsuperscript{1}Czech Technical University in Prague \quad
\textsuperscript{2}CISPA Helmholtz Center for Information Security \\
{\tt\small \{yermaand,cechj,matas\}@fel.cvut.cz} \quad
{\tt\small fritz@cispa.de}
}

\begin{document}
\maketitle
\begin{abstract}

Facial deepfake detectors are usually trained on blending fakes, where a generated face is pasted back into a real frame. Recent generators synthesize the entire face and leave no traces of blending. On a benchmark of 19 datasets released from 2019 to 2026, the best prior facial deepfake detector reaches a mean AUROC of 96.2\% on ten blending datasets, but none exceeds 78.1\% on five datasets of entire face synthesis (EFS). AI-generated content (AIGC) detectors do not close this gap. We propose \method, a detector trained on FaceForensics++ fakes and two kinds of twins: fakes made from a real frame. A reconstruction twin is a real frame reconstructed by an autoencoder, with the real pixels put back in a random region. It targets the neural traces of a decoder. A blob twin blends an altered copy of a real frame with the frame itself, targeting blending artifacts. It needs no generator and no face-shaped mask. Together, these twins provide two complementary forensic cues that generalize to unseen manipulations. \method reaches a mean AUROC of 96.3\%, 8.9 points above the prior state-of-the-art, with 98.1\% on the blending datasets and 95.6\% on the EFS datasets. Trained on twins alone, without a deepfake dataset, single models still reach 94.3\% on average over three seeds. For reproducibility, the code, trained model, and training data are at: \url{https://github.com/yermandy/TwinD}

\end{abstract}

\section{Introduction}
\label{sec:intro}

\begin{figure*}[t]
\centering
\includegraphics[width=\textwidth]{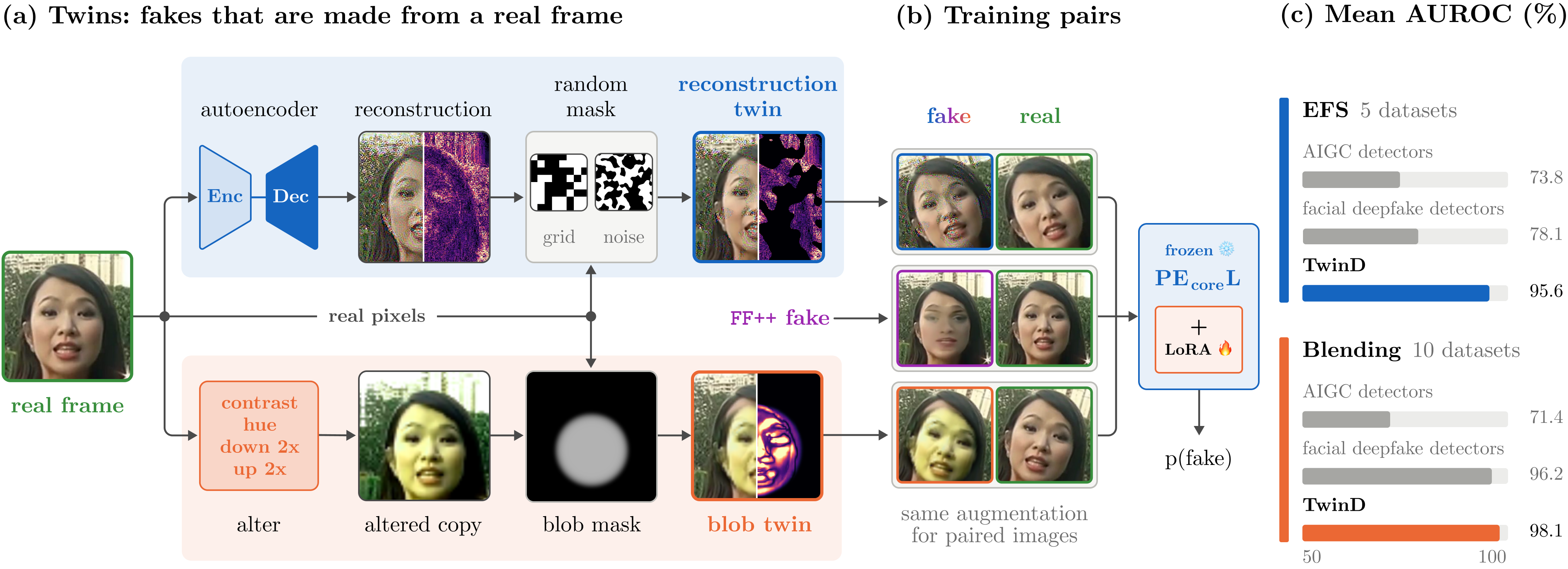}
\caption{\label{fig:overview}
    \textbf{Overview of \method.} (a) A \emph{twin} is a fake made from a real frame. A \emph{reconstruction twin} is an autoencoder reconstruction with the real pixels put back inside a random grid or noise mask. A \emph{blob twin} blends an altered copy of the frame with the frame itself through a blob mask. Right halves of the twins show the difference to the real frame. The traces of the autoencoder are amplified $10\times$ for visualization. (b) Each fake is paired with the real frame it was made from. Both get the same augmentation, so they differ only by the manipulation. \method is a frozen \PEcoreL with LoRA, trained on batches of such pairs. (c) Mean video-level AUROC (\%) on the \texttt{EFS} and \texttt{Blending} datasets of \cref{tab:sota}. Gray bars show the best prior detector of each kind.
}
\end{figure*}

Most facial deepfake datasets contain blending fakes, where a swapped or reenacted face is blended back into the real frame~\cite{FF,DFD,CDFv2,FFIW,FakeAVCeleb,DFDM,PolyGlotFake,IDForge,HiDF}. Recent generators synthesize the entire video frame without blending, and datasets of such entire face synthesis (EFS) fakes have appeared only recently~\cite{TalkingHeadBench,OmniFake,DigiFakeAV,MintVid,DF26}. A facial deepfake detector has to catch both kinds. The fundamental challenge is to generalize to unseen generators.

Facial deepfake detectors~\cite{SBI,Effort,ForAda,VLAForge,FS-VFM,DFD-HR,GenD} are usually trained on the FaceForensics++ dataset (FF++)~\cite{FF}, which contains only blending fakes. They can learn the traces of blending and generalize well to other blending datasets~\cite{FaceXray,SBI}. EFS leaves no such traces. In \cref{tab:sota}, the best of them reaches a mean AUROC of 96.2\% on ten blending datasets, but none exceed 78.1\% on five EFS datasets.

Detectors of AI-generated content (AIGC) come from a separate line of research. They target fully synthetic images and promise to generalize to many unseen generators~\cite{UniFD,DGS-Net,LTD,DDA,B-Free}. Some of them are trained on real images and their generative reconstructions~\cite{DRCT,DDA,B-Free,AlignedForensics}, which differ from the real images only by the neural traces of the generator. So, they could be expected to handle EFS deepfakes. They do not. The best AIGC detector reaches a mean AUROC of 73.8\% on the EFS datasets, below the best facial deepfake detector. On the blending datasets, none exceeds 71.4\%.

No evaluated detector works well on both blending and EFS fakes. We close this gap with a better training set. We introduce two kinds of \emph{twins}, \q{pseudo-fakes} that are made from a real frame. Reconstruction twins improve generalization on EFS fakes and blob twins on blending fakes, see \cref{fig:overview}.

A \emph{reconstruction twin} is a real face frame reconstructed by an autoencoder, as in AIGC detectors, but with the real pixels put back inside a random mask. So, the detector learns decoder traces even when they cover only part of the frame.

A \emph{blob twin} blends an altered copy of a real frame with the frame itself through a smooth round mask, so it needs no generator.
It simplifies the popular self-blended images (SBI)~\cite{SBI}: the mask is round instead of aligned with the facial boundaries, so it needs no landmarks.

We propose \method, a detector trained on FF++ fakes and two kinds of twins. Each fake is paired with the real frame from which it was made, and both receive the same training augmentation, so the only difference between them is the manipulation. The contributions of this paper are:

\begin{enumerate}
    \item The proposed \textbf{\method} is evaluated against 15 recent detectors on 19 facial deepfake datasets released from 2019 to 2026. To our knowledge, it is the largest evaluation of released detectors on the same face crops that covers both older blending fakes and recent EFS fakes.
    \method outperforms the previous state-of-the-art by a large margin~-- a mean AUROC of 96.3\%, 8.9 points above the best prior detector, with 98.1\% on the blending datasets and 95.6\% on the EFS datasets. It reaches at least 90.4\% on each of the 19 datasets.
    \item We show that the two kinds of twins are complementary: blob twins cover blending fakes, and reconstruction twins cover EFS fakes. Trained on twins alone, without seeing any deepfake dataset, single models still reach AUROC of 94.3\% averaged over three seeds, which is above all 15 prior detectors.
\end{enumerate}

\section{Related work}
\label{sec:related_work}

\paragraph{Pseudo-fakes} Pseudo-fake methods build training fakes from real images by using only non-generative manipulations. Face X-ray~\cite{FaceXray} pastes the face of another real person, matched by facial landmarks, into a frame. Self-blended images (SBI)~\cite{SBI} replace a person's face with an altered copy of the same face, blend it through a face-shaped mask built from facial landmarks, and randomize the mask and the geometry of the copy. The proposed blob twins (\cref{sec:blob-twins}) are simpler than both. Their mask is a round blob instead of a face shape, so it needs no landmarks, and a small shift is the only change in geometry. They outperform SBI on the blending datasets (\cref{tab:ablations}).

\paragraph{Training on reconstructions} Current state-of-the-art AIGC detectors are trained on real images and their reconstructions, so the two classes differ only by the traces of the generator. DRCT~\cite{DRCT} and B-Free~\cite{B-Free} reconstruct real images using a diffusion model. AlignedForensics~\cite{AlignedForensics} and DDA~\cite{DDA} use only the autoencoder of a latent diffusion model. AlignedForensics also puts each reconstruction in the same batch as its real image and gives both the same augmentation, which \method does as well (\cref{sec:pairs}). Two methods also mix real pixels back into the reconstruction. DDA averages the reconstruction and the real image with one random weight, so no region remains purely real. B-Free puts the real pixels back outside one object, marked by its segmentation mask or bounding box, and stores the result in its training set. All of them are trained on general images, and the methods we evaluate, AlignedForensics, DDA, and B-Free, do not transfer to facial deepfakes (\cref{tab:sota}). Reconstruction twins (\cref{sec:rec-twins}) build on these methods. They are reconstructions of real face frames, with the real pixels put back inside a random mask. Unlike in B-Free, the mask needs no segmentation labels and is redrawn at every use.

\paragraph{Parameter-efficient fine-tuning} Recent deepfake detectors adapt a large pre-trained backbone by fine-tuning a small part of it. GenD~\cite{GenD} adapts the image encoders of \PEcoreL~\cite{PE}, CLIP~\cite{CLIP} and DINOv3~\cite{DINOv3} by training only their LayerNorm parameters (LN-tuning~\cite{LN-tuning}). Effort~\cite{Effort} adapts CLIP by training the last singular components of each weight, like MiLoRA~\cite{MiLoRA}. DDA~\cite{DDA} trains LoRA~\cite{LoRA} on a DINOv2~\cite{DINOv2} backbone. \method trains LoRA on \PEcoreL (\cref{sec:model}).

\paragraph{Benchmarks} Facial deepfake detectors are commonly trained on FF++~\cite{FF} and tested on a few other datasets. DeepfakeBench~\cite{DeepfakeBench} unified the preprocessing and evaluation protocol. Its test datasets, like those used by most detectors, contain mostly blending fakes. Recent EFS fakes~\cite{TalkingHeadBench,OmniFake,DigiFakeAV,MintVid,DF26} are not yet part of this protocol. GenD~\cite{GenD} uses 14 datasets from 2019 to 2025. The benchmark of this paper (\cref{sec:benchmark}) includes 19 datasets released between 2019 and 2026 and groups them by how their fakes are made. For a fair comparison, it crops the faces in the same way for all detectors and keeps the resizing and normalization each detector was trained with.

\begin{table*}[t]
\centering
\caption{\label{tab:sota}
    Video-level AUROC (\%) of SOTA methods on 19 datasets sorted by their macro-average (Avg). Above the dataset names are their release years. \textsuperscript{\dag}A temporal model, which takes several frames as input in a single forward pass.
}
\tabcolsep=4pt
\resizebox{\textwidth}{!}{
\begin{tabular}{l|cccccccccc|cccc|ccccc|c}
\toprule
 & \multicolumn{10}{c|}{\textbf{\texttt{Blending}}} & \multicolumn{4}{c|}{\textbf{\texttt{Blending + EFS}}} & \multicolumn{5}{c|}{\textbf{\texttt{EFS}}} &  \\
 & 2019 & 2020 & 2020 & 2021 & 2021 & 2022 & 2024 & 2024 & 2024 & 2025 & 2025 & 2025 & 2025 & 2026 & 2026 & 2026 & 2026 & 2026 & 2026 &  \\
 & \texttt{DFD} & \texttt{FSh} & \texttt{CDFv2} & \texttt{FFIW} & \texttt{FAVC} & \texttt{DFDM} & \texttt{PGF} & \texttt{IDF} & \texttt{DSv1} & \texttt{HiDF} & \texttt{CDFv3} & \texttt{DSv2} & \texttt{RF} & \texttt{HF} & \texttt{THB} & \texttt{OF} & \texttt{DFAV} & \texttt{MV} & \texttt{DF26} & Avg \\
\midrule
\multicolumn{21}{l}{\textit{AIGC detectors}} \\[2pt]
Effort-SDv1.4~\cite{Effort} \C{Effort-sdv14-D26CVPR} & 51.5 & 58.9 & 48.8 & 56.5 & 55.4 & 55.3 & 38.4 & 63.7 & 31.6 & 55.0 & 63.5 & 51.5 & 79.4 & 75.8 & 79.6 & 77.3 & 48.9 & 38.6 & 62.3 & 57.5 \\
DGS-Net~\cite{DGS-Net} \C{DGS-Net-D26CVPR} & 58.4 & 71.9 & 55.4 & 73.6 & 45.1 & 72.8 & 49.4 & 38.1 & 41.5 & 47.8 & 71.9 & 41.7 & 56.8 & 66.2 & 66.0 & 49.0 & 68.5 & 69.8 & 75.0 & 58.9 \\
LTD-UFD~\cite{LTD} \C{LTD-UFD-D26CVPR} & 83.3 & 77.2 & 57.2 & 72.6 & 61.9 & 52.1 & 60.4 & 50.9 & 47.2 & 64.6 & 66.2 & 38.7 & 64.1 & 53.2 & 75.9 & 30.8 & 75.1 & 55.9 & 87.6 & 61.8 \\
AlignedFor.~\cite{AlignedForensics} \C{AlignedForensics-Sync-D26CVPR} & 55.0 & 62.9 & 61.0 & 78.3 & 93.2 & 65.5 & 56.8 & 93.7 & 43.0 & 47.8 & 28.6 & 74.8 & 68.8 & 90.3 & 93.5 & 56.9 & 83.6 & 40.0 & 67.8 & 66.4 \\
DDA~\cite{DDA} \C{DDA-D26CVPR} & 64.4 & 53.2 & 55.5 & 57.5 & 73.8 & 51.6 & 60.2 & 70.2 & 71.2 & 66.8 & 60.4 & 84.2 & 75.4 & 97.0 & 85.3 & 36.5 & 93.5 & 71.2 & 75.1 & 68.6 \\
B-Free~\cite{B-Free} \C{B-Free-D26CVPR} & 69.4 & 69.6 & 57.1 & 58.1 & 79.9 & 74.6 & 78.5 & 75.8 & 82.0 & 69.4 & 59.5 & 82.7 & 77.2 & \textbf{98.5} & 86.2 & 48.6 & 91.6 & 74.8 & 68.0 & 73.8 \\
\midrule
\multicolumn{21}{l}{\textit{Facial deepfake detectors}} \\[2pt]
SBI~\cite{SBI} \C{SBI-FFraw-D26CVPR} & 90.2 & 77.2 & 87.5 & 90.9 & 98.7 & 99.8 & 75.0 & 98.4 & 63.1 & 82.1 & 59.9 & 59.2 & 62.1 & 61.4 & 71.2 & 75.9 & 57.8 & 43.3 & 60.1 & 74.4 \\
FSFM~\cite{FSFM} \C{FSFM-D26CVPR} & 90.8 & 72.2 & 90.5 & 78.4 & 90.9 & 96.8 & 85.8 & 82.5 & 83.5 & 91.9 & 79.6 & 69.0 & 66.2 & 72.1 & 47.4 & 49.5 & 74.2 & 62.7 & 52.6 & 75.6 \\
ForAda~\cite{ForAda} \C{ForAda-D26CVPR} & 98.1 & 81.3 & 95.5 & 92.8 & 93.1 & 97.3 & 87.3 & 90.7 & 83.0 & 91.6 & 75.6 & 73.3 & 69.5 & 57.5 & 60.2 & 65.1 & 73.1 & 65.7 & 56.5 & 79.3 \\
Effort~\cite{Effort} \C{Effort-D26CVPR} & 96.6 & 91.3 & 93.4 & 95.1 & 92.4 & 98.1 & 86.8 & 96.0 & 83.2 & 90.9 & 78.7 & 65.3 & 64.8 & 62.6 & 73.1 & 75.4 & 72.5 & 57.1 & 44.0 & 79.9 \\
DFD-FCG\textsuperscript{\dag}~\cite{DFD-FCG} \C{DFD-FCG-x1.4-D26CVPR} & 94.7 & 91.8 & 94.0 & 94.0 & 90.6 & 99.6 & 78.5 & 92.8 & 82.7 & 94.6 & \textbf{93.1} & 78.5 & 64.5 & 65.6 & 75.3 & 76.0 & 65.2 & 65.4 & 53.0 & 81.6 \\
VLAForge~\cite{VLAForge} \C{VLAForge-D26CVPR} & 98.1 & 92.9 & 95.0 & 96.8 & 94.5 & 99.5 & 89.7 & 97.6 & 87.0 & 93.9 & 72.4 & 70.0 & 69.0 & 57.9 & 77.2 & 71.7 & 74.0 & 62.8 & 54.3 & 81.8 \\
FS-VFM~\cite{FS-VFM} \C{FS-VFM-D26CVPR} & 97.8 & 80.7 & 94.4 & 92.0 & 97.4 & 98.4 & 90.8 & 94.7 & 92.3 & 97.3 & 85.0 & 79.5 & 74.6 & 77.7 & 61.9 & 67.3 & 81.7 & 60.1 & 57.7 & 83.2 \\
DFD-HR~\cite{DFD-HR} \C{DFD-HR-D26CVPR} & 98.7 & 92.9 & 94.2 & 98.0 & 98.7 & 99.7 & 94.6 & 98.7 & 90.7 & 96.3 & 81.6 & 83.2 & 67.5 & 59.6 & 81.2 & 74.8 & 86.3 & 80.8 & 63.4 & 86.4 \\
GenD-PE~\cite{GenD} \C{GenD-PE-HF-D26CVPR} & 98.6 & 88.3 & 95.9 & 95.4 & 97.5 & 98.8 & 94.3 & 98.0 & 89.8 & 93.7 & 89.9 & 78.5 & 76.6 & 75.3 & 83.7 & 72.3 & 85.1 & 79.8 & 69.7 & 87.4 \\
\midrule
\textbf{\method} \C{eCVPR27-M1-rsLoRA-r32-soup-D26CVPR} & \textbf{98.9} & \textbf{96.8} & \textbf{97.1} & \textbf{98.9} & \textbf{99.9} & \textbf{99.9} & \textbf{97.3} & \textbf{99.5} & \textbf{93.5} & \textbf{98.7} & 90.4 & \textbf{92.1} & \textbf{92.3} & 96.1 & \textbf{98.5} & \textbf{94.9} & \textbf{93.7} & \textbf{95.0} & \textbf{95.8} & \textbf{96.3} \\
\bottomrule
\end{tabular}
} %
\end{table*}

\section{Method}
\label{sec:method}

\method is a frame-level model that outputs the probability that a face crop is fake. It learns from real faces and three kinds of fakes: fakes from the FaceForensics++ dataset, reconstruction twins, and blob twins (\cref{sec:data}). A \emph{twin} is a fake that we make from a real frame. Each fake, twin or not, is paired with the real frame it was made from, and both receive the same augmentation. The only difference between them is the manipulation. The backbone is a pre-trained \PEcoreL adapted with LoRA (\cref{sec:model}). The final model is the weighted average of three runs with different seeds (\cref{sec:training}). At test time, the video score is the mean of its frame scores (\cref{sec:inference}). The flowchart is shown in \cref{fig:overview}.

\subsection{Training data}
\label{sec:data}

\subsubsection{Real faces}
\label{sec:real}

Real frames come from two sources: the FaceForensics++ (FF++) training videos~\cite{FF}, 8 evenly spaced frames per video (5,739 frames), and 1,913 random frames from Casual Conversations v2 (CCv2)~\cite{CCv2}. This gives a 3\,:\,1 ratio of FF++ to CCv2 frames. From FF++, we drop face crops with a side of at most 100\,px, often false detections.

\subsubsection{FF++ fakes}
\label{sec:fake}

We use all four FF++ generators: Deepfakes, Face2Face, FaceSwap, and NeuralTextures. Each fake is cropped with the face box of its real frame, so the two crops match pixel by pixel. We drop fakes whose ground-truth mask covers less than 10\% of the crop, as these are false face detections.

\subsubsection{Reconstruction twins}
\label{sec:rec-twins}

Most recent generators operate in a latent space. Their last step decodes latents into pixels with the decoder of an autoencoder. The decoder leaves traces in every image it outputs, regardless of what the previous generation steps were. A \emph{reconstruction twin} thus teaches \method to detect these traces. We will show that training on the traces of a single decoder is enough to detect fakes from other generators, see \cref{tab:generators}.

We pass every real frame $I$ through the CogVideoX video autoencoder~\cite{CogVideoX}. Its reconstruction $I'$ is aligned with $I$ and differs from it only by the decoder's traces. These traces cover the whole frame. To make the detector find traces that cover only part of a frame, we put the real pixels back inside a region defined by a binary mask $M_\mathrm{r}$.
 
\begin{equation}
  \hat I = M_\mathrm{r} \odot I + (1 - M_\mathrm{r}) \odot I',
  \label{eq:partial}
\end{equation}
and label $\hat I$ as fake, and $\odot$ denotes the pixel-wise product. This partial reconstruction is the reconstruction twin of $I$. The real region covers a random share of 0 to 90\% of the frame. At 0\%, the twin is the full reconstruction. Half of the time, $M_\mathrm{r}$ is a \emph{grid mask}: a random set of cells of a $32 \times 32$ grid. Otherwise, $M_\mathrm{r}$ is a \emph{noise mask}: a set of irregular blobs. To make it, we draw white noise over the frame and blur it with a Gaussian filter whose width is 3 to 9\% of the image side. The real region is where the blurred noise exceeds the threshold that gives exactly the sampled share of the frame. The mask edges are hard. \Cref{fig:overview} shows both masks. Details are in the supplementary \cref{sec:supp-partial}. The masking is meant to reduce overfitting to a specific decoder.

\subsubsection{Blob twins}
\label{sec:blob-twins}

Blending fakes paste a generated/manipulated face into a real frame. This leaves compositing artifacts: the pasted region and the rest of the frame have different image statistics. A \emph{blob twin} teaches \method to detect them. 

A blob twin blends an image $F$ into an image $B$ inside a round \emph{blob} defined by a soft mask $M_\mathrm{b}$. One image, chosen with equal chance, is a real frame $I$. The other is an altered copy of $I$: its contrast and hue are changed, and it is downsampled $2\times$ (nearest-neighbor) and upsampled back (bilinear). The mask blends them into the twin $\tilde I$:
\begin{equation}
  \tilde I = s M_\mathrm{b} \odot F + (1 - s M_\mathrm{b}) \odot B,
  \label{eq:blob}
\end{equation}
where the blend strength $s$ is 0.3 or 1 with equal probability. The twin has the label fake. The mask is $M_\mathrm{b} = 2^{-(d/\rho)^8}$, where $d$ is the distance of a pixel from the blob's center and $\rho$ is the blob's radius. It is close to 1 inside the blob, 0.5 on its edge, and close to 0 outside. The blob is centered at 50\% of the crop's width and 60\% of its height and covers 8 to 50\% of the crop at random. Before blending, a random shift moves $F$ and $M_\mathrm{b}$ together by up to 1.8\% of the crop along each axis, details are in the supplementary \cref{sec:supp-blob}.

\subsubsection{Training pairs}
\label{sec:pairs}

We train on pairs of a fake and the real frame from which it was made, which mitigates shortcut learning~\cite{GenD}. FF++ has four fakes for every real frame, one per manipulation. To have as many twins, we set the number of twin pairs in an epoch to four times the number of FF++ real frames. In each twin pair, the fake is a twin of a real frame of \cref{sec:real}: a reconstruction twin or a blob twin, with equal probability. FF++ fake pairs, re-sampled at random every epoch, add half as many pairs. So each kind of fake makes up about a third of an epoch, which has 34,434 pairs in batches of 64.

Each time a pair is used, both images receive the same new random augmentation, as in~\cite{AlignedForensics}: a horizontal flip, an affine transform, Gaussian blur, brightness and contrast jitter, sharpening, and JPEG compression; details are in the supplementary \cref{sec:supp-augmentation}. The augmentation is applied to the face crop before resizing and normalization (\cref{sec:inference}).

\subsection{Model architecture}
\label{sec:model}

\method uses the vision part of the Perception Encoder~\cite{PE} (\PEcoreL), Meta's pre-trained foundation model, with a $224 \times 224$ input. We add a binary linear classifier to the image embedding from \PEcoreL that predicts the real or fake class. We adapt the backbone with LoRA~\cite{LoRA}, which keeps each pre-trained weight $W_0$ frozen and trains a low-rank update $AB$ of rank $r$:
\begin{equation}
  W = W_0 + \frac{\alpha}{r} A B.
  \label{eq:lora}
\end{equation}
Training starts from $AB = 0$, so at the start of training, the model is equal to the pre-trained one. We use $r = 32$ and $\alpha = 32$, and apply LoRA to the query ($W_q$), key ($W_k$), value ($W_v$), and output ($W_o$) projection matrices of the self-attention in all 24 blocks. This trains 6.29\,M parameters, 1.95\% of the entire model.

\subsection{Training}
\label{sec:training}

\method is trained using a standard cross-entropy loss for 10 epochs, at a constant learning rate of $3 \cdot 10^{-4}$ after one epoch of linear warm-up from $10^{-5}$. Training uses the Adam~\cite{Adam} optimizer and bf16 mixed precision. Instead of weight decay, we use L2-SP~\cite{L2SP} in the formulation of~\cite{L2SPAdam}, which after each Adam step pulls the parameters back toward their initial values (\cref{sec:supp-training} in the supplementary).

A validation set tells little about which checkpoint is the best when the test datasets come from a different distribution~\cite{EoA}. So \method uses no validation set. The number of epochs was fixed in advance and never tuned. The learning rate is a common default for Adam and was not tuned either. With no way to tell which checkpoint is the best, \method does not select one. It uses stochastic weight averaging (SWA)~\cite{SWA}, which averages all checkpoints of the last 5 epochs into a single model. SWA was shown~\cite{SWA} to improve generalization compared to individual checkpoints, needs no held-out data, and has a negligible computational cost. Every experiment in this paper uses the same schedule and the same number of samples per epoch.

We repeat training three times with different random seeds and average the final weights into a single model: a uniform \emph{model soup}~\cite{ModelSoups}. Averaging weights works only if any weighted average of two runs also has a low loss, a property called linear mode connectivity~\cite{LMC}. We expect it to hold because the three runs start from the same pre-trained model, and L2-SP keeps them close to its weights. The soup improves accuracy and out-of-distribution robustness~\cite{ModelSoups}.

Before averaging, we merge each LoRA update into its frozen weight, \cref{eq:lora}, because averaging $A$ and $B$ separately differs from averaging the product $AB$. So \method has the same inference cost as \PEcoreL with a linear classifier.

One training run takes about an hour on one A100 GPU.

\begin{table}[t]
\centering
    \caption{\label{tab:test-datasets}
    \textbf{Summary of test datasets.} The table reports the number of real and fake test videos. For image datasets (\textsuperscript{\dag}) and image subsets (\textsuperscript{\ddag}) each image counts as a video with a single frame. \q{Gen.} is the number of generators.
    }
\tabcolsep=4pt
\begin{tabular}{@{}c@{\hspace{4pt}}lrrrr@{}}
\toprule
& \textbf{Dataset} & \textbf{Venue} & \textbf{Gen.} & \textbf{Real} & \textbf{Fake} \\
\midrule
\multirow{10}{*}{\rotatebox[origin=c]{90}{\textbf{\texttt{Blending}}}} & \texttt{DFD}~\cite{DFD} & Google'19 & 5 & 363 & 3,068 \\
 & \texttt{FSh}~\cite{FSh} & CVPR'20 & 1 & 140 & 140 \\
 & \texttt{CDFv2}~\cite{CDFv2} & CVPR'20 & 1 & 178 & 340 \\
 & \texttt{FFIW}~\cite{FFIW} & CVPR'21 & 3 & 1,357 & 1,357 \\
 & \texttt{FAVC}~\cite{FakeAVCeleb} & NeurIPS'21 & 4 & 500 & 20,566 \\
 & \texttt{DFDM}~\cite{DFDM} & ICIP'22 & 4 & 415 & 1,720 \\
 & \texttt{PGF}~\cite{PolyGlotFake} & ICPR'24 & 10 & 708 & 11,685 \\
 & \texttt{IDF}~\cite{IDForge} & MM'24 & 9 & 18,834 & 2,987 \\
 & \texttt{DSv1}~\cite{DSv1} & CVPR'26 & 5 & 1,381 & 1,395 \\
 & \texttt{HiDF}~\cite{HiDF} & KDD'25 & 2 & 4,361 & 4,361 \\
\midrule
\multirow{4}{*}{\rotatebox[origin=c]{90}{\textbf{\texttt{B+EFS}}}} & \texttt{CDFv3}~\cite{CDFv3} & arXiv'25 & 22 & 178 & 5,239 \\
 & \texttt{DSv2}~\cite{DSv1} & CVPR'26 & 6 & 1,766 & 1,289 \\
 & \texttt{RF}\textsuperscript{\dag}~\cite{RedFace} & arXiv'25 & 9 & 7,411 & 4,536 \\
 & \texttt{HF}\textsuperscript{\ddag}~\cite{HydraFake} & ICLR'26 & 20 & 15,907 & 11,078 \\
\midrule
\multirow{5}{*}{\rotatebox[origin=c]{90}{\textbf{\texttt{EFS}}}} & \texttt{THB}~\cite{TalkingHeadBench} & WACV'26 & 5 & 334 & 844 \\
 & \texttt{OF}~\cite{OmniFake} & CVPR'26 & 4 & 3,146 & 3,400 \\
 & \texttt{DFAV}~\cite{DigiFakeAV} & ICASSP'26 & 3 & 8,223 & 21,597 \\
 & \texttt{MV}~\cite{MintVid} & ICML'26 & 3 & 705 & 2,114 \\
 & \texttt{DF26}~\cite{DF26} & arXiv'26 & 7 & 273 & 2,423 \\
\midrule
\multicolumn{2}{@{}l}{\textbf{Total} (19)} &  & -- & 66,180 & 100,139 \\
\bottomrule
\end{tabular}
\end{table}

\begin{table*}[t]
\centering
\caption{\label{tab:generators}
    \textbf{Results per generator.} Video-level AUROC (\%) on \texttt{DF26}, \texttt{OF} and \texttt{MV} for each of their generators. The row of \method is the AUROC. Every other row is the difference to \method in percentage points, green if higher and red if lower. The prior detectors are the four best AIGC detectors and the four best facial deepfake detectors of \cref{tab:sota}. The other seven are in \cref{tab:supp-generators} in the supplementary. The rows under \method make the reconstruction twins with another autoencoder instead of CogVideoX. HV is HunyuanVideo, T2V is text-to-video and I2V is image-to-video. Mean is the average over the 17 generators. The best value in each column is in bold.
}
\tabcolsep=3.4pt
\resizebox{\textwidth}{!}{
\begin{tabular}{@{}l|cccccccccc|cccc|ccc|c@{}}
\toprule
 & \multicolumn{10}{c|}{\texttt{DF26}~\cite{DF26}} & \multicolumn{4}{c|}{\texttt{OF}~\cite{OmniFake}} & \multicolumn{3}{c|}{\texttt{MV}~\cite{MintVid}} & \\
 & \multicolumn{4}{c}{\it closed-source} & \multicolumn{6}{c|}{\it open-source} & \multicolumn{4}{c|}{} & \multicolumn{3}{c|}{} & \\[-2ex]
 & \multicolumn{4}{@{\hskip 4pt}c@{\hskip 4pt}}{\leaders\hrule height 0.3pt\hskip 0pt plus 1filll\null} & \multicolumn{6}{@{\hskip 4pt}c@{\hskip 4pt}|}{\leaders\hrule height 0.3pt\hskip 0pt plus 1filll\null} & \multicolumn{4}{c|}{} & \multicolumn{3}{c|}{} & \\
 & \rotatebox{90}{Grok T2V} & \rotatebox{90}{Kling T2V} & \rotatebox{90}{Veo3 T2V} & \rotatebox{90}{Wan 2.6 T2V} & \rotatebox{90}{HV 1.5 I2V} & \rotatebox{90}{HV 1.5 T2V} & \rotatebox{90}{Wan 2.2 I2V} & \rotatebox{90}{Wan 2.2 T2V} & \rotatebox{90}{LTX 2.3 I2V} & \rotatebox{90}{LTX 2.3 T2V} & \rotatebox{90}{ACTalker} & \rotatebox{90}{Ditto} & \rotatebox{90}{EDTalk} & \rotatebox{90}{FLOAT} & \rotatebox{90}{FantasyTalk.} & \rotatebox{90}{OmniAvatar} & \rotatebox{90}{Phantom 14B} & \rotatebox{90}{Mean} \\
\midrule
\multicolumn{19}{@{}l}{\textit{AIGC detectors}} \\[2pt]
LTD-UFD~\cite{LTD} \C{LTD-UFD-D26CVPR} & \dn{2.7} & \dn{40.7} & \dn{0.5} & \up{2.8} & \dn{9.3} & \dn{2.8} & \dn{16.4} & \up{0.7} & \dn{13.2} & \dn{4.7} & \dn{34.5} & \dn{64.3} & \dn{79.6} & \dn{60.4} & \dn{64.5} & \dn{47.1} & \dn{5.6} & \dn{26.0} \\
AlignedFor.~\cite{AlignedForensics} \C{AlignedForensics-Sync-D26CVPR} & \dn{5.4} & \textbf{\boldmath\up{3.5}} & \dn{15.0} & \dn{4.0} & \dn{40.5} & \dn{51.0} & \dn{48.6} & \dn{54.4} & \dn{21.4} & \dn{21.9} & \dn{2.1} & \dn{33.4} & \dn{30.9} & \dn{64.1} & \dn{86.5} & \dn{64.4} & \dn{13.9} & \dn{32.6} \\
DDA~\cite{DDA} \C{DDA-D26CVPR} & \dn{8.4} & \up{1.6} & \dn{9.8} & \textbf{\boldmath\up{7.6}} & \dn{14.8} & \dn{28.3} & \dn{25.9} & \dn{33.3} & \dn{24.0} & \dn{34.7} & \dn{0.2} & \dn{70.1} & \dn{66.4} & \dn{62.0} & \dn{54.1} & \dn{11.1} & \dn{6.1} & \dn{25.9} \\
B-Free~\cite{B-Free} \C{B-Free-D26CVPR} & \dn{53.0} & \dn{23.2} & \dn{41.1} & \up{6.5} & \dn{20.1} & \dn{18.3} & \dn{22.2} & \dn{33.4} & \dn{21.6} & \dn{41.6} & \dn{0.1} & \dn{48.5} & \dn{62.8} & \dn{46.0} & \dn{27.7} & \dn{9.4} & \dn{23.4} & \dn{28.6} \\
\midrule
\multicolumn{19}{@{}l}{\textit{Facial deepfake detectors}} \\[2pt]
VLAForge~\cite{VLAForge} \C{VLAForge-D26CVPR} & \dn{64.0} & \dn{36.1} & \dn{64.0} & \dn{30.2} & \dn{46.1} & \dn{20.2} & \dn{33.8} & \dn{46.9} & \dn{33.7} & \dn{49.3} & \dn{5.4} & \dn{34.8} & \dn{15.0} & \dn{26.9} & \dn{48.5} & \dn{18.7} & \dn{29.2} & \dn{35.5} \\
FS-VFM~\cite{FS-VFM} \C{FS-VFM-D26CVPR} & \dn{43.7} & \dn{35.2} & \dn{40.3} & \dn{16.5} & \dn{48.0} & \dn{19.2} & \dn{32.5} & \dn{51.5} & \dn{27.0} & \dn{47.4} & \dn{4.7} & \dn{36.8} & \dn{16.8} & \dn{38.4} & \dn{44.4} & \dn{17.5} & \dn{42.7} & \dn{33.1} \\
DFD-HR~\cite{DFD-HR} \C{DFD-HR-D26CVPR} & \dn{53.5} & \dn{23.1} & \dn{45.2} & \dn{27.0} & \dn{33.8} & \dn{15.2} & \dn{22.2} & \dn{45.4} & \dn{13.8} & \dn{40.0} & \dn{2.0} & \dn{31.1} & \dn{6.1} & \dn{30.3} & \dn{22.5} & \dn{4.7} & \dn{15.3} & \dn{25.4} \\
GenD-PE~\cite{GenD} \C{GenD-PE-HF-D26CVPR} & \dn{30.9} & \dn{24.2} & \dn{18.6} & \dn{3.1} & \dn{30.3} & \dn{17.0} & \dn{29.2} & \dn{35.0} & \dn{16.2} & \dn{25.1} & \dn{2.8} & \dn{37.0} & \dn{7.2} & \dn{31.4} & \dn{27.3} & \dn{6.9} & \dn{11.4} & \dn{20.8} \\
\midrule
\textbf{\method} \C{eCVPR27-M1-rsLoRA-r32-soup-D26CVPR} & 97.4 & 95.2 & 95.1 & 91.4 & 95.1 & 99.7 & \textbf{95.3} & 93.7 & 97.6 & 96.5 & \textbf{99.9} & \textbf{88.8} & \textbf{98.9} & \textbf{95.0} & \textbf{92.9} & \textbf{92.1} & 99.9 & \textbf{95.6} \\
\quad w/ KVAE 2D \C{eCVPR27-M2-dec-kvae2d-soup-D26CVPR} & \dn{1.3} & \dn{7.2} & \dn{1.0} & \dn{2.6} & \dn{4.2} & \dn{1.4} & \dn{7.4} & \dn{18.2} & \dn{2.9} & \dn{2.0} & \dn{0.1} & \dn{10.5} & \dn{4.2} & \dn{7.3} & \dn{19.9} & \dn{15.5} & \textbf{\boldmath\up{0.1}} & \dn{6.2} \\
\quad w/ LTX-Video \C{eCVPR27-M2-dec-ltxvideo-soup-D26CVPR} & \up{0.7} & \dn{6.5} & \up{3.8} & \up{3.2} & \up{1.0} & \dn{0.2} & \dn{8.0} & \dn{10.7} & \dn{0.4} & \up{1.5} & \dn{0.2} & \dn{10.5} & \dn{4.0} & \dn{12.4} & \dn{10.4} & \dn{10.7} & \textbf{\boldmath\up{0.1}} & \dn{3.7} \\
\quad w/ HV 1.0 \C{eCVPR27-M2-dec-hunyuanvideo-soup-D26CVPR} & \up{1.1} & \dn{6.3} & \up{1.2} & \up{1.1} & \up{0.5} & \dn{0.4} & \dn{1.4} & \dn{1.3} & \dn{1.0} & \up{1.3} & \dn{1.3} & \dn{18.9} & \dn{7.6} & \dn{15.4} & \dn{4.9} & \dn{4.7} & \textbf{\boldmath\up{0.1}} & \dn{3.4} \\
\quad w/ LTX-2 \C{eCVPR27-M2-dec-ltx2-soup-D26CVPR} & \textbf{\boldmath\up{1.8}} & \dn{8.7} & \textbf{\boldmath\up{4.1}} & \textbf{\boldmath\up{7.6}} & \textbf{\boldmath\up{3.2}} & \textbf{\boldmath\up{0.2}} & \dn{0.3} & \textbf{\boldmath\up{1.6}} & \textbf{\boldmath\up{0.9}} & \textbf{\boldmath\up{2.7}} & \textbf{\boldmath\ns{0.0}} & \dn{25.9} & \dn{8.7} & \dn{27.5} & \dn{1.3} & \dn{5.8} & \textbf{\boldmath\up{0.1}} & \dn{3.3} \\
\bottomrule
\end{tabular}
} %
\end{table*}

\subsection{Inference}
\label{sec:inference}

From each test video we take up to 32 evenly spaced frames. In each frame, we detect the face, align it with a similarity transform on its five keypoints, enlarge the face box $1.3\times$, and crop it (\cref{sec:preprocessing}). The crop is then resized (bicubic) to $224 \times 224$ and normalized to $[-1, 1]$ range (mean and standard deviation 0.5 per channel), the standard preprocessing of \PEcoreL~\cite{PE}. \method gives the probability that a frame is fake, and the video score is the mean over its frames. With batches of 128 frames, the torch-compiled model processes about 1,000 frames per second on one A100 GPU.

\begin{table}[t]
\centering
\caption{\label{tab:ablations}
    \textbf{Ablations.} The first row is the video-level AUROC (\%) of the single models averaged into \method, mean of 3 seeds. Each other row changes one thing in training and shows the change in percentage points. Gray: mean$\pm$std intervals overlap the first row.
}
\small\tabcolsep=2pt
\begin{tabular}{@{}l|rrr|r@{}}
\toprule
 & \textbf{\texttt{Blending}} & \textbf{\texttt{B+EFS}} & \textbf{\texttt{EFS}} & \bf All \\
\midrule
\textbf{\method} single models (\%) & 97.9 & 92.6 & 95.3 & 96.1 \\
\midrule
\multicolumn{5}{@{}l}{\textit{Training fakes}} \\
\quad w/o FF++ fakes           & \dn{1.8}  & \dn{3.0}  & \ns{$-$0.7} & \dn{1.8} \\
\quad w/o blob twins           & \dn{5.8}  & \ns{$-$0.4} & \ns{$-$0.9} & \dn{3.4} \\
\quad w/o reconstruction twins & \ns{$+$0.4} & \dn{10.3} & \dn{18.1} & \dn{6.7} \\
\quad blob twins only          & \dn{0.7}  & \dn{15.2} & \dn{21.9} & \dn{9.3} \\
\quad FF++ fakes only          & \dn{1.3}  & \dn{15.7} & \dn{24.4} & \dn{10.4} \\
\quad reconstruction twins only & \dn{33.0} & \dn{9.4} & \dn{3.9}  & \dn{20.4} \\
\midrule
\multicolumn{5}{@{}l}{\textit{Design of twins}} \\
\quad w/o partial reconstructions & \ns{0.0} & \dn{3.5} & \dn{2.7} & \dn{1.5} \\
\quad SBI instead of blob twins & \dn{0.9} & \ns{$-$0.1} & \ns{$-$0.4} & \dn{0.6} \\
\midrule
\multicolumn{5}{@{}l}{\textit{Training sources}} \\
\quad w/o CCv2 reals and twins  & \dn{1.8} & \dn{2.4} & \ns{$-$1.2} & \dn{1.8} \\
\midrule
\multicolumn{5}{@{}l}{\textit{Training}} \\
\quad w/o L2-SP                & \ns{$-$0.2} & \ns{$-$1.3} & \ns{$-$1.0} & \dn{0.7} \\
\midrule
\multicolumn{5}{@{}l}{\textit{LoRA rank}} \\
\quad $r = 2$                  & \ns{$-$0.3} & \ns{$-$2.3} & \ns{$-$0.8} & \dn{0.9} \\
\quad $r = 4$                  & \ns{0.0} & \ns{$-$1.4} & \ns{$-$0.1} & \dn{0.4} \\
\quad $r = 8$                  & \ns{$+$0.2} & \ns{$-$1.1} & \ns{$-$0.1} & \ns{$-$0.2} \\
\quad $r = 16$                 & \ns{$+$0.1} & \ns{$-$0.4} & \ns{$-$0.1} & \ns{$-$0.1} \\
\quad $r = 64$                 & \ns{$-$0.1} & \ns{$-$0.9} & \ns{$-$0.3} & \dn{0.3} \\
\bottomrule
\end{tabular}
\end{table}

\section{Experiments}
\label{sec:experiments}

\subsection{Benchmark}
\label{sec:benchmark}

We evaluate all methods on 19 datasets released between 2019 and 2026, using test splits where available (otherwise, the full dataset). We divide them into three groups by how their generators create fakes: \texttt{Blending}, \texttt{EFS}, and \texttt{B+EFS}. We use these groups throughout for analysis.

\paragraph{\texttt{Blending}} Every generator blends the generated content back into the original frame. These 10 datasets are: Google's \texttt{DFD} dataset~\cite{DFD}, FaceShifter (\texttt{FSh})~\cite{FSh}, Celeb-DF-v2 (\texttt{CDFv2})~\cite{CDFv2}, Face Forensics in the Wild (\texttt{FFIW})~\cite{FFIW}, FakeAVCeleb (\texttt{FAVC})~\cite{FakeAVCeleb}, DeepFakes from Different Models (\texttt{DFDM})~\cite{DFDM}, PolyGlotFake (\texttt{PGF})~\cite{PolyGlotFake}, IDForge (\texttt{IDF})~\cite{IDForge}, DeepSpeak v1.1 (\texttt{DSv1})~\cite{DSv1}, and Human-Indistinguishable Deepfake Dataset (\texttt{HiDF})~\cite{HiDF}. 

\paragraph{\texttt{EFS}} Every generator synthesizes the entire video frame with a face, which is called entire face synthesis (EFS). No step of the deepfake production pipeline blends generated content back into the original frame. These 5 datasets are: TalkingHeadBench (\texttt{THB})~\cite{TalkingHeadBench}, OmniFake (\texttt{OF})~\cite{OmniFake}, DigiFakeAV (\texttt{DFAV})~\cite{DigiFakeAV}, MintVid (\texttt{MV})~\cite{MintVid}, and \texttt{DF26}~\cite{DF26}. 

\paragraph{\texttt{B+EFS}} These 4 datasets contain both blending and EFS generators: Celeb-DF++ (\texttt{CDFv3})~\cite{CDFv3}, DeepSpeak v2.0 (\texttt{DSv2})~\cite{DSv1}, RedFace (\texttt{RF})~\cite{RedFace}, and HydraFake (\texttt{HF})~\cite{HydraFake}.

\paragraph{Cleaning} We clean the test datasets in two ways. First, in \texttt{Blending} datasets, where each fake can be matched with its source video, we remove failed fakes that are pixel-identical to their source. Second, we remove videos where the face in the original footage is smaller than 100\,px on any side. Such detections are often not faces at all, and when they are, the fakes are too small and blurred to deceive anyone.

\paragraph{No overlap with training data} The training and test data do not overlap. \texttt{FF++}~\cite{FF} is the common training set in facial deepfake detection. The videos of CCv2~\cite{CCv2} are unique recordings, and no test dataset uses them. To confirm this, we ranked CCv2 faces by their similarity to the faces of every test dataset using the face recognition model from the InsightFace \texttt{buffalo\_l} pack~\cite{InsightFace} and inspected the 100 most similar pairs per dataset. None showed the same person. Importantly, no test dataset contains fakes generated by CogVideoX~\cite{CogVideoX}, whose autoencoder makes the reconstruction twins (\cref{sec:rec-twins}), and no open-source generator in the test datasets is built on it. \Cref{tab:test-datasets} lists the number of generators and real and fake videos in each dataset.

\paragraph{Evaluation metric} The primary metric is the video-level area under the ROC curve (AUROC), where each score represents one video, as described in \cref{sec:inference}. The equal error rate (EER) and the mean average precision over the two classes (mAP) are reported in the supplementary \cref{sec:supp-eer-map}.

\subsection{Preprocessing}
\label{sec:preprocessing}

All methods, including \method, get the same face crops. We follow DeepfakeBench~\cite{DeepfakeBench} but use a state-of-the-art face detector, SCRFD-10GF~\cite{SCRFD} from the InsightFace \texttt{buffalo\_l} model pack~\cite{InsightFace}. From each test video we take up to 32 evenly spaced frames. In each frame, we detect the face and align it with a similarity transform on its five keypoints, which keeps the face close to its original size. We then enlarge the face box $1.3\times$ and crop it. Each method resizes and normalizes the crop with its own preprocessing, the one it was trained with (for \method, see \cref{sec:inference}).

\subsection{Comparison with the state-of-the-art}
\label{sec:sota}

\Cref{tab:sota} compares \method with six AIGC detectors and nine facial deepfake detectors. We run the released model checkpoints of every method on the same face crops (\cref{sec:preprocessing}).

\paragraph{Overall} \method reaches a mean AUROC of 96.3\%, 8.9 percentage points above GenD-PE~\cite{GenD}, the best prior detector. \method is the best method on 17 of the 19 datasets. The two exceptions are \texttt{CDFv3}, where the temporal model DFD-FCG~\cite{DFD-FCG} is 2.7 points higher, and \texttt{HF}, where the AIGC detector B-Free~\cite{B-Free} is 2.4 points higher. \method is at least 90.4\% on every dataset, signifying no major failures.

\paragraph{\texttt{Blending}} Prior facial deepfake detectors are strong on \texttt{Blending}, where DFD-HR~\cite{DFD-HR} reaches a mean of 96.2\%. \method reaches 98.1\% and is the best on each dataset.

\paragraph{\texttt{EFS}} Prior detectors fall behind on the five \texttt{EFS} datasets. The best prior facial deepfake detector, GenD-PE, reaches a mean of 78.1\%, and the best AIGC detector, B-Free, 73.8\%. \method reaches 95.6\%. On the four \texttt{B+EFS} datasets, \method reaches 92.7\% against 80.1\% for GenD-PE.

\paragraph{AIGC detectors} AIGC detectors generalize poorly on facial datasets. Over all 19 datasets, B-Free, the best one, also reaches 73.8\%, with 98.5\% on \texttt{HF} but 48.6\% on \texttt{OF}.

\paragraph{Training set of \method} \method trains on 38,211 frames. Of these, 7,652 are real: 5,739 from FF++, 8 per video, and 1,913 from CCv2. The rest are fake: 7,652 CogVideoX reconstructions, one per real frame, and 22,907 FF++ fake frames from four FF++ manipulations. Blob twins and the masks of reconstruction twins are generated during training.

\paragraph{Training sets of prior detectors} The gain of \method in \cref{tab:sota} cannot be simply explained by the number of training samples. Prior facial deepfake detectors sample FF++ at least as densely. SBI~\cite{SBI}, Effort~\cite{Effort} and DFD-HR also sample 8 frames per FF++ video. ForAda~\cite{ForAda} and GenD-PE sample 32, and GenD-PE, the best prior detector, trains on about 115,000 frames. FSFM~\cite{FSFM} samples 128 frames per real video and pre-trains on about 3.3\,M face images. The AIGC detectors that report their training set, AlignedForensics~\cite{AlignedForensics}, DDA~\cite{DDA} and B-Free, train on 51,517 to 179,257 real images and 118,000 to 309,102 fake images.

\subsection{Generalization across generators}
\label{sec:generators}

The EFS fakes of \texttt{DF26}~\cite{DF26}, \texttt{OF}~\cite{OmniFake} and \texttt{MV}~\cite{MintVid} are labeled with the generator that made them. \Cref{tab:generators} gives the AUROC of \method for each generator and, for every other detector, the difference to \method. \texttt{DF26} has seven generators. Four are closed-source, and the three open-source ones each have a text-to-video and an image-to-video mode. With the four generators of \texttt{OF} and the three of \texttt{MV}, the table has 17 generators. The table shows the four best AIGC and the four best facial deepfake detectors of \cref{tab:sota}. The other seven detectors are in \cref{tab:supp-generators} in the supplementary. The same comparison for the generators of \texttt{THB}, \texttt{DFAV} and \texttt{HF} is in the supplementary (\cref{tab:supp-generators-thb-dfav,tab:supp-generators-hf}), where some prior detectors are above \method on single generators.

\paragraph{Unseen generators} \method saw none of these generators in training (\cref{sec:benchmark}). Its AUROC is 95.6\% on average and at least 88.8\% in every column. On the four closed-source generators, it is between 91.4\% and 97.4\%.

\paragraph{Prior detectors} GenD-PE~\cite{GenD}, the best prior detector on average, is 20.8 points below \method. Prior detectors are also uneven across generators. B-Free~\cite{B-Free} is 0.1 points below \method on ACTalker but 62.8 points below on EDTalk. AlignedForensics~\cite{AlignedForensics} is 3.5 points above \method on Kling but 86.5 points below on FantasyTalking. In the table, a prior detector is above \method in six cells only. All six are AIGC detectors on Kling, Wan 2.6 or Wan 2.2. No prior facial deepfake detector is above \method in any of the 17 columns of \cref{tab:generators}. None of the seven detectors of \cref{tab:sota} that the table leaves out is above \method on even a single generator (\cref{tab:supp-generators} in the supplementary).

\vspace{10pt}
\paragraph{Other autoencoders} The last four rows of \cref{tab:generators} make the reconstruction twins with another autoencoder instead of CogVideoX (\cref{sec:rec-twins}): KVAE 2D~\cite{KVAE}, LTX-Video~\cite{LTX-Video}, HunyuanVideo 1.0~\cite{HunyuanVideo-1.0} or LTX-2~\cite{LTX-2}. Each of them is trained exactly as \method and is also a soup of three seeds (\cref{sec:training}). Only the autoencoder differs. None of the open-source generators in \cref{tab:generators} decodes with CogVideoX or with one of the four other autoencoders. The autoencoders of the four closed-source generators are not public. On average, all four other autoencoders are below CogVideoX, by 3.3 to 6.2 points. This is still far above GenD-PE ($-20.8$). So the gain of \method does not depend on the choice of CogVideoX. No autoencoder is the best on every generator. LTX-2 is above CogVideoX in 8 of the 10 columns of \texttt{DF26}, by up to 7.6 points, but below it on \texttt{OF}, by up to 27.5 points. All four are below CogVideoX on Kling, Ditto and FLOAT.

\begin{figure*}[t]
\centering
\includegraphics[width=\textwidth]{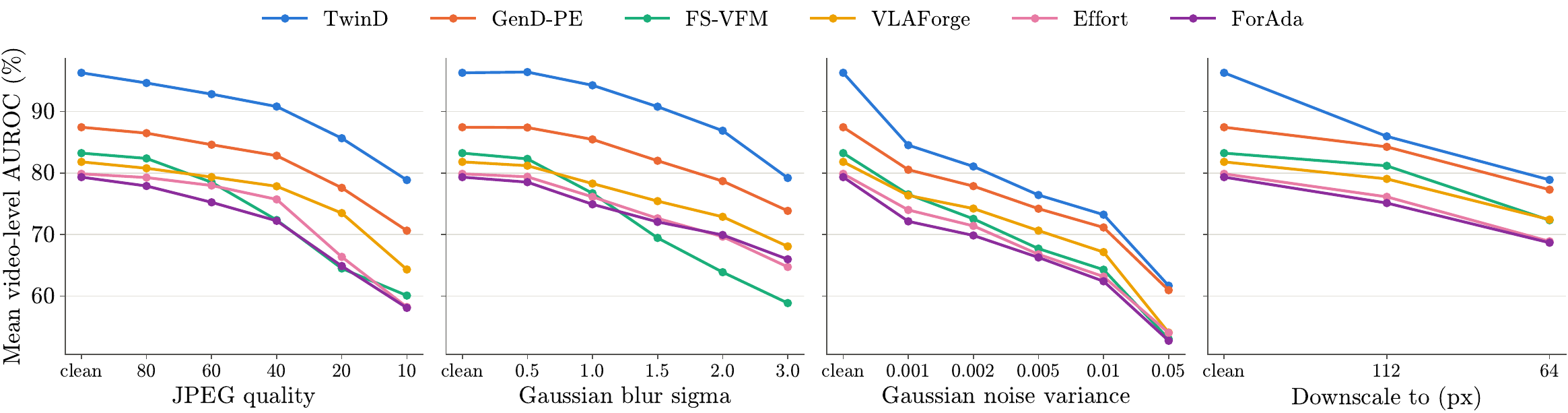}
\caption{\label{fig:degradation}
    \textbf{Robustness to image degradations.} Mean video-level AUROC (\%) over the 19 test datasets as the strength of each degradation increases. \q{Clean} is the undegraded benchmark of \cref{tab:sota}.
}
\end{figure*}

\subsection{Ablations}
\label{sec:ablations}

We conducted an ablation study, see \cref{tab:ablations}. Every row is the mean AUROC of three models trained with different seeds, which allows us to distinguish the effect of a change from seed variance. Therefore, the first row is the mean AUROC of three models that make the soup of \method (\cref{sec:training}), not the AUROC of the soup itself from \cref{tab:sota}. Each subsequent row changes one factor in training and reports the change compared to the first row. In every row, an epoch has 34,434 pairs. When a kind of fake is removed, the epoch is filled with pairs of the other kinds. The supplementary material gives further details (\cref{sec:supp-ablation-fakes}) and the AUROC on each dataset (\cref{tab:supp-ablation-fakes,tab:supp-ablation-sbi,tab:supp-ablation-partial}).

\paragraph{FF++ fakes only} This row is the usual training protocol for the facial deepfake detectors in \cref{tab:sota}: training on FF++ reals and FF++ fakes, with no CCv2 reals and no twins. \texttt{Blending} drops slightly ($-1.3$), but \texttt{EFS} drops by 24.4 points. This indicates that FF++ fakes mostly teach the compositing artifacts, while EFS generators leave no such traces. On All, this detector reaches 85.7\%, which is 1.7 points below GenD-PE~\cite{GenD}, the best prior detector. GenD-PE samples four times more frames per FF++ video (32 against 8). So, on the standard training data, the model and the training recipe of \method do not surpass prior detectors. The gain of \method comes from the training data: mainly the twins and partially the CCv2 reals.

\paragraph{Twins alone surpass prior detectors} Without FF++ fakes, all fakes are twins made from the FF++ and CCv2 reals. No fake in the training set changes the identity or the expression of a face. This detector reaches 94.3\% on All, 6.9 points above GenD-PE, the best prior detector. Compared to the full training set, it loses 1.8 points, mostly on \texttt{Blending} ($-1.8$) and \texttt{B+EFS} ($-3.0$), while the change on \texttt{EFS} ($-0.7$) is within the seed variance. So, state-of-the-art generalization needs no face swap or reenactment in the training set. Here, \emph{twins alone surpass all prior detectors.} Making the twins from more diverse reals helps further. Trained on FF++ reals and their twins only, without FF++ fakes and CCv2, the detector reaches 91.3\% (\cref{tab:supp-ccv2-reals} in the supplementary). This is 3.0 points lower but still above prior work.

\paragraph{Each kind of twin covers one kind of fake} Without reconstruction twins, \texttt{EFS} drops by 18.1 points, while \texttt{Blending} is within the seed variance. Therefore, \texttt{EFS} accuracy comes from reconstruction twins. Without blob twins, \texttt{Blending} drops by 5.8 points, while \texttt{EFS} is within the seed variance.  

\paragraph{No single kind of fake is enough} With reconstruction twins only, \texttt{Blending} drops by 33.0 points, because the decoder's traces do not teach blending artifacts. Blob twins only, made without any generator, do at least as well as FF++ fakes only ($-9.3$ vs $-10.4$ on All).

\paragraph{Design of twins} Putting back the real pixels into the reconstruction twins matters for \texttt{EFS}. Without partial reconstructions, \texttt{EFS} drops by 2.7 points and \texttt{B+EFS} by 3.5, while \texttt{Blending} does not change. Replacing blob twins with the more complex self-blended images (SBI)~\cite{SBI} lowers \texttt{Blending} by 0.9 points (\cref{sec:supp-ablation-sbi} in the supplementary).

\paragraph{CCv2 reals} FF++ reals are mostly news-style YouTube videos with similar framing and compression. The CCv2 reals add many identities, cameras, and lighting conditions. Removing them and their twins from the full training set lowers All by 1.8 points. With CCv2, the detector improves mostly on the real videos of the test datasets and not on their fakes (\cref{sec:supp-gap-split} in the supplementary). In a training set without FF++ fakes, the CCv2 reals help only together with their twins (\cref{sec:supp-ccv2-reals} in the supplementary).

\paragraph{L2-SP} Removing L2-SP lowers All by 0.7 points, see supplementary \cref{sec:supp-training}.

\paragraph{LoRA rank} We started with $r = 8$, the rank used by DDA~\cite{DDA}, and then trained \method with ranks from 2 to 64. With a fixed $\alpha$, the factor $\alpha / r$ in \cref{eq:lora} gets smaller as the rank grows. So we follow rank-stabilized LoRA~\cite{rsLoRA} and set $\alpha = \sqrt{32 r}$, which gives $\alpha = 32$ at $r = 32$. The sweep then changes the dimensionality of the update, not the scale. The rank matters little. The mean AUROC on All grows from 95.2\% at $r = 2$ to 96.1\% at $r = 32$ and drops to 95.8\% at $r = 64$. Ranks 8 and 16 are within the seed variance of rank 32. Every rank surpasses all prior detectors. Even $r = 2$ is 7.8 points above GenD-PE, the best of them (\cref{tab:sota}). 

\subsection{Robustness to image degradations}
\label{sec:robustness}

Videos shared online are often recompressed, resized, or of low quality. We test how \method and five facial deepfake detectors from \cref{tab:sota}, GenD-PE~\cite{GenD}, FS-VFM~\cite{FS-VFM}, VLAForge~\cite{VLAForge}, Effort~\cite{Effort}, and ForAda~\cite{ForAda}, hold up under four degradations. We leave out DFD-HR because its released code is 15 times slower than \method, which makes it too expensive to run on every degradation. We apply each degradation to the face crops of \cref{sec:preprocessing}, before the preprocessing pipeline of each method, one degradation and one strength at a time.
JPEG compression with PIL uses quality 80, 60, 40, 20, and 10.
Gaussian blur uses $\sigma$ of 0.5, 1.0, 1.5, 2.0, and 3.0\,px. The kernel size is $2\lceil 3\sigma \rceil + 1$.
Gaussian noise has zero mean and variance of 0.001, 0.002, 0.005, 0.01, and 0.05 on pixel values in $[0, 1]$. Its standard deviation is the square root of the variance, rounded to three decimals. It is drawn independently for each pixel and color channel, and the result is clipped to $[0, 1]$.
Downscaling resizes the square face crop to $112 \times 112$ and $64 \times 64$\,px with nearest-neighbor interpolation.

\Cref{fig:degradation} shows the results. Averaged over the 19 datasets, \method has the highest AUROC under every degradation and strength. Under Gaussian noise and downscaling, the margin over GenD-PE narrows, and at the strongest noise the two methods are on par. Like GenD-PE and Effort, \method trains with JPEG compression and Gaussian blur (\cref{sec:supp-augmentation} in the supplementary). It trains without noise or downscaling.

\section{Limitations and future work}
\label{sec:limitations}

\paragraph{Noise and downscaling} Both lower the AUROC of \method more than for prior detectors (\cref{fig:degradation}), most on \texttt{EFS} (\cref{tab:supp-degradation} in the supplementary). Unlike JPEG and blur, neither is used in training, and both likely mask high-frequency decoder traces. Training for robustness to them is future work.

\paragraph{One autoencoder} Reconstruction twins come from a single autoencoder. No autoencoder that we tested is the best on every generator (\cref{tab:generators}). Making twins with several autoencoders at once is future work.

\section{Conclusions}
\label{sec:conclusions}

The cross-dataset generalization of \method comes from two complementary training signals, the twins: fakes made from a real frame and paired with it. Reconstruction twins target the traces of a decoder, and blob twins the traces of blending. Twins alone surpass all 15 prior detectors. The proposed \method fine-tunes the \PEcoreL foundation model with techniques to prevent overfitting: LoRA, L2-SP, and weight averaging. \method outperforms the prior art by a large margin. Against 15 recent detectors on 19 datasets from 2019 to 2026, \method reaches a mean AUROC of 96.3\%, 8.9 points above the best of them, and at least 90.4\% on each dataset.

\section*{LLM Usage Statement}

\noindent\textbf{Tool.} Claude Code (CC) from Anthropic was used in the research and in the writing of this paper.

\noindent\textbf{Usage:}
\begin{itemize}
    \item \textbf{Ideation:} CC wrote quick throwaway code that allowed us to iterate over ideas.
    \item \textbf{Code:} CC wrote and debugged parts of the training, evaluation, and plotting code.
    \item \textbf{Experiments and analysis:} CC cross-checked the numbers in the text against the result tables.
    \item \textbf{Related work:} CC helped find related work and facts about prior detectors and datasets.
    \item \textbf{Text:} CC edited the text for clarity, grammar, and consistency.
    \item \textbf{Figures:} the fake faces in \cref{fig:overview} are examples of the data: an FF++ fake and twins made by \method, including a reconstruction by the CogVideoX autoencoder. No generative AI images are present elsewhere.
\end{itemize}

\noindent\textbf{Responsibility.} The authors reviewed all output of CC and are responsible for the content of this paper.

{
    \small
    \interlinepenalty=10000
    \bibliographystyle{ieeenat_fullname}
    \bibliography{main}

\begin{thebibliography}{60}
\providecommand{\natexlab}[1]{#1}
\providecommand{\url}[1]{\texttt{#1}}
\expandafter\ifx\csname urlstyle\endcsname\relax
  \providecommand{\doi}[1]{doi: #1}\else
  \providecommand{\doi}{doi: \begingroup \urlstyle{rm}\Url}\fi

\bibitem[Arpit et~al.(2022)Arpit, Wang, Zhou, and Xiong]{EoA}
Devansh Arpit, Huan Wang, Yingbo Zhou, and Caiming Xiong.
\newblock Ensemble of averages: Improving model selection and boosting performance in domain generalization.
\newblock In \emph{Advances in Neural Information Processing Systems 35: Annual Conference on Neural Information Processing Systems 2022, NeurIPS 2022, New Orleans, LA, USA, November 28 - December 9, 2022}, 2022.

\bibitem[Barrington et~al.(2026)Barrington, Bohacek, and Farid]{DSv1}
Sarah Barrington, Maty Bohacek, and Hany Farid.
\newblock The deepspeak dataset.
\newblock In \emph{Proceedings of the IEEE/CVF conference on computer vision and pattern recognition}, pages 1893--1902, 2026.

\bibitem[Bolya et~al.(2025)Bolya, Huang, Sun, Cho, Madotto, Wei, Ma, Zhi, Rajasegaran, Rasheed, Wang, Monteiro, Xu, Dong, Ravi, Li, Doll{\'{a}}r, and Feichtenhofer]{PE}
Daniel Bolya, Po{-}Yao Huang, Peize Sun, Jang~Hyun Cho, Andrea Madotto, Chen Wei, Tengyu Ma, Jiale Zhi, Jathushan Rajasegaran, Hanoona Rasheed, Junke Wang, Marco Monteiro, Hu Xu, Shiyu Dong, Nikhila Ravi, Shang{-}Wen Li, Piotr Doll{\'{a}}r, and Christoph Feichtenhofer.
\newblock Perception encoder: The best visual embeddings are not at the output of the network.
\newblock In \emph{Advances in Neural Information Processing Systems 38: Annual Conference on Neural Information Processing Systems 2025, NeurIPS 2025, San Diego, CA, USA, December 2-7, 2025 / Mexico City, Mexico, November 30 - December 5, 2025}, 2025.

\bibitem[Chen et~al.(2024)Chen, Zeng, Yang, and Yang]{DRCT}
Baoying Chen, Jishen Zeng, Jianquan Yang, and Rui Yang.
\newblock {DRCT:} diffusion reconstruction contrastive training towards universal detection of diffusion generated images.
\newblock In \emph{Forty-first International Conference on Machine Learning, {ICML} 2024, Vienna, Austria, July 21-27, 2024}. OpenReview.net, 2024.

\bibitem[Chen et~al.(2025)Chen, Xi, Yan, Zhang, Wu, Xie, Chen, Xu, Guan, Yao, and Ding]{DDA}
Ruoxin Chen, Junwei Xi, Zhiyuan Yan, Ke{-}Yue Zhang, Shuang Wu, Jingyi Xie, Xu Chen, Lei Xu, Isabel Guan, Taiping Yao, and Shouhong Ding.
\newblock Dual data alignment makes {AI-Generated} image detector easier generalizable.
\newblock In \emph{Advances in Neural Information Processing Systems 38: Annual Conference on Neural Information Processing Systems 2025, NeurIPS 2025, San Diego, CA, USA, December 2-7, 2025 / Mexico City, Mexico, November 30 - December 5, 2025}, 2025.

\bibitem[Cui et~al.(2025)Cui, Li, Luo, Zhou, and Dong]{ForAda}
Xinjie Cui, Yuezun Li, Ao Luo, Jiaran Zhou, and Junyu Dong.
\newblock Forensics adapter: Adapting {CLIP} for generalizable face forgery detection.
\newblock In \emph{{IEEE/CVF} Conference on Computer Vision and Pattern Recognition, {CVPR} 2025, Nashville, TN, USA, June 11-15, 2025}, pages 19207--19217. Computer Vision Foundation / {IEEE}, 2025.

\bibitem[Deng et~al.(2018)Deng, Guo, et~al.]{InsightFace}
Jiankang Deng, Jia Guo, et~al.
\newblock {InsightFace}: {2D} and {3D} face analysis project.
\newblock \url{https://github.com/deepinsight/insightface}, 2018.
\newblock Model pack: \texttt{buffalo\_l}.

\bibitem[Dufour et~al.(2019)Dufour, Gully, Karlsson, Vorbyov, Leung, Childs, and Bregler]{DFD}
Nicholas Dufour, Andrew Gully, Per Karlsson, Alexey~Victor Vorbyov, Thomas Leung, Jeremiah Childs, and Christoph Bregler.
\newblock {Deepfakes Detection Dataset by Google \& Jigsaw}.
\newblock \url{https://research.google/blog/contributing-data-to-deepfake-detection-research/}, 2019.

\bibitem[Frankle et~al.(2020)Frankle, Dziugaite, Roy, and Carbin]{LMC}
Jonathan Frankle, Gintare~Karolina Dziugaite, Daniel Roy, and Michael Carbin.
\newblock Linear mode connectivity and the lottery ticket hypothesis.
\newblock In \emph{Proceedings of the 37th International Conference on Machine Learning, {ICML} 2020, 13-18 July 2020, Virtual Event}, pages 3259--3269. {PMLR}, 2020.

\bibitem[Guillaro et~al.(2025)Guillaro, Zingarini, Usman, Sud, Cozzolino, and Verdoliva]{B-Free}
Fabrizio Guillaro, Giada Zingarini, Ben Usman, Avneesh Sud, Davide Cozzolino, and Luisa Verdoliva.
\newblock A {Bias-Free} training paradigm for more general {AI-generated} image detection.
\newblock In \emph{{IEEE/CVF} Conference on Computer Vision and Pattern Recognition, {CVPR} 2025, Nashville, TN, USA, June 11-15, 2025}, pages 18685--18694. Computer Vision Foundation / {IEEE}, 2025.

\bibitem[Guo et~al.(2022)Guo, Deng, Lattas, and Zafeiriou]{SCRFD}
Jia Guo, Jiankang Deng, Alexandros Lattas, and Stefanos Zafeiriou.
\newblock Sample and computation redistribution for efficient face detection.
\newblock In \emph{The Tenth International Conference on Learning Representations, {ICLR} 2022, Virtual Event, April 25-29, 2022}. OpenReview.net, 2022.

\bibitem[HaCohen et~al.(2024)HaCohen, Chiprut, Brazowski, Shalem, Moshe, Richardson, Levin, Shiran, Zabari, Gordon, et~al.]{LTX-Video}
Yoav HaCohen, Nisan Chiprut, Benny Brazowski, Daniel Shalem, Dudu Moshe, Eitan Richardson, Eran Levin, Guy Shiran, Nir Zabari, Ori Gordon, et~al.
\newblock {LTX}-video: Realtime video latent diffusion.
\newblock \emph{arXiv preprint arXiv:2501.00103}, 2024.

\bibitem[HaCohen et~al.(2026)HaCohen, Brazowski, Chiprut, Bitterman, Kvochko, Berkowitz, Shalem, Lifschitz, Moshe, Porat, et~al.]{LTX-2}
Yoav HaCohen, Benny Brazowski, Nisan Chiprut, Yaki Bitterman, Andrew Kvochko, Avishai Berkowitz, Daniel Shalem, Daphna Lifschitz, Dudu Moshe, Eitan Porat, et~al.
\newblock {LTX-2}: Efficient joint audio-visual foundation model.
\newblock \emph{arXiv preprint arXiv:2601.03233}, 2026.

\bibitem[Han et~al.(2025)Han, Huang, Hua, and Chen]{DFD-FCG}
Yue{-}Hua Han, Tai{-}Ming Huang, Kai{-}Lung Hua, and Jun{-}Cheng Chen.
\newblock Towards more general video-based deepfake detection through facial component guided adaptation for foundation model.
\newblock In \emph{{IEEE/CVF} Conference on Computer Vision and Pattern Recognition, {CVPR} 2025, Nashville, TN, USA, June 11-15, 2025}, pages 22995--23005. Computer Vision Foundation / {IEEE}, 2025.

\bibitem[Hou et~al.(2024)Hou, Fu, Chen, Li, Zhang, and Zhao]{PolyGlotFake}
Yang Hou, Haitao Fu, Chunkai Chen, Zida Li, Haoyu Zhang, and Jianjun Zhao.
\newblock {PolyGlotFake}: A novel multilingual and multimodal deepfake dataset.
\newblock In \emph{International Conference on Pattern Recognition}, pages 180--193. Springer, 2024.

\bibitem[Hu et~al.(2022)Hu, Shen, Wallis, Allen{-}Zhu, Li, Wang, Wang, and Chen]{LoRA}
Edward~J. Hu, Yelong Shen, Phillip Wallis, Zeyuan Allen{-}Zhu, Yuanzhi Li, Shean Wang, Lu Wang, and Weizhu Chen.
\newblock {LoRA}: {Low-Rank} adaptation of large language models.
\newblock In \emph{The Tenth International Conference on Learning Representations, {ICLR} 2022, Virtual Event, April 25-29, 2022}. OpenReview.net, 2022.

\bibitem[Izmailov et~al.(2018)Izmailov, Podoprikhin, Garipov, Vetrov, and Wilson]{SWA}
Pavel Izmailov, Dmitrii Podoprikhin, Timur Garipov, Dmitry~P. Vetrov, and Andrew~Gordon Wilson.
\newblock Averaging weights leads to wider optima and better generalization.
\newblock In \emph{Proceedings of the Thirty-Fourth Conference on Uncertainty in Artificial Intelligence, {UAI} 2018, Monterey, California, USA, August 6-10, 2018}, pages 876--885. {AUAI} Press, 2018.

\bibitem[Jia et~al.(2022)Jia, Li, and Lyu]{DFDM}
Shan Jia, Xin Li, and Siwei Lyu.
\newblock Model attribution of face-swap deepfake videos.
\newblock In \emph{2022 IEEE International Conference on Image Processing (ICIP)}, pages 2356--2360. IEEE, 2022.

\bibitem[Kalajdzievski(2023)]{rsLoRA}
D Kalajdzievski.
\newblock A rank stabilization scaling factor for fine-tuning with lora.
\newblock \emph{arXiv preprint arXiv:2312.03732}, 2023.

\bibitem[Kang et~al.(2025)Kang, Jeong, Lee, Choi, Woo, and Han]{HiDF}
Chaewon Kang, Seoyoon Jeong, Jonghyun Lee, Daejin Choi, Simon~S. Woo, and Jinyoung Han.
\newblock {HiDF}: {A} {Human-Indistinguishable} deepfake dataset.
\newblock In \emph{Proceedings of the 31st {ACM} {SIGKDD} Conference on Knowledge Discovery and Data Mining, V.2, {KDD} 2025, Toronto ON, Canada, August 3-7, 2025}, pages 5527--5538. {ACM}, 2025.

\bibitem[Khalid et~al.(2021)Khalid, Tariq, Kim, and Woo]{FakeAVCeleb}
Hasam Khalid, Shahroz Tariq, Minha Kim, and Simon~S. Woo.
\newblock {FakeAVCeleb}: {A} novel {Audio-Video} multimodal deepfake dataset.
\newblock In \emph{Proceedings of the Neural Information Processing Systems Track on Datasets and Benchmarks 1, NeurIPS Datasets and Benchmarks 2021, December 2021, virtual}, 2021.

\bibitem[Kingma and Ba(2015)]{Adam}
Diederik~P. Kingma and Jimmy Ba.
\newblock Adam: {A} method for stochastic optimization.
\newblock In \emph{3rd International Conference on Learning Representations, {ICLR} 2015, San Diego, CA, USA, May 7-9, 2015, Conference Track Proceedings}, 2015.

\bibitem[Kong et~al.(2024)Kong, Tian, Zhang, Min, Dai, Zhou, Xiong, Li, Wu, Zhang, et~al.]{HunyuanVideo-1.0}
Weijie Kong, Qi Tian, Zijian Zhang, Rox Min, Zuozhuo Dai, Jin Zhou, Jiangfeng Xiong, Xin Li, Bo Wu, Jianwei Zhang, et~al.
\newblock {HunyuanVideo}: A systematic framework for large video generative models.
\newblock \emph{arXiv preprint arXiv:2412.03603}, 2024.

\bibitem[Li et~al.(2020{\natexlab{a}})Li, Bao, Yang, Chen, and Wen]{FSh}
Lingzhi Li, Jianmin Bao, Hao Yang, Dong Chen, and Fang Wen.
\newblock Advancing high fidelity identity swapping for forgery detection.
\newblock In \emph{2020 {IEEE/CVF} Conference on Computer Vision and Pattern Recognition, {CVPR} 2020, Seattle, WA, USA, June 13-19, 2020}, pages 5073--5082. {IEEE}, 2020{\natexlab{a}}.

\bibitem[Li et~al.(2020{\natexlab{b}})Li, Bao, Zhang, Yang, Chen, Wen, and Guo]{FaceXray}
Lingzhi Li, Jianmin Bao, Ting Zhang, Hao Yang, Dong Chen, Fang Wen, and Baining Guo.
\newblock Face {X}-ray for more general face forgery detection.
\newblock In \emph{2020 {IEEE/CVF} Conference on Computer Vision and Pattern Recognition, {CVPR} 2020, Seattle, WA, USA, June 13-19, 2020}, pages 5000--5009. {IEEE}, 2020{\natexlab{b}}.

\bibitem[Li et~al.(2026)Li, Huang, Wen, He, Li, Zhu, Duan, Chen, Fu, Dong, et~al.]{OmniFake}
Tianxiao Li, Zhenglin Huang, Haiquan Wen, Yiwei He, Xinze Li, Bingyu Zhu, Wuhui Duan, Congang Chen, Zeyu Fu, Yi Dong, et~al.
\newblock {OMNI-fake}: benchmarking unified multimodal social media deepfake detection.
\newblock In \emph{Proceedings of the IEEE/CVF conference on computer vision and pattern recognition}, pages 30299--30311, 2026.

\bibitem[Li et~al.(2018)Li, Grandvalet, and Davoine]{L2SP}
Xuhong Li, Yves Grandvalet, and Franck Davoine.
\newblock Explicit inductive bias for transfer learning with convolutional networks.
\newblock In \emph{Proceedings of the 35th International Conference on Machine Learning, {ICML} 2018, Stockholmsm{\"{a}}ssan, Stockholm, Sweden, July 10-15, 2018}, pages 2830--2839. {PMLR}, 2018.

\bibitem[Li et~al.(2020{\natexlab{c}})Li, Yang, Sun, Qi, and Lyu]{CDFv2}
Yuezun Li, Xin Yang, Pu Sun, Honggang Qi, and Siwei Lyu.
\newblock {Celeb-DF}: {A} large-scale challenging dataset for deepfake forensics.
\newblock In \emph{2020 {IEEE/CVF} Conference on Computer Vision and Pattern Recognition, {CVPR} 2020, Seattle, WA, USA, June 13-19, 2020}, pages 3204--3213. {IEEE}, 2020{\natexlab{c}}.

\bibitem[Li et~al.(2025)Li, Zhu, Cui, and Lyu]{CDFv3}
Y Li, D Zhu, X Cui, and S Lyu.
\newblock Celeb-df++: A large-scale challenging video deepfake benchmark for generalizable forensics.
\newblock \emph{arXiv preprint arXiv:2507.18015}, 2025.

\bibitem[Liu et~al.(2026)Liu, Wang, Hou, Ren, Wu, Ma, Pei, and He]{DigiFakeAV}
Jiaxin Liu, Jia Wang, Saihui Hou, Min Ren, Huijia Wu, Long Ma, Renwang Pei, and Zhaofeng He.
\newblock Beyond face swapping: A {Diffusion-Based} digital human benchmark for multimodal deepfake detection.
\newblock In \emph{ICASSP 2026 - 2026 IEEE International Conference on Acoustics, Speech and Signal Processing (ICASSP)}, pages 14087--14091, 2026.

\bibitem[Ojha et~al.(2023)Ojha, Li, and Lee]{UniFD}
Utkarsh Ojha, Yuheng Li, and Yong~Jae Lee.
\newblock Towards universal fake image detectors that generalize across generative models.
\newblock In \emph{{IEEE/CVF} Conference on Computer Vision and Pattern Recognition, {CVPR} 2023, Vancouver, BC, Canada, June 17-24, 2023}, pages 24480--24489. {IEEE}, 2023.

\bibitem[Oquab et~al.(2024)Oquab, Darcet, Moutakanni, Vo, Szafraniec, Khalidov, Fernandez, Haziza, Massa, El{-}Nouby, Assran, Ballas, Galuba, Howes, Huang, Li, Misra, Rabbat, Sharma, Synnaeve, Xu, J{\'{e}}gou, Mairal, Labatut, Joulin, and Bojanowski]{DINOv2}
Maxime Oquab, Timoth{\'{e}}e Darcet, Th{\'{e}}o Moutakanni, Huy~V. Vo, Marc Szafraniec, Vasil Khalidov, Pierre Fernandez, Daniel Haziza, Francisco Massa, Alaaeldin El{-}Nouby, Mido Assran, Nicolas Ballas, Wojciech Galuba, Russell Howes, Po{-}Yao Huang, Shang{-}Wen Li, Ishan Misra, Michael Rabbat, Vasu Sharma, Gabriel Synnaeve, Hu Xu, Herv{\'{e}} J{\'{e}}gou, Julien Mairal, Patrick Labatut, Armand Joulin, and Piotr Bojanowski.
\newblock {DINOv2}: Learning robust visual features without supervision.
\newblock \emph{Trans. Mach. Learn. Res.}, 2024, 2024.

\bibitem[Porgali et~al.(2023)Porgali, Albiero, Ryda, Ferrer, and Hazirbas]{CCv2}
Bilal Porgali, V{\'\i}tor Albiero, Jordan Ryda, Cristian~Canton Ferrer, and Caner Hazirbas.
\newblock The casual conversations {V2} dataset: A diverse, large benchmark for measuring fairness and robustness in audio/vision/speech models.
\newblock In \emph{2023 IEEE/CVF Conference on Computer Vision and Pattern Recognition Workshops (CVPRW)}, pages 10--17. IEEE, 2023.

\bibitem[Radford et~al.(2021)Radford, Kim, Hallacy, Ramesh, Goh, Agarwal, Sastry, Askell, Mishkin, Clark, Krueger, and Sutskever]{CLIP}
Alec Radford, Jong~Wook Kim, Chris Hallacy, Aditya Ramesh, Gabriel Goh, Sandhini Agarwal, Girish Sastry, Amanda Askell, Pamela Mishkin, Jack Clark, Gretchen Krueger, and Ilya Sutskever.
\newblock Learning transferable visual models from natural language supervision.
\newblock In \emph{Proceedings of the 38th International Conference on Machine Learning, {ICML} 2021, 18-24 July 2021, Virtual Event}, pages 8748--8763. {PMLR}, 2021.

\bibitem[Rajan et~al.(2025)Rajan, Ojha, Schloesser, and Lee]{AlignedForensics}
Anirudh~Sundara Rajan, Utkarsh Ojha, Jedidiah Schloesser, and Yong~Jae Lee.
\newblock Aligned datasets improve detection of latent {Diffusion-Generated} images.
\newblock In \emph{The Thirteenth International Conference on Learning Representations, {ICLR} 2025, Singapore, April 24-28, 2025}. OpenReview.net, 2025.

\bibitem[R{\"{o}}ssler et~al.(2019)R{\"{o}}ssler, Cozzolino, Verdoliva, Riess, Thies, and Nie{\ss}ner]{FF}
Andreas R{\"{o}}ssler, Davide Cozzolino, Luisa Verdoliva, Christian Riess, Justus Thies, and Matthias Nie{\ss}ner.
\newblock {FaceForensics++}: Learning to detect manipulated facial images.
\newblock In \emph{2019 {IEEE/CVF} International Conference on Computer Vision, {ICCV} 2019, Seoul, Korea (South), October 27 - November 2, 2019}, pages 1--11. {IEEE}, 2019.

\bibitem[Shi et~al.(2025)Shi, Li, Zuo, Yu, Lin, Hu, Zhou, Zhang, Wan, Xu, et~al.]{RedFace}
Junyu Shi, Minghui Li, Junguo Zuo, Zhifei Yu, Yipeng Lin, Shengshan Hu, Ziqi Zhou, Yechao Zhang, Wei Wan, Yinzhe Xu, et~al.
\newblock Towards real-world deepfake detection: A diverse in-the-wild dataset of forgery faces.
\newblock \emph{arXiv preprint arXiv:2510.08067}, 2025.

\bibitem[Shiohara and Yamasaki(2022)]{SBI}
Kaede Shiohara and Toshihiko Yamasaki.
\newblock Detecting deepfakes with self-blended images.
\newblock In \emph{Proceedings of the IEEE/CVF Conference on Computer Vision and Pattern Recognition}, pages 18720--18729, 2022.

\bibitem[Shutkin et~al.(2026)Shutkin, Parkhomenko, Kirillov, Chernyshev, Malakhov, Vasiliev, Trushkin, Kobenko, Chikovani, Ivanov, et~al.]{KVAE}
Andrey Shutkin, Denis Parkhomenko, Ivan Kirillov, Kirill Chernyshev, Kirill Malakhov, Ilia Vasiliev, Ilia Trushkin, Valeriya Kobenko, David Chikovani, Alexander Ivanov, et~al.
\newblock Kvae: Family of tokenizers for multimodal generative models.
\newblock \emph{arXiv preprint arXiv:2608.05798}, 2026.

\bibitem[Shykula et~al.(2026)Shykula, Yermakov, Samarskyi, Mishkin, Cech, and Mishchuk]{DF26}
Severyn Shykula, Andrii Yermakov, Ivan Samarskyi, Dmytro Mishkin, Jan Cech, and Anastasiia Mishchuk.
\newblock Df26: We cannot tell fake from real anymore.
\newblock \emph{arXiv preprint arXiv:2609.07369}, 2026.

\bibitem[Sim{\'{e}}oni et~al.(2026)Sim{\'{e}}oni, Vo, Seitzer, Baldassarre, Oquab, Jose, Khalidov, Szafraniec, Yi, Ramamonjisoa, Massa, Haziza, Wehrstedt, Wang, Darcet, Moutakanni, Sentana, Roberts, Vedaldi, Tolan, Brandt, Couprie, Mairal, J{\'{e}}gou, Labatut, and Bojanowski]{DINOv3}
Oriane Sim{\'{e}}oni, Huy~V. Vo, Maximilian Seitzer, Federico Baldassarre, Maxime Oquab, Cijo Jose, Vasil Khalidov, Marc Szafraniec, Seung~Eun Yi, Micha{\"{e}}l Ramamonjisoa, Francisco Massa, Daniel Haziza, Luca Wehrstedt, Jianyuan Wang, Timoth{\'{e}}e Darcet, Th{\'{e}}o Moutakanni, Leonel Sentana, Claire Roberts, Andrea Vedaldi, Jamie Tolan, John Brandt, Camille Couprie, Julien Mairal, Herv{\'{e}} J{\'{e}}gou, Patrick Labatut, and Piotr Bojanowski.
\newblock {DINOv3}.
\newblock \emph{Trans. Mach. Learn. Res.}, 2026, 2026.

\bibitem[Sun et~al.(2026)Sun, Yan, Zhang, Yao, and Ding]{DFD-HR}
Jiamu Sun, Zhiyuan Yan, Ke-Yue Zhang, Taiping Yao, and Shouhong Ding.
\newblock {DFD-HR}: Generalizable deepfake detection via hierarchical routing learning.
\newblock In \emph{Proceedings of the IEEE/CVF Conference on Computer Vision and Pattern Recognition}, pages 13984--13995, 2026.

\bibitem[Tan et~al.(2026{\natexlab{a}})Tan, jun lan, Shi, Tan, Yu, Zhu, Wang, Wan, and Lei]{MintVid}
Hao Tan, jun lan, Senyuan Shi, Zichang Tan, Zijian Yu, Huijia Zhu, Weiqiang Wang, Jun Wan, and Zhen Lei.
\newblock {VideoVeritas}: {AI}-generated video detection via perception pretext reinforcement learning.
\newblock In \emph{Forty-third International Conference on Machine Learning}, 2026{\natexlab{a}}.

\bibitem[Tan et~al.(2026{\natexlab{b}})Tan, Tan, Shi, Liu, Song, Zhu, Wang, Wan, Lei, et~al.]{HydraFake}
Hao Tan, Zichang Tan, Senyuan Shi, Ajian Liu, Chuanbiao Song, Huijia Zhu, Weiqiang Wang, Jun Wan, Zhen Lei, et~al.
\newblock Veritas: Generalizable deepfake detection via pattern-aware reasoning.
\newblock In \emph{International Conference on Learning Representations}, pages 66420--66477, 2026{\natexlab{b}}.

\bibitem[Tian et~al.(2024)Tian, Huang, and Kira]{L2SPAdam}
Junjiao Tian, Chengyue Huang, and Zsolt Kira.
\newblock Rethinking weight decay for robust {Fine-Tuning} of foundation models.
\newblock In \emph{Advances in Neural Information Processing Systems 37: Annual Conference on Neural Information Processing Systems 2024, NeurIPS 2024, Vancouver, BC, Canada, December 10 - 15, 2024}, 2024.

\bibitem[Wang et~al.(2025{\natexlab{a}})Wang, Lin, Wu, Liu, Ba, and Ren]{FSFM}
Gaojian Wang, Feng Lin, Tong Wu, Zhenguang Liu, Zhongjie Ba, and Kui Ren.
\newblock {FSFM:} {A} generalizable face security foundation model via {Self-Supervised} facial representation learning.
\newblock In \emph{{IEEE/CVF} Conference on Computer Vision and Pattern Recognition, {CVPR} 2025, Nashville, TN, USA, June 11-15, 2025}, pages 24364--24376. Computer Vision Foundation / {IEEE}, 2025{\natexlab{a}}.

\bibitem[Wang et~al.(2025{\natexlab{b}})Wang, Lin, Wu, Yan, and Ren]{FS-VFM}
Gaojian Wang, Feng Lin, Tong Wu, Zhisheng Yan, and Kui Ren.
\newblock Scalable face security vision foundation model for deepfake, diffusion, and spoofing detection.
\newblock \emph{arXiv preprint arXiv:2510.10663}, 2025{\natexlab{b}}.

\bibitem[Wang et~al.(2025{\natexlab{c}})Wang, Li, Wang, Chen, and Chen]{MiLoRA}
Hanqing Wang, Yixia Li, Shuo Wang, Guanhua Chen, and Yun Chen.
\newblock {M}i{L}o{RA}: Harnessing minor singular components for {Parameter-Efficient} {LLM} finetuning.
\newblock In \emph{Proceedings of the 2025 Conference of the Nations of the Americas Chapter of the Association for Computational Linguistics: Human Language Technologies (Volume 1: Long Papers)}, pages 4823--4836, Albuquerque, New Mexico, 2025{\natexlab{c}}. Association for Computational Linguistics.

\bibitem[Wortsman et~al.(2022)Wortsman, Ilharco, Gadre, Roelofs, Lopes, Morcos, Namkoong, Farhadi, Carmon, Kornblith, and Schmidt]{ModelSoups}
Mitchell Wortsman, Gabriel Ilharco, Samir~Yitzhak Gadre, Rebecca Roelofs, Raphael~Gontijo Lopes, Ari~S. Morcos, Hongseok Namkoong, Ali Farhadi, Yair Carmon, Simon Kornblith, and Ludwig Schmidt.
\newblock Model soups: averaging weights of multiple fine-tuned models improves accuracy without increasing inference time.
\newblock In \emph{International Conference on Machine Learning, {ICML} 2022, 17-23 July 2022, Baltimore, Maryland, {USA}}, pages 23965--23998. {PMLR}, 2022.

\bibitem[Xiong et~al.(2026)Xiong, Patel, Fan, Wadhwa, Selvam, Guo, Qi, Liu, and Sengupta]{TalkingHeadBench}
Xinqi Xiong, Prakrut Patel, Qingyuan Fan, Amisha Wadhwa, Sarathy Selvam, Xiao Guo, Luchao Qi, Xiaoming Liu, and Roni Sengupta.
\newblock {TalkingHeadBench}: {A} {Multi-Modal} benchmark {\&} analysis of {Talking-Head} {DeepFake} detection.
\newblock In \emph{{IEEE/CVF} Winter Conference on Applications of Computer Vision, {WACV} 2026, Tucson, AZ, USA, March 6-10, 2026}, pages 4139--4149. {IEEE}, 2026.

\bibitem[Xu et~al.(2024)Xu, Chen, Song, Han, Shan, and Jiang]{IDForge}
Junhao Xu, Jingjing Chen, Xue Song, Feng Han, Haijun Shan, and Yu-Gang Jiang.
\newblock Identity-driven multimedia forgery detection via reference assistance.
\newblock In \emph{Proceedings of the 32nd ACM International Conference on Multimedia}, pages 3887--3896, 2024.

\bibitem[Yan et~al.(2026)Yan, Li, Wang, Wang, He, and Fu]{DGS-Net}
Jiazhen Yan, Ziqiang Li, Fan Wang, Boyu Wang, Ziwen He, and Zhangjie Fu.
\newblock {DGS}-net: {Distillation-Guided} gradient surgery for {CLIP} {Fine-Tuning} in {AI}-generated image detection.
\newblock In \emph{Forty-third International Conference on Machine Learning}, 2026.

\bibitem[Yan et~al.(2023)Yan, Zhang, Yuan, Lyu, and Wu]{DeepfakeBench}
Zhiyuan Yan, Yong Zhang, Xinhang Yuan, Siwei Lyu, and Baoyuan Wu.
\newblock {DeepfakeBench}: {A} comprehensive benchmark of deepfake detection.
\newblock In \emph{Advances in Neural Information Processing Systems 36: Annual Conference on Neural Information Processing Systems 2023, NeurIPS 2023, New Orleans, LA, USA, December 10 - 16, 2023}, 2023.

\bibitem[Yan et~al.(2025)Yan, Wang, Jin, Zhang, Liu, Chen, Yao, Ding, Wu, and Yuan]{Effort}
Zhiyuan Yan, Jiangming Wang, Peng Jin, Ke{-}Yue Zhang, Chengchun Liu, Shen Chen, Taiping Yao, Shouhong Ding, Baoyuan Wu, and Li Yuan.
\newblock Orthogonal subspace decomposition for generalizable {AI-Generated} image detection.
\newblock In \emph{Forty-second International Conference on Machine Learning, {ICML} 2025, Vancouver, BC, Canada, July 13-19, 2025}. {PMLR} / OpenReview.net, 2025.

\bibitem[Yang et~al.(2026)Yang, Li, Kong, Diao, Gao, Shi, and Wang]{LTD}
Yawen Yang, Feng Li, Shuqi Kong, Yunfeng Diao, Xinjian Gao, Zenglin Shi, and Meng Wang.
\newblock Layer consistency matters: Elegant latent transition discrepancy for generalizable synthetic image detection.
\newblock In \emph{Proceedings of the IEEE/CVF Conference on Computer Vision and Pattern Recognition (CVPR)}, pages 38111--38121, 2026.

\bibitem[Yang et~al.(2025)Yang, Teng, Zheng, Ding, Huang, Xu, Yang, Hong, Zhang, Feng, Yin, Zhang, Wang, Cheng, Xu, Gu, Dong, and Tang]{CogVideoX}
Zhuoyi Yang, Jiayan Teng, Wendi Zheng, Ming Ding, Shiyu Huang, Jiazheng Xu, Yuanming Yang, Wenyi Hong, Xiaohan Zhang, Guanyu Feng, Da Yin, Yuxuan Zhang, Weihan Wang, Yean Cheng, Bin Xu, Xiaotao Gu, Yuxiao Dong, and Jie Tang.
\newblock {CogVideoX}: {Text-to-Video} diffusion models with an expert transformer.
\newblock In \emph{The Thirteenth International Conference on Learning Representations, {ICLR} 2025, Singapore, April 24-28, 2025}. OpenReview.net, 2025.

\bibitem[Yermakov et~al.(2026)Yermakov, Cech, Matas, and Fritz]{GenD}
Andrii Yermakov, Jan Cech, Jiri Matas, and Mario Fritz.
\newblock Deepfake detection that generalizes across benchmarks.
\newblock In \emph{{IEEE/CVF} Winter Conference on Applications of Computer Vision, {WACV} 2026, Tucson, AZ, USA, March 6-10, 2026}, pages 773--783. {IEEE}, 2026.

\bibitem[Zhao et~al.(2024)Zhao, Tu, Wei, Mei, and Xie]{LN-tuning}
Bingchen Zhao, Haoqin Tu, Chen Wei, Jieru Mei, and Cihang Xie.
\newblock Tuning layernorm in attention: Towards efficient multi-modal {LLM} finetuning.
\newblock In \emph{The Twelfth International Conference on Learning Representations, {ICLR} 2024, Vienna, Austria, May 7-11, 2024}, 2024.

\bibitem[Zhou et~al.(2021)Zhou, Wang, Liang, and Shen]{FFIW}
Tianfei Zhou, Wenguan Wang, Zhiyuan Liang, and Jianbing Shen.
\newblock Face forensics in the wild.
\newblock In \emph{{IEEE} Conference on Computer Vision and Pattern Recognition, {CVPR} 2021, virtual, June 19-25, 2021}, pages 5778--5788. Computer Vision Foundation / {IEEE}, 2021.

\bibitem[Zhu et~al.(2026)Zhu, Miao, Zhang, Deng, and Pang]{VLAForge}
Jiawen Zhu, Yunqi Miao, Xueyi Zhang, Jiankang Deng, and Guansong Pang.
\newblock Unleashing {Vision-Language} semantics for deepfake video detection.
\newblock In \emph{Proceedings of the IEEE/CVF Conference on Computer Vision and Pattern Recognition}, pages 42953--42963, 2026.

\end{thebibliography}


\begin{thebibliography}{13}
\providecommand{\natexlab}[1]{#1}
\providecommand{\url}[1]{\texttt{#1}}
\expandafter\ifx\csname urlstyle\endcsname\relax
  \providecommand{\doi}[1]{doi: #1}\else
  \providecommand{\doi}{doi: \begingroup \urlstyle{rm}\Url}\fi

\bibitem[Blattmann et~al.(2023)Blattmann, Dockhorn, Kulal, Mendelevitch, Kilian, Lorenz, Levi, English, Voleti, Letts, et~al.]{SVD}
Andreas Blattmann, Tim Dockhorn, Sumith Kulal, Daniel Mendelevitch, Maciej Kilian, Dominik Lorenz, Yam Levi, Zion English, Vikram Voleti, Adam Letts, et~al.
\newblock Stable video diffusion: Scaling latent video diffusion models to large datasets.
\newblock \emph{arXiv preprint arXiv:2311.15127}, 2023.

\bibitem[Gan et~al.(2025)Gan, Yang, Zhu, Xue, and Hoi]{OmniAvatar}
Qijun Gan, Ruizi Yang, Jianke Zhu, Shaofei Xue, and Steven Hoi.
\newblock {OmniAvatar}: Efficient audio-driven avatar video generation with adaptive body animation.
\newblock \emph{arXiv preprint arXiv:2506.18866}, 2025.

\bibitem[Guo et~al.(2024)Guo, Zhang, Liu, Zhong, Zhang, Wan, and Zhang]{LivePortrait}
Jianzhu Guo, Dingyun Zhang, Xiaoqiang Liu, Zhizhou Zhong, Yuan Zhang, Pengfei Wan, and Di Zhang.
\newblock {LivePortrait}: Efficient portrait animation with stitching and retargeting control.
\newblock \emph{arXiv preprint arXiv:2407.03168}, 2024.

\bibitem[Hong et~al.(2025)Hong, Xu, Zhou, Zhou, Li, Lin, Lu, and Xu]{ACTalker}
Fa{-}Ting Hong, Zunnan Xu, Zixiang Zhou, Jun Zhou, Xiu Li, Qin Lin, Qinglin Lu, and Dan Xu.
\newblock {Audio-Visual} controlled video diffusion with masked selective state spaces modeling for natural talking head generation.
\newblock In \emph{Proc. of ICCV}, pages 12549--12558. {IEEE}, 2025.

\bibitem[Ki et~al.(2025)Ki, Min, and Chae]{FLOAT}
Taekyung Ki, Dongchan Min, and Gyeongsu Chae.
\newblock {FLOAT:} generative motion latent flow matching for {Audio-Driven} talking portrait.
\newblock In \emph{Proc. of ICCV}, pages 14699--14710. {IEEE}, 2025.

\bibitem[Li et~al.(2025)Li, Zheng, Yang, Chen, and Yang]{Ditto}
Tianqi Li, Ruobing Zheng, Minghui Yang, Jingdong Chen, and Ming Yang.
\newblock Ditto: Motion-space diffusion for controllable realtime talking head synthesis.
\newblock In \emph{Proceedings of the 33rd ACM International Conference on Multimedia}, pages 9704--9713, 2025.

\bibitem[Liu et~al.(2025)Liu, Ma, Li, Chen, Liu, Li, Zhou, He, and Wu]{Phantom}
Lijie Liu, Tianxiang Ma, Bingchuan Li, Zhuowei Chen, Jiawei Liu, Gen Li, Siyu Zhou, Qian He, and Xinglong Wu.
\newblock Phantom: {Subject-Consistent} video generation via {Cross-Modal} alignment.
\newblock In \emph{Proc. of ICCV}, pages 14951--14961. {IEEE}, 2025.

\bibitem[Shapley(1953)]{Shapley}
Lloyd~S. Shapley.
\newblock A value for {n-Person} games.
\newblock In \emph{Contributions to the Theory of Games {II}}, pages 307--317. Princeton University Press, 1953.

\bibitem[Tan et~al.(2024)Tan, Ji, Bi, and Pan]{EDTalk}
Shuai Tan, Bin Ji, Mengxiao Bi, and Ye Pan.
\newblock {EDTalk}: Efficient disentanglement for emotional talking head synthesis.
\newblock In \emph{Computer Vision - {ECCV} 2024 - 18th European Conference, Milan, Italy, September 29-October 4, 2024, Proceedings, Part {VI}}, pages 398--416. Springer, 2024.

\bibitem[Wan et~al.(2025)Wan, Wang, Ai, Wen, Mao, Xie, Chen, Yu, Zhao, Yang, et~al.]{WAN-2.1}
Team Wan, Ang Wang, Baole Ai, Bin Wen, Chaojie Mao, Chen-Wei Xie, Di Chen, Feiwu Yu, Haiming Zhao, Jianxiao Yang, et~al.
\newblock Wan: Open and advanced large-scale video generative models.
\newblock \emph{arXiv preprint arXiv:2503.20314}, 2025.

\bibitem[Wang et~al.(2025)Wang, Wang, Jiang, Fan, Zhang, Qi, Zhao, and Xu]{FantasyTalking}
Mengchao Wang, Qiang Wang, Fan Jiang, Yaqi Fan, Yunpeng Zhang, Yonggang Qi, Kun Zhao, and Mu Xu.
\newblock Fantasytalking: Realistic talking portrait generation via coherent motion synthesis.
\newblock In \emph{Proceedings of the 33rd ACM International Conference on Multimedia}, pages 9891--9900, 2025.

\bibitem[Wang et~al.(2022)Wang, Yang, Br{\'{e}}mond, and Dantcheva]{LIA}
Yaohui Wang, Di Yang, Fran{\c{c}}ois Br{\'{e}}mond, and Antitza Dantcheva.
\newblock Latent image animator: Learning to animate images via latent space navigation.
\newblock In \emph{Proc. of ICLR}. OpenReview.net, 2022.

\bibitem[Zhang et~al.(2023)Zhang, Singh, Ghassemi, and Joshi]{ShiftAttribution}
Haoran Zhang, Harvineet Singh, Marzyeh Ghassemi, and Shalmali Joshi.
\newblock "why did the model fail?": Attributing model performance changes to distribution shifts.
\newblock In \emph{Proc. of ICML}, pages 41550--41578. {PMLR}, 2023.

\end{thebibliography}
}

\onecolumn 
\setcounter{page}{1}
\suppressfloats[t]
\vspace*{0pt}
{\centering
\LARGE \textbf{Supplementary Material}\par
}
\vspace{1.5em}
\renewcommand{\thefigure}{S\arabic{figure}}
\renewcommand{\thetable}{S\arabic{table}}
\renewcommand{\theequation}{S\arabic{equation}}
\renewcommand{\thesection}{S\arabic{section}}
\renewcommand{\theHfigure}{S\arabic{figure}}
\renewcommand{\theHtable}{S\arabic{table}}
\renewcommand{\theHequation}{S\arabic{equation}}
\renewcommand{\theHsection}{S\arabic{section}}

\setcounter{section}{0}
\setcounter{table}{0}
\setcounter{equation}{0}
\setcounter{figure}{0}

\section{Supplementary material overview}

This supplementary material gives the details needed to reproduce \method and the results per dataset behind the tables of the main paper.

\titlecontents{section}[0em]{\addvspace{.6em}\bfseries}{\thecontentslabel\hspace{1em}}{}{\hfill\contentspage}
\titlecontents{subsection}[1.5em]{}{\thecontentslabel\hspace{1em}}{}{\titlerule*[.8em]{.}\contentspage}
\startcontents[supp]
\printcontents[supp]{}{1}{\setcounter{tocdepth}{2}}

\clearpage
\section{Implementation details}
\label{sec:supp-impl}

\Cref{tab:supp-settings} summarizes the training and evaluation settings.

\begin{table}[h]
  \centering
  \small
  \caption{\label{tab:supp-settings} Training and evaluation settings.}
  \begin{tabular}{@{}lp{0.62\linewidth}@{}}
    \toprule
    Setting & Value \\
    \midrule
    Backbone & \PEcoreL, $224$\,px, linear classifier on the image embedding \\
    Adapter & LoRA, $r = 32$, $\alpha = 32$, on $W_q$, $W_k$, $W_v$ and $W_o$ of all 24 blocks \\
    Trainable parameters & 6.29\,M of 322.1\,M \\
    Reals & FF++ (5,739 frames) + CCv2 (1,913 frames) \\
    Fakes per epoch & $\tfrac13$ blob twins, $\tfrac13$ reconstruction twins, $\tfrac13$ FF++ fakes (4 methods) \\
    Epoch / batch & 34,434 pairs / 64 pairs (128 images) \\
    Optimizer & Adam, LR $3\cdot10^{-4}$ constant, 1-epoch warm-up \\
    Regularization & L2-SP, $\lambda = 1$, no weight decay \\
    Schedule & 10 epochs, SWA over epochs 6--10, no validation \\
    Final model & uniform soup of 3 seeds \\
    Test & $\le 32$ frames / video, mean fake probability, video AUROC \\
    \bottomrule
  \end{tabular}
\end{table}

\subsection{Reconstruction twins}
\label{sec:supp-partial}

Let $f \sim \mathcal U(0, 0.9)$ be the real share of the mask $M_\mathrm{r}$ in \cref{eq:partial}. The grid mask sets $\operatorname{round}(32 \times 32 \times f)$ random cells to real. The noise mask is Gaussian-smoothed noise, with $\sigma \sim \mathcal U(0.03, 0.09)$ of the side, thresholded at its $(1 - f)$ quantile.

\subsection{Blob twins}
\label{sec:supp-blob}

For a crop of width $w$ and height $h$, the radius of the blob in \cref{eq:blob} is
\begin{equation}
  \rho = \sqrt{a\,w h / \pi}, \qquad a \sim \mathcal U(0.08, 0.5),
\end{equation}
so that the disk of radius $\rho$ covers a share $a$ of the crop. The contrast change is up to $\pm 0.18$, and the hue shift is up to $\pm 0.03$. The shift is an integer pixel offset, drawn uniformly on each axis. $B$ is never shifted.

\subsection{Augmentation}
\label{sec:supp-augmentation}

The horizontal flip has $p = 0.5$. After it, the affine transform, blur, jitter and sharpening follow in a random order, drawn per pair. The affine transform rotates by up to $\pm 10^\circ$, translates by up to 10\% and scales by $\mathcal U(0.9, 1.1)$. It uses bicubic, bilinear or nearest interpolation with probabilities $0.45$, $0.45$ and $0.1$. The Gaussian blur has $p = 0.1$, kernel 7 and $\sigma \sim \mathcal U(0.1, 2)$. The brightness and contrast jitter is $\pm 0.1$. Sharpening is an unsharp mask $I + a\,(I - G_\sigma * I)$ with $a \sim \mathcal U(0.5, 3)$ and $\sigma \sim \mathcal U(0.5, 1.5)$\,px. The JPEG quality is drawn uniformly from 40 to 100.

\subsection{Model}
\label{sec:supp-model}

In \PEcoreL, the query, key and value share one fused projection. LoRA (\cref{eq:lora}) adapts its three parts, $W_q$, $W_k$ and $W_v$, separately, and the output projection $W_o$. The trainable parameters, 6,293,506 of 322,068,482, include the classifier.

\subsection{Training}
\label{sec:supp-training}

Training uses bf16 mixed precision. Before stochastic weight averaging, each LoRA update is folded into its dense weight (\cref{eq:lora}). The three runs of the soup use seeds 1, 2 and 3. The seed changes the classifier and LoRA initialization, the pair order, the fake parameters, and the augmentation draws. The soup averages the folded backbone weights and the classifier.

After each Adam step, L2-SP~\cite{L2SP,L2SPAdam} pulls the parameters back toward their values at the start of training:
\begin{equation}
  \theta_t \leftarrow \tilde{\theta}_t - \lambda \eta (\tilde{\theta}_t - \theta_0),
  \label{eq:l2sp}
\end{equation}
where $\tilde{\theta}_t$ are the parameters after the Adam step $t$, $\theta_0$ are the parameters at the start of training, $\eta$ is the learning rate, and $\lambda$ is the regularization strength. We set $\lambda = 1$ and did not tune it. Without L2-SP and with no other regularization, the mean AUROC of single models drops by 0.7 points (\cref{tab:ablations}). We keep it in all experiments.

\clearpage
\section{EER and mAP}
\label{sec:supp-eer-map}

\Cref{tab:supp-sota-eer,tab:supp-sota-map} repeat the comparison of \cref{tab:sota} with the equal error rate (EER) and the mean average precision (mAP). The EER is the error rate at the threshold where real and fake videos are misclassified at the same rate. The mAP is the mean over the two classes of the average precision (AP), computed once with fake and once with real as the positive class. So it is not directly comparable with an AP that takes only fake as the positive class.
\method has the lowest EER on 17 of the 19 datasets, 9.0\% on average against 19.1\% for GenD-PE~\cite{GenD}, and the highest mAP on 16, 93.6\% on average against 83.0\% for DFD-HR~\cite{DFD-HR}. As for the AUROC, the exceptions are \texttt{CDFv3} and \texttt{HF}, and for the mAP also \texttt{DFD}.

\begin{table}[h]
\centering
\caption{\label{tab:supp-sota-eer}
    \textbf{Equal error rate.} Video-level EER (\%) of the methods of \cref{tab:sota} on the 19 datasets, lower is better. The rows are in the order of \cref{tab:sota}. Above the dataset names are their release years. Avg is the macro-average. \textsuperscript{\dag}A temporal model, which takes several frames as input. The best value in each column is in bold.
}
\tabcolsep=4pt
\resizebox{\textwidth}{!}{
\begin{tabular}{l|cccccccccc|cccc|ccccc|c}
\toprule
 & \multicolumn{10}{c|}{\textbf{\texttt{Blending}}} & \multicolumn{4}{c|}{\textbf{\texttt{Blending + EFS}}} & \multicolumn{5}{c|}{\textbf{\texttt{EFS}}} &  \\
 & 2019 & 2020 & 2020 & 2021 & 2021 & 2022 & 2024 & 2024 & 2024 & 2025 & 2025 & 2025 & 2025 & 2026 & 2026 & 2026 & 2026 & 2026 & 2026 &  \\
 & \texttt{DFD} & \texttt{FSh} & \texttt{CDFv2} & \texttt{FFIW} & \texttt{FAVC} & \texttt{DFDM} & \texttt{PGF} & \texttt{IDF} & \texttt{DSv1} & \texttt{HiDF} & \texttt{CDFv3} & \texttt{DSv2} & \texttt{RF} & \texttt{HF} & \texttt{THB} & \texttt{OF} & \texttt{DFAV} & \texttt{MV} & \texttt{DF26} & Avg \\
\midrule
\multicolumn{21}{l}{\textit{AIGC detectors}} \\[2pt]
Effort-SDv1.4~\cite{Effort} \C{Effort-sdv14-D26CVPR} & 49.6 & 45.0 & 49.4 & 46.9 & 46.0 & 46.8 & 59.5 & 40.5 & 64.2 & 46.5 & 39.2 & 49.3 & 28.2 & 31.7 & 26.4 & 29.3 & 50.4 & 57.7 & 40.8 & 44.6 \\
DGS-Net~\cite{DGS-Net} \C{DGS-Net-D26CVPR} & 44.1 & 32.1 & 46.1 & 30.4 & 53.6 & 31.6 & 50.7 & 57.9 & 55.4 & 51.6 & 33.7 & 57.2 & 43.0 & 34.4 & 37.9 & 49.8 & 36.3 & 35.3 & 32.1 & 42.8 \\
LTD-UFD~\cite{LTD} \C{LTD-UFD-D26CVPR} & 24.4 & 27.9 & 45.9 & 31.6 & 41.2 & 49.9 & 42.4 & 51.0 & 50.6 & 38.7 & 40.3 & 60.2 & 40.2 & 48.1 & 28.7 & 63.8 & 29.2 & 48.9 & 18.7 & 41.1 \\
AlignedFor.~\cite{AlignedForensics} \C{AlignedForensics-Sync-D26CVPR} & 45.5 & 38.6 & 40.5 & 27.4 & 12.4 & 39.0 & 45.6 & 13.9 & 58.0 & 51.7 & 64.7 & 31.3 & 36.4 & 15.7 & 12.3 & 45.6 & 24.3 & 58.9 & 37.4 & 36.8 \\
DDA~\cite{DDA} \C{DDA-D26CVPR} & 40.8 & 48.6 & 44.7 & 46.7 & 32.2 & 49.9 & 42.7 & 35.0 & 35.8 & 37.2 & 42.7 & 25.1 & 32.1 & 8.6 & 23.5 & 62.5 & 14.6 & 32.9 & 33.2 & 36.2 \\
B-Free~\cite{B-Free} \C{B-Free-D26CVPR} & 35.3 & 35.7 & 44.9 & 44.5 & 27.4 & 32.5 & 27.8 & 31.7 & 26.5 & 35.7 & 44.7 & 26.0 & 30.5 & \textbf{5.8} & 21.6 & 53.4 & 15.7 & 32.0 & 37.0 & 32.0 \\
\midrule
\multicolumn{21}{l}{\textit{Facial deepfake detectors}} \\[2pt]
SBI~\cite{SBI} \C{SBI-FFraw-D26CVPR} & 17.6 & 25.7 & 21.5 & 18.4 & 5.4 & 1.6 & 31.1 & 7.3 & 41.9 & 25.6 & 43.8 & 44.2 & 41.3 & 40.8 & 33.2 & 31.6 & 43.9 & 54.7 & 43.6 & 30.2 \\
FSFM~\cite{FSFM} \C{FSFM-D26CVPR} & 18.5 & 34.3 & 15.9 & 28.7 & 18.6 & 8.9 & 22.6 & 25.1 & 23.9 & 15.8 & 27.5 & 37.1 & 38.2 & 33.8 & 52.7 & 51.3 & 32.1 & 40.7 & 48.2 & 30.2 \\
ForAda~\cite{ForAda} \C{ForAda-D26CVPR} & 6.6 & 26.4 & 12.4 & 14.1 & 15.2 & 8.0 & 21.6 & 18.5 & 24.6 & 16.1 & 32.5 & 30.3 & 35.5 & 44.9 & 42.8 & 39.9 & 33.0 & 38.9 & 46.1 & 26.7 \\
Effort~\cite{Effort} \C{Effort-D26CVPR} & 8.2 & 16.4 & 14.1 & 11.9 & 16.2 & 6.7 & 22.0 & 10.9 & 24.7 & 17.2 & 29.8 & 39.2 & 39.1 & 41.4 & 35.3 & 31.1 & 33.9 & 46.4 & 55.7 & 26.3 \\
DFD-FCG\textsuperscript{\dag}~\cite{DFD-FCG} \C{DFD-FCG-x1.4-D26CVPR} & 13.2 & 15.7 & 14.1 & 13.6 & 17.7 & 3.1 & 28.9 & 14.6 & 24.9 & 12.2 & \textbf{14.6} & 29.3 & 39.3 & 38.4 & 32.2 & 31.1 & 39.7 & 38.9 & 47.4 & 24.7 \\
VLAForge~\cite{VLAForge} \C{VLAForge-D26CVPR} & 6.8 & 15.0 & 12.9 & 8.5 & 13.5 & 3.9 & 18.6 & 8.1 & 22.8 & 13.4 & 35.0 & 34.2 & 35.2 & 44.2 & 28.4 & 34.0 & 31.6 & 40.2 & 47.2 & 23.9 \\
FS-VFM~\cite{FS-VFM} \C{FS-VFM-D26CVPR} & 7.4 & 27.9 & 12.9 & 16.7 & 8.2 & 5.5 & 17.9 & 14.2 & 15.1 & 7.8 & 23.0 & 27.4 & 33.2 & 29.8 & 42.4 & 41.1 & 26.0 & 43.8 & 47.2 & 23.5 \\
DFD-HR~\cite{DFD-HR} \C{DFD-HR-D26CVPR} & 5.8 & 15.0 & 14.4 & 6.4 & 5.6 & 2.5 & 11.9 & 5.5 & 17.1 & 9.6 & 27.0 & 26.5 & 34.0 & 44.5 & 27.8 & 33.2 & 21.7 & 26.5 & 40.7 & 19.8 \\
GenD-PE~\cite{GenD} \C{GenD-PE-HF-D26CVPR} & 6.1 & 20.7 & 10.7 & 10.8 & 9.1 & 6.0 & 12.4 & 7.4 & 19.2 & 13.7 & 16.9 & 29.2 & 28.4 & 29.6 & 23.8 & 33.9 & 22.0 & 26.6 & 35.9 & 19.1 \\
\midrule
\textbf{\method} \C{eCVPR27-M1-rsLoRA-r32-soup-D26CVPR} & \textbf{4.5} & \textbf{10.7} & \textbf{8.4} & \textbf{4.6} & \textbf{0.6} & \textbf{1.2} & \textbf{9.0} & \textbf{2.8} & \textbf{14.3} & \textbf{5.3} & 16.3 & \textbf{16.8} & \textbf{14.1} & 7.7 & \textbf{5.3} & \textbf{12.5} & \textbf{13.6} & \textbf{12.6} & \textbf{10.8} & \textbf{9.0} \\
\bottomrule
\end{tabular}
} %
\end{table}

\begin{table}[h]
\centering
\caption{\label{tab:supp-sota-map}
    \textbf{Mean average precision.} Video-level mAP (\%) of the methods of \cref{tab:sota} on the 19 datasets, higher is better. The mAP is the mean over the two classes: the average precision with fake as the positive class and with real as the positive class. The rows are in the order of \cref{tab:sota}. Above the dataset names are their release years. Avg is the macro-average over the datasets. \textsuperscript{\dag}A temporal model, which takes several frames as input. The best value in each column is in bold.
}
\tabcolsep=4pt
\resizebox{\textwidth}{!}{
\begin{tabular}{l|cccccccccc|cccc|ccccc|c}
\toprule
 & \multicolumn{10}{c|}{\textbf{\texttt{Blending}}} & \multicolumn{4}{c|}{\textbf{\texttt{Blending + EFS}}} & \multicolumn{5}{c|}{\textbf{\texttt{EFS}}} &  \\
 & 2019 & 2020 & 2020 & 2021 & 2021 & 2022 & 2024 & 2024 & 2024 & 2025 & 2025 & 2025 & 2025 & 2026 & 2026 & 2026 & 2026 & 2026 & 2026 &  \\
 & \texttt{DFD} & \texttt{FSh} & \texttt{CDFv2} & \texttt{FFIW} & \texttt{FAVC} & \texttt{DFDM} & \texttt{PGF} & \texttt{IDF} & \texttt{DSv1} & \texttt{HiDF} & \texttt{CDFv3} & \texttt{DSv2} & \texttt{RF} & \texttt{HF} & \texttt{THB} & \texttt{OF} & \texttt{DFAV} & \texttt{MV} & \texttt{DF26} & Avg \\
\midrule
\multicolumn{21}{l}{\textit{AIGC detectors}} \\[2pt]
Effort-SDv1.4~\cite{Effort} \C{Effort-sdv14-D26CVPR} & 50.9 & 58.2 & 49.6 & 56.3 & 50.4 & 52.7 & 48.1 & 58.1 & 39.7 & 54.3 & 51.5 & 51.2 & 79.2 & 75.2 & 72.3 & 76.0 & 48.7 & 44.2 & 55.0 & 56.4 \\
DGS-Net~\cite{DGS-Net} \C{DGS-Net-D26CVPR} & 53.9 & 69.2 & 54.2 & 70.0 & 49.6 & 66.6 & 49.9 & 46.4 & 44.4 & 48.4 & 52.3 & 46.0 & 58.4 & 62.0 & 60.4 & 49.2 & 63.4 & 64.8 & 60.9 & 56.3 \\
LTD-UFD~\cite{LTD} \C{LTD-UFD-D26CVPR} & 68.8 & 76.9 & 55.9 & 69.7 & 51.0 & 51.5 & 52.4 & 50.9 & 48.5 & 62.0 & 52.0 & 45.6 & 67.0 & 52.5 & 68.4 & 38.4 & 66.3 & 55.0 & 68.9 & 58.0 \\
AlignedFor.~\cite{AlignedForensics} \C{AlignedForensics-Sync-D26CVPR} & 51.7 & 61.7 & 57.6 & 75.8 & 64.0 & 58.8 & 52.1 & 84.5 & 47.9 & 48.7 & 48.4 & 74.6 & 69.5 & 89.5 & 91.5 & 57.3 & 82.1 & 47.6 & 55.8 & 64.2 \\
DDA~\cite{DDA} \C{DDA-D26CVPR} & 60.4 & 54.5 & 56.6 & 58.7 & 53.1 & 52.3 & 54.4 & 57.1 & 72.3 & 66.0 & 51.6 & 84.6 & 75.8 & 97.0 & 79.9 & 45.2 & 92.2 & 62.1 & 64.0 & 65.2 \\
B-Free~\cite{B-Free} \C{B-Free-D26CVPR} & 57.3 & 68.2 & 57.1 & 57.8 & 54.4 & 68.5 & 59.0 & 60.6 & 82.6 & 68.3 & 51.6 & 83.2 & 77.7 & \textbf{98.4} & 79.5 & 52.9 & 88.7 & 68.8 & 56.8 & 68.0 \\
\midrule
\multicolumn{21}{l}{\textit{Facial deepfake detectors}} \\[2pt]
SBI~\cite{SBI} \C{SBI-FFraw-D26CVPR} & 77.2 & 77.1 & 86.5 & 90.9 & 89.3 & 99.6 & 61.3 & 96.1 & 62.7 & 81.6 & 51.0 & 58.9 & 61.0 & 58.4 & 65.1 & 75.5 & 55.1 & 46.4 & 53.0 & 70.9 \\
FSFM~\cite{FSFM} \C{FSFM-D26CVPR} & 80.1 & 71.7 & 90.0 & 78.3 & 64.7 & 95.4 & 71.1 & 75.2 & 82.6 & 91.8 & 56.1 & 68.2 & 64.5 & 70.5 & 48.9 & 50.8 & 69.4 & 58.4 & 51.3 & 70.5 \\
ForAda~\cite{ForAda} \C{ForAda-D26CVPR} & 92.5 & 81.1 & 95.2 & 92.1 & 65.9 & 94.7 & 66.4 & 85.6 & 81.7 & 91.4 & 53.6 & 71.3 & 68.5 & 56.4 & 56.7 & 64.5 & 68.0 & 60.0 & 52.1 & 73.6 \\
Effort~\cite{Effort} \C{Effort-D26CVPR} & 86.0 & 90.5 & 92.1 & 95.0 & 61.7 & 96.4 & 69.2 & 93.1 & 81.7 & 90.7 & 54.2 & 63.9 & 63.2 & 61.8 & 68.0 & 73.1 & 67.3 & 54.8 & 48.9 & 74.3 \\
DFD-FCG\textsuperscript{\dag}~\cite{DFD-FCG} \C{DFD-FCG-x1.4-D26CVPR} & 84.2 & 92.1 & 93.7 & 93.9 & 61.6 & 99.2 & 63.3 & 85.8 & 82.7 & 94.5 & \textbf{76.0} & 78.2 & 63.5 & 63.6 & 69.6 & 75.1 & 60.7 & 61.5 & 50.9 & 76.3 \\
VLAForge~\cite{VLAForge} \C{VLAForge-D26CVPR} & 93.9 & 93.1 & 94.8 & 96.7 & 68.2 & 99.0 & 72.2 & 95.4 & 87.0 & 93.9 & 53.7 & 67.9 & 67.7 & 57.0 & 70.3 & 69.2 & 69.4 & 57.7 & 51.5 & 76.8 \\
FS-VFM~\cite{FS-VFM} \C{FS-VFM-D26CVPR} & 91.5 & 81.5 & 94.3 & 92.2 & 81.5 & 96.8 & 82.1 & 90.2 & 92.2 & 97.3 & 58.6 & 78.8 & 74.7 & 77.1 & 60.9 & 69.5 & 76.0 & 56.1 & 53.5 & 79.2 \\
DFD-HR~\cite{DFD-HR} \C{DFD-HR-D26CVPR} & \textbf{95.7} & 93.2 & 94.0 & 97.9 & 91.8 & 99.4 & 87.5 & 97.9 & 90.8 & 96.3 & 55.4 & 83.4 & 66.8 & 59.8 & 77.3 & 74.9 & 83.3 & 76.9 & 54.9 & 83.0 \\
GenD-PE~\cite{GenD} \C{GenD-PE-HF-D26CVPR} & 94.3 & 88.3 & 95.7 & 95.2 & 83.0 & 97.7 & 85.4 & 96.6 & 89.7 & 93.7 & 59.7 & 78.9 & 76.4 & 71.8 & 79.4 & 71.3 & 81.4 & 73.5 & 60.1 & 82.7 \\
\midrule
\textbf{\method} \C{eCVPR27-M1-rsLoRA-r32-soup-D26CVPR} & 94.8 & \textbf{97.0} & \textbf{96.5} & \textbf{98.9} & \textbf{98.4} & \textbf{99.7} & \textbf{90.2} & \textbf{99.2} & \textbf{93.6} & \textbf{98.7} & 61.3 & \textbf{92.4} & \textbf{91.7} & 95.4 & \textbf{97.8} & \textbf{95.0} & \textbf{92.5} & \textbf{93.8} & \textbf{90.7} & \textbf{93.6} \\
\bottomrule
\end{tabular}
} %
\end{table}

\clearpage

\section{Ablation of training fakes}
\label{sec:supp-ablation-fakes}

In the full training set, FF++ fake pairs take a third of the 34,434 pairs of an epoch. The other two thirds are made from the real frames of \cref{sec:real}, and each becomes a reconstruction twin or a blob twin with equal probability (\cref{sec:pairs}). The number of pairs per epoch is the same in every ablation. When reconstruction twins or blob twins are removed, the other kind of twin takes all of their pairs. When FF++ fakes are removed, the twins fill the entire epoch. FF++ fakes are made from FF++ reals, while twins are made from both FF++ and CCv2 reals. So when only FF++ fakes are left, the CCv2 reals have no fakes to pair with and are dropped as well.

\Cref{tab:supp-ablation-fakes} gives the per-dataset AUROC of the training fakes ablation in \cref{tab:ablations}.

\begin{table}[h]
\centering
\small
\caption{\label{tab:supp-ablation-fakes}
    \textbf{Per-dataset ablation of training fakes.} Video-level AUROC (\%) of single models, mean $\pm$ standard deviation over three seeds. Columns follow the training fakes rows of \cref{tab:ablations}. FF++, Rec and Blob are FF++ fakes, reconstruction twins and blob twins. The best value in each row is in bold.
}
\begin{tabular}{@{}l|ccccccc@{}}
\toprule
FF++ & \checkmark & $-$ & \checkmark & \checkmark & $-$ & \checkmark & $-$ \\
Rec & \checkmark & \checkmark & \checkmark & $-$ & $-$ & $-$ & \checkmark \\
Blob & \checkmark & \checkmark & $-$ & \checkmark & \checkmark & $-$ & $-$ \\
\midrule
\multicolumn{8}{@{}l}{\textbf{\texttt{Blending}}} \\[2pt]
\texttt{DFD} & 98.7$\pm$0.1 & 96.8$\pm$0.7 & 90.7$\pm$1.6 & \textbf{99.5$\pm$0.1} & 99.1$\pm$0.1 & 97.9$\pm$0.2 & 58.6$\pm$0.7 \\
\texttt{FSh} & 96.6$\pm$0.3 & 93.5$\pm$2.2 & 93.2$\pm$0.5 & \textbf{96.8$\pm$0.4} & 94.1$\pm$0.6 & 95.9$\pm$0.8 & 72.8$\pm$0.9 \\
\texttt{CDFv2} & 97.0$\pm$0.2 & 96.1$\pm$0.2 & 88.4$\pm$0.3 & \textbf{97.3$\pm$0.3} & 96.7$\pm$0.3 & 90.2$\pm$0.8 & 68.0$\pm$1.7 \\
\texttt{FFIW} & 98.6$\pm$0.4 & 95.9$\pm$1.0 & 93.0$\pm$0.4 & \textbf{99.6$\pm$0.0} & 99.5$\pm$0.1 & 98.4$\pm$0.6 & 67.2$\pm$1.8 \\
\texttt{FAVC} & \textbf{99.9$\pm$0.0} & \textbf{99.9$\pm$0.0} & 99.4$\pm$0.1 & \textbf{99.9$\pm$0.0} & \textbf{99.9$\pm$0.0} & 99.1$\pm$0.2 & 94.8$\pm$0.8 \\
\texttt{DFDM} & 99.8$\pm$0.1 & 99.7$\pm$0.1 & 93.2$\pm$2.1 & \textbf{100.0$\pm$0.0} & \textbf{100.0$\pm$0.0} & 99.8$\pm$0.1 & 0.2$\pm$0.1 \\
\texttt{PGF} & 96.8$\pm$0.7 & 94.0$\pm$0.9 & 89.4$\pm$0.5 & \textbf{98.1$\pm$0.3} & 96.0$\pm$0.2 & 96.4$\pm$0.3 & 70.4$\pm$0.6 \\
\texttt{IDF} & 99.1$\pm$0.1 & 98.8$\pm$0.1 & 97.2$\pm$0.1 & \textbf{99.8$\pm$0.1} & 99.6$\pm$0.0 & 99.7$\pm$0.1 & 73.1$\pm$4.6 \\
\texttt{DSv1} & \textbf{93.7$\pm$0.4} & 89.8$\pm$1.2 & 85.9$\pm$2.8 & 93.0$\pm$1.2 & 88.2$\pm$0.5 & 90.5$\pm$0.5 & 72.1$\pm$3.5 \\
\texttt{HiDF} & 98.4$\pm$0.1 & 96.2$\pm$0.3 & 90.1$\pm$0.6 & \textbf{98.9$\pm$0.1} & 98.7$\pm$0.2 & 97.6$\pm$0.3 & 71.9$\pm$2.1 \\
\midrule
\multicolumn{8}{@{}l}{\textbf{\texttt{B+EFS}}} \\[2pt]
\texttt{CDFv3} & \textbf{92.0$\pm$1.3} & 85.0$\pm$3.0 & 88.9$\pm$1.1 & 82.0$\pm$0.6 & 73.7$\pm$0.9 & 78.1$\pm$0.5 & 66.0$\pm$2.2 \\
\texttt{DSv2} & 92.4$\pm$0.7 & 89.1$\pm$0.3 & \textbf{93.1$\pm$1.1} & 75.1$\pm$0.7 & 69.6$\pm$0.5 & 74.9$\pm$0.9 & 86.1$\pm$1.5 \\
\texttt{RF} & 90.9$\pm$1.1 & 89.6$\pm$1.9 & \textbf{92.1$\pm$0.2} & 81.1$\pm$1.1 & 76.7$\pm$3.7 & 76.3$\pm$1.1 & 90.1$\pm$0.3 \\
\texttt{HF} & \textbf{95.0$\pm$0.3} & 94.5$\pm$0.9 & 94.6$\pm$0.6 & 91.1$\pm$0.2 & 89.6$\pm$4.0 & 78.1$\pm$2.1 & 90.6$\pm$3.6 \\
\midrule
\multicolumn{8}{@{}l}{\textbf{\texttt{EFS}}} \\[2pt]
\texttt{THB} & \textbf{98.0$\pm$0.4} & 97.9$\pm$0.4 & 96.0$\pm$0.7 & 86.1$\pm$0.9 & 84.0$\pm$0.4 & 75.8$\pm$2.3 & 92.5$\pm$1.3 \\
\texttt{OF} & 94.3$\pm$0.6 & \textbf{95.3$\pm$0.5} & 95.1$\pm$0.4 & 77.7$\pm$2.1 & 78.5$\pm$0.4 & 64.0$\pm$3.1 & 91.4$\pm$2.6 \\
\texttt{DFAV} & \textbf{93.3$\pm$0.9} & 90.1$\pm$1.0 & 89.8$\pm$0.9 & 83.5$\pm$0.3 & 76.7$\pm$2.3 & 80.9$\pm$0.4 & 85.9$\pm$2.2 \\
\texttt{MV} & 95.6$\pm$0.5 & 94.7$\pm$0.4 & \textbf{96.6$\pm$1.2} & 72.0$\pm$1.7 & 64.9$\pm$1.9 & 71.0$\pm$1.6 & 95.4$\pm$0.8 \\
\texttt{DF26} & \textbf{95.2$\pm$0.4} & 94.9$\pm$1.0 & 94.2$\pm$1.2 & 66.7$\pm$0.6 & 62.7$\pm$2.2 & 62.9$\pm$1.3 & 91.9$\pm$1.6 \\
\midrule
All & \textbf{96.1$\pm$0.1} & 94.3$\pm$0.3 & 92.7$\pm$0.4 & 89.4$\pm$0.1 & 86.8$\pm$0.8 & 85.7$\pm$0.5 & 75.7$\pm$0.2 \\
\bottomrule
\end{tabular}
\end{table}

\clearpage
\section{Blob twins against SBI}
\label{sec:supp-ablation-sbi}

Blob twins are simpler than the self-blended images of SBI~\cite{SBI}: the mask is round instead of a face shape built from facial landmarks, and a small shift is the only change of geometry (\cref{sec:blob-twins}). \Cref{tab:supp-ablation-sbi} tests whether this simplification costs accuracy. We keep the full training set and recipe of \method and change only how the blob twins are made. The SBI variant replaces the round mask with the face-shaped mask of SBI and takes the parameters of SBI for altering the copy. It is 0.6 points below blob twins on All.

\section{Partial against full reconstructions}
\label{sec:supp-ablation-partial}

A reconstruction twin puts the real pixels back inside a random region of the reconstruction (\cref{eq:partial}). \Cref{tab:supp-ablation-partial} compares this with full reconstructions, where no real pixels are put back. Everything else is the full training set and recipe of \method.

\begin{table}[h]
\begin{minipage}[t]{0.48\linewidth}
\centering
\small
\caption{\label{tab:supp-ablation-sbi}
    \textbf{Per-dataset comparison of blob twins and SBI.} Video-level AUROC (\%) of single models, mean $\pm$ standard deviation over three seeds. The blob twins of the full training set are replaced by the self-blended images of SBI. The best value in each row is in bold.
}
\begin{tabular}{@{}l|cc@{}}
\toprule
 & Blob twins & SBI \\
\midrule
\multicolumn{3}{@{}l}{\textbf{\texttt{Blending}}} \\[2pt]
\texttt{DFD} & \textbf{98.7$\pm$0.1} & 97.1$\pm$0.6 \\
\texttt{FSh} & \textbf{96.6$\pm$0.3} & \textbf{96.6$\pm$0.3} \\
\texttt{CDFv2} & \textbf{97.0$\pm$0.2} & 96.1$\pm$0.2 \\
\texttt{FFIW} & \textbf{98.6$\pm$0.4} & 97.5$\pm$0.4 \\
\texttt{FAVC} & \textbf{99.9$\pm$0.0} & 99.6$\pm$0.0 \\
\texttt{DFDM} & \textbf{99.8$\pm$0.1} & \textbf{99.8$\pm$0.0} \\
\texttt{PGF} & \textbf{96.8$\pm$0.7} & 95.6$\pm$0.5 \\
\texttt{IDF} & \textbf{99.1$\pm$0.1} & 98.4$\pm$0.3 \\
\texttt{DSv1} & \textbf{93.7$\pm$0.4} & 91.6$\pm$1.5 \\
\texttt{HiDF} & \textbf{98.4$\pm$0.1} & 96.9$\pm$0.6 \\
\midrule
\multicolumn{3}{@{}l}{\textbf{\texttt{B+EFS}}} \\[2pt]
\texttt{CDFv3} & \textbf{92.0$\pm$1.3} & \textbf{92.0$\pm$0.4} \\
\texttt{DSv2} & 92.4$\pm$0.7 & \textbf{93.6$\pm$1.7} \\
\texttt{RF} & \textbf{90.9$\pm$1.1} & \textbf{90.9$\pm$1.2} \\
\texttt{HF} & \textbf{95.0$\pm$0.3} & 93.6$\pm$0.5 \\
\midrule
\multicolumn{3}{@{}l}{\textbf{\texttt{EFS}}} \\[2pt]
\texttt{THB} & \textbf{98.0$\pm$0.4} & 97.6$\pm$0.3 \\
\texttt{OF} & 94.3$\pm$0.6 & \textbf{94.5$\pm$0.5} \\
\texttt{DFAV} & \textbf{93.3$\pm$0.9} & 92.3$\pm$0.9 \\
\texttt{MV} & 95.6$\pm$0.5 & \textbf{96.2$\pm$0.8} \\
\texttt{DF26} & \textbf{95.2$\pm$0.4} & 93.8$\pm$1.7 \\
\midrule
All & \textbf{96.1$\pm$0.1} & 95.5$\pm$0.3 \\
\bottomrule
\end{tabular}

\end{minipage}\hfill
\begin{minipage}[t]{0.48\linewidth}
\centering
\small
\caption{\label{tab:supp-ablation-partial}
    \textbf{Partial against full reconstructions.} Video-level AUROC (\%) of single models, mean $\pm$ standard deviation over three seeds. Partial is the full training set of \method. Full uses the reconstruction with no real pixels put back. The best value in each row is in bold.
}
\begin{tabular}{@{}l|cc@{}}
\toprule
 & Partial & Full \\
\midrule
\multicolumn{3}{@{}l}{\textbf{\texttt{Blending}}} \\[2pt]
\texttt{DFD} & \textbf{98.7$\pm$0.1} & 98.5$\pm$0.2 \\
\texttt{FSh} & 96.6$\pm$0.3 & \textbf{97.3$\pm$0.1} \\
\texttt{CDFv2} & \textbf{97.0$\pm$0.2} & 96.7$\pm$0.8 \\
\texttt{FFIW} & 98.6$\pm$0.4 & \textbf{99.0$\pm$0.3} \\
\texttt{FAVC} & \textbf{99.9$\pm$0.0} & \textbf{99.9$\pm$0.0} \\
\texttt{DFDM} & 99.8$\pm$0.1 & \textbf{99.9$\pm$0.0} \\
\texttt{PGF} & 96.8$\pm$0.7 & \textbf{97.2$\pm$0.6} \\
\texttt{IDF} & 99.1$\pm$0.1 & \textbf{99.4$\pm$0.1} \\
\texttt{DSv1} & \textbf{93.7$\pm$0.4} & 92.3$\pm$0.5 \\
\texttt{HiDF} & 98.4$\pm$0.1 & \textbf{98.7$\pm$0.4} \\
\midrule
\multicolumn{3}{@{}l}{\textbf{\texttt{B+EFS}}} \\[2pt]
\texttt{CDFv3} & \textbf{92.0$\pm$1.3} & 87.5$\pm$0.4 \\
\texttt{DSv2} & \textbf{92.4$\pm$0.7} & 87.5$\pm$1.9 \\
\texttt{RF} & \textbf{90.9$\pm$1.1} & 87.9$\pm$1.3 \\
\texttt{HF} & \textbf{95.0$\pm$0.3} & 93.5$\pm$0.7 \\
\midrule
\multicolumn{3}{@{}l}{\textbf{\texttt{EFS}}} \\[2pt]
\texttt{THB} & \textbf{98.0$\pm$0.4} & 97.2$\pm$0.5 \\
\texttt{OF} & \textbf{94.3$\pm$0.6} & 92.2$\pm$0.5 \\
\texttt{DFAV} & \textbf{93.3$\pm$0.9} & 89.6$\pm$0.6 \\
\texttt{MV} & \textbf{95.6$\pm$0.5} & 90.2$\pm$2.1 \\
\texttt{DF26} & \textbf{95.2$\pm$0.4} & 93.6$\pm$0.3 \\
\midrule
All & \textbf{96.1$\pm$0.1} & 94.6$\pm$0.2 \\
\bottomrule
\end{tabular}

\end{minipage}
\end{table}

\clearpage

\section{Where the gain from CCv2 appears}
\label{sec:supp-gap-split}

Removing the CCv2 reals and their twins lowers the mean AUROC by 1.8 points (\cref{tab:ablations}). This section splits this gain into a part due to the real videos of each test dataset and a part due to its fakes. It gives the hypothesis, the two detectors, why a reference set is needed, the split, a worked example, the results per test dataset, and the limits of the method.

\subsection{Hypothesis}

With twins, the \emph{fake} class is defined by what is done to a real frame, while the \emph{real} class is defined only by the real frames seen in training. FF++ reals are mostly news-style YouTube videos with similar framing and compression. So we add the real videos of CCv2~\cite{CCv2}, which include many identities, cameras, and lighting conditions, and make reconstruction and blob twins from them as well. Each twin is paired with the real frame it was made from (\cref{sec:pairs}), so CCv2's capture conditions appear in both classes and cannot become a cue for either. We hypothesize that a detector trained on FF++ reals alone loses AUROC on other datasets mainly on their real videos, and that the gain from CCv2 appears on these real videos.

\subsection{The two detectors}
\label{sec:supp-gap-detectors}

The detector $h_1$ is trained on FF++ reals, their reconstruction and blob twins, and FF++ fakes. The detector $h_2$ also uses the CCv2 reals and their twins, like the single models of \method. In \cref{tab:ablations}, $h_1$ is the row \q{w/o CCv2 reals and twins} and $h_2$ is the first row. Neither is a soup. Each is trained with seeds 1, 2 and 3, and $h_1$ with seed $k$ is compared with $h_2$ with seed $k$. The two configurations differ only in the CCv2 data. Both draw the same number of pairs per epoch, so $h_2$ sees each FF++ real 16.7\% less often than $h_1$ does. The change from $h_1$ to $h_2$ is therefore a change of the training mixture at a fixed budget, not the same training with extra reals. Both detectors are trained on FF++ reals and FF++ fakes, so the FF++ test set is in-domain for both, in both classes.

\subsection{Why a reference set is needed}

Let $s_h(x)$ be the score that detector $h$ gives to video $x$, the mean probability of fake over its frames. For a set of real videos $\mathcal R$ and a set of fake videos $\mathcal F$, the AUROC is the share of (real, fake) pairs that the detector orders correctly, where a tie counts as half:
\begin{equation}
  \mathrm{AUC}_h(\mathcal R, \mathcal F) = \Pr\big[s_h(x') > s_h(x)\big] + \tfrac12 \Pr\big[s_h(x') = s_h(x)\big], \qquad x \sim \mathcal R,\; x' \sim \mathcal F.
  \label{eq:auc}
\end{equation}
A wrongly ordered pair is one real and one fake with the real scored higher. Nothing in that pair says whether the real is scored too high or the fake too low. So the AUROC on a test dataset $\mathcal T$ alone cannot say which class a gain comes from. The scores of two detectors cannot be compared directly either. The AUROC does not change when the scores of a detector are passed through any strictly increasing function, so the scores of $h_1$ and $h_2$ are not on a common scale. Any attribution to one class needs a reference from outside $\mathcal T$.

\subsection{The gap split}

We compare each test dataset $\mathcal T$ with a reference, the FF++ test set $\mathcal S$. It is in-domain for both detectors, in both classes. The generalization gap of $h$ on $\mathcal T$, with reals $\mathcal T^{r}$ and fakes $\mathcal T^{f}$, is
\begin{equation}
  G_h(\mathcal T) = \mathrm{AUC}_h(\mathcal T^{r}, \mathcal T^{f}) - \mathrm{AUC}_h(\mathcal S^{r}, \mathcal S^{f}).
  \label{eq:gap}
\end{equation}
Going from $\mathcal S$ to $\mathcal T$ swaps the reals and the fakes. We do it in two steps, one class at a time. The \emph{real part} of the gap, $\phi_h^{r}$, is the change in AUROC at the step that swaps the reals. It depends on which fakes the reals are ranked against, those of $\mathcal S$ or of $\mathcal T$, so we average both cases:
\begin{equation}
  \begin{aligned}
    \phi_h^{r}(\mathcal T) = {} & \tfrac12 \big[ \mathrm{AUC}_h(\mathcal T^{r}, \mathcal S^{f}) - \mathrm{AUC}_h(\mathcal S^{r}, \mathcal S^{f}) \big] \\
    + {} & \tfrac12 \big[ \mathrm{AUC}_h(\mathcal T^{r}, \mathcal T^{f}) - \mathrm{AUC}_h(\mathcal S^{r}, \mathcal T^{f}) \big].
  \end{aligned}
  \label{eq:phi-r}
\end{equation}
The \emph{fake part} $\phi_h^{f}$ is defined likewise for the fakes:
\begin{equation}
  \begin{aligned}
    \phi_h^{f}(\mathcal T) = {} & \tfrac12 \big[ \mathrm{AUC}_h(\mathcal S^{r}, \mathcal T^{f}) - \mathrm{AUC}_h(\mathcal S^{r}, \mathcal S^{f}) \big] \\
    + {} & \tfrac12 \big[ \mathrm{AUC}_h(\mathcal T^{r}, \mathcal T^{f}) - \mathrm{AUC}_h(\mathcal T^{r}, \mathcal S^{f}) \big].
  \end{aligned}
  \label{eq:phi-f}
\end{equation}
This average is the Shapley value~\scite{Shapley} with the reals and the fakes as two players, as in~\scite{ShiftAttribution}. The two parts sum to the gap exactly, $\phi_h^{r} + \phi_h^{f} = G_h$. A negative real part means that the detector scores the reals of $\mathcal T$ as more fake than the reals of $\mathcal S$. A negative fake part means that it scores the fakes of $\mathcal T$ as more real than the fakes of $\mathcal S$. So both parts are relative to the reference, and another reference would give other values.

By \cref{eq:gap}, the AUROC on $\mathcal T$ is the AUROC on $\mathcal S$ plus the two parts. Taking the difference between $h_2$ and $h_1$ splits the gain on $\mathcal T$ into three terms:
\begin{equation}
  \Delta\mathrm{AUROC}(\mathcal T) = \Delta\phi^{r} + \Delta\phi^{f} + \Delta\mathrm{AUROC}(\mathcal S),
  \label{eq:gain}
\end{equation}
where $\Delta$ is the value for $h_2$ minus the value for $h_1$, so $\Delta\phi^{r} = \phi_{h_2}^{r} - \phi_{h_1}^{r}$. Every AUROC ranks the scores of one detector only. So the split needs no threshold or calibration and never compares scores of $h_1$ with scores of $h_2$.

\subsection{The four AUROCs and the interaction}

For one detector $h$ and one test dataset $\mathcal T$, the split uses four AUROCs, all computed from the scores of $h$:
\begin{center}
\small
\begin{tabular}{@{}l|cc@{}}
\toprule
 & fakes of $\mathcal S$ & fakes of $\mathcal T$ \\
\midrule
reals of $\mathcal S$ & $a = \mathrm{AUC}_h(\mathcal S^{r}, \mathcal S^{f})$ & $c = \mathrm{AUC}_h(\mathcal S^{r}, \mathcal T^{f})$ \\
reals of $\mathcal T$ & $b = \mathrm{AUC}_h(\mathcal T^{r}, \mathcal S^{f})$ & $d = \mathrm{AUC}_h(\mathcal T^{r}, \mathcal T^{f})$ \\
\bottomrule
\end{tabular}
\end{center}
Here $a$ is the in-domain AUROC, $d$ is the actual AUROC on $\mathcal T$, and the gap of \cref{eq:gap} is $G_h = d - a$. In $b$ only the reals come from $\mathcal T$, and in $c$ only the fakes. The parts of \cref{eq:phi-r,eq:phi-f} are
\begin{equation}
  \phi_h^{r} = \tfrac12 \big[(b - a) + (d - c)\big], \qquad \phi_h^{f} = \tfrac12 \big[(c - a) + (d - b)\big].
\end{equation}
Each bracket holds one class fixed and swaps the other from $\mathcal S$ to $\mathcal T$. The first difference swaps a class while the other class is still from $\mathcal S$. The second swaps it when the other class is already from $\mathcal T$. Averaging the two removes the dependence on the order of the swaps, which is the Shapley value~\scite{Shapley} for two players. The \emph{interaction} is what the two swaps do together beyond what each does alone,
\begin{equation}
  I_h = d - b - c + a,
\end{equation}
and each part is the effect of its swap done first plus half of the interaction: $\phi_h^{r} = (b - a) + \tfrac12 I_h$ and $\phi_h^{f} = (c - a) + \tfrac12 I_h$. So a part is not a property of the videos of one class alone. It also depends on the other class, and the reported value is the average over both cases.

The Shapley value is linear in the quantity that is split. So $\Delta\phi^{r}$ and $\Delta\phi^{f}$ in \cref{eq:gain} are the Shapley split of the difference between the gaps of the two detectors, $G_{h_2}(\mathcal T) - G_{h_1}(\mathcal T)$.

\subsection{Worked example}

\Cref{tab:supp-gap-example} gives the four AUROCs of both detectors for $\mathcal T = \texttt{DSv1}$. For $h_1$,
\begin{equation}
  \phi_{h_1}^{r} = \tfrac12 \big[(87.8 - 99.4) + (81.1 - 99.9)\big] = -15.2, \qquad
  \phi_{h_1}^{f} = \tfrac12 \big[(99.9 - 99.4) + (81.1 - 87.8)\big] = -3.1.
\end{equation}
Of the 18.3 points that $h_1$ loses on \texttt{DSv1}, 15.2 come from the reals of \texttt{DSv1} and 3.1 from its fakes. For $h_2$, the parts are $\phi_{h_2}^{r} = -4.4$ and $\phi_{h_2}^{f} = -1.2$. The gain of $+12.6$ points from $h_1$ to $h_2$ splits into $+10.8$ from reals, $+1.8$ from fakes, and $-0.1$ from the \texttt{FF++} test set. The values are rounded, so the sums can be off by 0.1.

\begin{table}[h]
  \centering
  \small
  \caption{\label{tab:supp-gap-example}
    \textbf{The four AUROCs (\%) of the gap split for \texttt{DSv1}.} Mean of 3 seeds.
  }
  \begin{tabular}{@{}l cc cc@{}}
    \toprule
    & \multicolumn{2}{c}{$h_1$: without CCv2} & \multicolumn{2}{c}{$h_2$: with CCv2} \\
    \cmidrule(lr){2-3} \cmidrule(l){4-5}
    & \texttt{FF++} fakes & \texttt{DSv1} fakes & \texttt{FF++} fakes & \texttt{DSv1} fakes \\
    \midrule
    \texttt{FF++} reals & 99.4 & 99.9 & 99.4 & 99.8 \\
    \texttt{DSv1} reals & 87.8 & 81.1 & 96.7 & 93.7 \\
    \bottomrule
  \end{tabular}
\end{table}

\subsection{Results per test dataset}

\Cref{tab:gap-split} gives the split per test dataset with the spread over the three seeds, and \cref{tab:supp-gap-detectors} gives the gap of each detector and its two parts.

Adding CCv2 raises the mean AUROC by 1.8 points. Of these, $+2.0$ points are the real part, $-0.1$ the fake part, and $-0.1$ the change on the FF++ test set. So little of the gain is an in-domain change.

The real part of the gain is positive on 17 of the 19 test datasets. On 15 of them it is positive for every seed, while \texttt{IDF} and \texttt{FFIW} have one negative seed. It is exactly zero on \texttt{FSh}, whose reals are the FF++ test reals. This is a built-in check of the computation. The fake part has no common sign, so CCv2 has no consistent effect on the fakes. It is positive on \texttt{DSv1}, \texttt{DFAV}, \texttt{MV}, \texttt{CDFv3}, \texttt{PGF}, \texttt{FSh} and \texttt{HiDF}, zero on \texttt{FAVC}, and negative on 11 test datasets.

The largest AUROC gains are on \texttt{DSv1}, \texttt{DFAV}, and \texttt{DSv2}, where the gain of the real part is also the largest. The reals of \texttt{DSv1} and \texttt{DSv2} are self-recorded by participants on their own webcams~\cite{DSv1}, unlike the news-style YouTube videos of FF++. On \texttt{THB}, the negative fake part outweighs a small positive real part. The AUROC drops most on \texttt{OF} ($-3.3$), where both parts are negative.

For $h_1$, the mean gap is mostly on the real side, $-4.4$ points from reals and $-0.7$ from fakes. The $-0.7$ is an average of negative and positive test datasets (\texttt{DFAV} $-5.2$, \texttt{FSh} $-4.0$ and \texttt{RF} $-3.8$ against \texttt{DFD} $+2.5$ and \texttt{DFDM} $+2.3$). A positive fake part means that the fakes of the test dataset are easier for the detector than the FF++ fakes. Adding CCv2 changes the mean real part from $-4.4$ to $-2.4$ and the mean fake part from $-0.7$ to $-0.9$. The remaining gap of $h_2$ is still mostly on the real side (\texttt{CDFv3} $-5.5$, \texttt{DSv2} $-4.9$, \texttt{DSv1} $-4.4$). The exceptions are \texttt{RF}, \texttt{OF} and \texttt{DF26}, where the fake part is the larger one.

\paragraph{Interpretation} The results support the hypothesis. Relative to FF++, the gap of $h_1$ is mostly on the real side: $h_1$ scores the real videos of the other datasets as more fake than the FF++ test reals (\cref{tab:supp-gap-detectors}). Most of the gain from CCv2 appears on these real videos. The split shows on which test class the gain appears, not what in the added training data causes it. In a training set without FF++ fakes, adding the CCv2 reals without their twins does not raise the mean AUROC, while adding them with their twins does (\cref{sec:supp-ccv2-reals}). So real frames help only with their twins.

\begin{table}[h]
  \centering
  \small
  \setlength{\tabcolsep}{5pt}
  \caption{\label{tab:gap-split}
    \textbf{Which test class the gain from CCv2 shows up on.} Change in video-level AUROC (\%) when the CCv2 reals and their twins are added to training, split by \cref{eq:gain}: $\Delta\mathrm{AUROC} = \Delta\phi^{r} + \Delta\phi^{f} + \Delta\mathrm{AUROC}(\mathcal S)$. $\Delta\phi^{r}$ and $\Delta\phi^{f}$ are the real and the fake part of the gain, relative to the FF++ test set $\mathcal S$, and $\Delta\mathrm{AUROC}(\mathcal S) = -0.05$ in every row. Values are rounded, so a row can be off by 0.1. The detector $h_1$ is trained without CCv2. Mean of 3 paired seeds, with the minimum and maximum over the seeds in parentheses. Green: positive in all three seeds. Red: negative in all three seeds. Gray: the sign differs between the seeds, or the value is zero. Sorted by $\Delta\mathrm{AUROC}$.
  }
  \begin{tabular}{@{}l r r rr@{}}
    \toprule
    & \bf AUROC & \bf Gain & \bf Reals & \bf Fakes \\
    \textbf{Dataset} & of $h_1$ & $\Delta\mathrm{AUROC}$ & $\Delta\phi^{r}$ & $\Delta\phi^{f}$ \\
    \midrule
    \texttt{DSv1} & 81.1 & \up{12.6} {\scriptsize($+11.5$, $+13.7$)} & \up{10.8} {\scriptsize($+9.4$, $+11.9$)} & \up{1.8} {\scriptsize($+1.4$, $+2.2$)} \\
    \texttt{DFAV} & 86.8 & \up{6.5} {\scriptsize($+4.1$, $+8.3$)} & \up{4.4} {\scriptsize($+3.2$, $+5.3$)} & \up{2.2} {\scriptsize($+1.0$, $+3.0$)} \\
    \texttt{DSv2} & 86.9 & \up{5.5} {\scriptsize($+4.1$, $+6.7$)} & \up{6.2} {\scriptsize($+5.5$, $+7.7$)} & \ns{$-$0.7} {\scriptsize($-1.4$, $+0.1$)} \\
    \texttt{MV} & 92.4 & \up{3.1} {\scriptsize($+1.8$, $+4.2$)} & \up{2.6} {\scriptsize($+1.9$, $+2.9$)} & \ns{$+$0.6} {\scriptsize($0.0$, $+1.4$)} \\
    \texttt{CDFv3} & 89.9 & \up{2.1} {\scriptsize($+1.9$, $+2.6$)} & \up{1.5} {\scriptsize($+1.1$, $+1.9$)} & \up{0.6} {\scriptsize($+0.3$, $+0.9$)} \\
    \texttt{PGF} & 94.9 & \up{1.9} {\scriptsize($+1.7$, $+2.1$)} & \up{1.6} {\scriptsize($+1.5$, $+1.8$)} & \up{0.4} {\scriptsize($+0.3$, $+0.6$)} \\
    \texttt{HF} & 93.2 & \up{1.8} {\scriptsize($+1.6$, $+2.0$)} & \up{2.6} {\scriptsize($+2.1$, $+3.0$)} & \dn{0.7} {\scriptsize($-1.1$, $-0.4$)} \\
    \texttt{FSh} & 95.4 & \up{1.1} {\scriptsize($+0.9$, $+1.3$)} & \ns{0.0} {\scriptsize($0.0$, $0.0$)} & \up{1.2} {\scriptsize($+0.9$, $+1.3$)} \\
    \texttt{DFD} & 97.6 & \up{1.1} {\scriptsize($+0.9$, $+1.3$)} & \up{2.3} {\scriptsize($+1.7$, $+2.8$)} & \dn{1.1} {\scriptsize($-1.5$, $-0.7$)} \\
    \texttt{HiDF} & 97.6 & \up{0.8} {\scriptsize($+0.1$, $+1.4$)} & \up{0.7} {\scriptsize($+0.3$, $+1.1$)} & \ns{$+$0.1} {\scriptsize($-0.1$, $+0.3$)} \\
    \texttt{DF26} & 94.7 & \up{0.5} {\scriptsize($+0.1$, $+0.8$)} & \up{1.2} {\scriptsize($+1.0$, $+1.3$)} & \dn{0.7} {\scriptsize($-0.9$, $-0.4$)} \\
    \texttt{CDFv2} & 96.7 & \ns{$+$0.4} {\scriptsize($-0.4$, $+0.8$)} & \up{0.7} {\scriptsize($0.0$, $+1.0$)} & \dn{0.2} {\scriptsize($-0.3$, $-0.1$)} \\
    \texttt{RF} & 90.6 & \ns{$+$0.3} {\scriptsize($-0.2$, $+1.1$)} & \up{1.9} {\scriptsize($+1.1$, $+3.1$)} & \dn{1.5} {\scriptsize($-1.9$, $-1.3$)} \\
    \texttt{FAVC} & 99.9 & \ns{0.0} {\scriptsize($0.0$, $+0.1$)} & \up{0.1} {\scriptsize($+0.1$, $+0.3$)} & \ns{0.0} {\scriptsize($-0.2$, $+0.1$)} \\
    \texttt{DFDM} & 99.8 & \ns{0.0} {\scriptsize($0.0$, $+0.1$)} & \up{0.5} {\scriptsize($+0.2$, $+0.6$)} & \dn{0.4} {\scriptsize($-0.5$, $-0.1$)} \\
    \texttt{IDF} & 99.2 & \ns{$-$0.1} {\scriptsize($-0.4$, $+0.1$)} & \ns{$+$0.3} {\scriptsize($-0.2$, $+0.8$)} & \dn{0.3} {\scriptsize($-0.8$, $-0.1$)} \\
    \texttt{FFIW} & 98.9 & \dn{0.2} {\scriptsize($-0.6$, $-0.1$)} & \ns{$+$0.1} {\scriptsize($-0.3$, $+0.3$)} & \dn{0.3} {\scriptsize($-0.4$, $-0.1$)} \\
    \texttt{THB} & 98.7 & \dn{0.7} {\scriptsize($-1.2$, $-0.4$)} & \up{0.3} {\scriptsize($0.0$, $+0.7$)} & \dn{1.0} {\scriptsize($-1.2$, $-0.6$)} \\
    \texttt{OF} & 97.6 & \dn{3.3} {\scriptsize($-3.6$, $-2.7$)} & \dn{0.7} {\scriptsize($-0.9$, $-0.6$)} & \dn{2.5} {\scriptsize($-2.9$, $-2.0$)} \\
    \midrule
    \textbf{Mean} (19) & 94.3 & \up{1.8} {\scriptsize($+1.6$, $+1.9$)} & \up{2.0} {\scriptsize($+1.7$, $+2.1$)} & \ns{$-$0.1} {\scriptsize($-0.2$, $0.0$)} \\
    \bottomrule
  \end{tabular}
\end{table}

\begin{table}[h]
  \centering
  \small
  \caption{\label{tab:supp-gap-detectors}
    \textbf{Gap of each detector, split into reals and fakes.} Video-level AUROC (\%) on each test dataset and the two parts of the gap to the AUROC on the FF++ test set (\cref{eq:phi-r,eq:phi-f}), which is 99.4 for both $h_1$ and $h_2$. Mean of 3 seeds. Same order as \cref{tab:gap-split}.
  }
  \begin{tabular}{@{}l rrr rrr@{}}
    \toprule
    & \multicolumn{3}{c}{$h_1$: without CCv2} & \multicolumn{3}{c}{$h_2$: with CCv2} \\
    \cmidrule(lr){2-4} \cmidrule(l){5-7}
    \textbf{Dataset} & AUROC & $\phi^{r}$ & $\phi^{f}$ & AUROC & $\phi^{r}$ & $\phi^{f}$ \\
    \midrule
    \texttt{DSv1} & 81.1 & $-15.2$ & $-3.1$ & 93.7 & $-4.4$ & $-1.2$ \\
    \texttt{DFAV} & 86.8 & $-7.4$ & $-5.2$ & 93.3 & $-3.1$ & $-3.0$ \\
    \texttt{DSv2} & 86.9 & $-11.1$ & $-1.4$ & 92.4 & $-4.9$ & $-2.1$ \\
    \texttt{MV} & 92.4 & $-5.0$ & $-2.0$ & 95.6 & $-2.4$ & $-1.4$ \\
    \texttt{CDFv3} & 89.9 & $-7.0$ & $-2.5$ & 92.0 & $-5.5$ & $-1.9$ \\
    \texttt{PGF} & 94.9 & $-3.6$ & $-1.0$ & 96.8 & $-2.0$ & $-0.6$ \\
    \texttt{HF} & 93.2 & $-5.6$ & $-0.7$ & 95.0 & $-3.0$ & $-1.4$ \\
    \texttt{FSh} & 95.4 & $0.0$ & $-4.0$ & 96.6 & $0.0$ & $-2.8$ \\
    \texttt{DFD} & 97.6 & $-4.4$ & $+2.5$ & 98.7 & $-2.1$ & $+1.4$ \\
    \texttt{HiDF} & 97.6 & $-2.0$ & $+0.2$ & 98.4 & $-1.3$ & $+0.4$ \\
    \texttt{DF26} & 94.7 & $-3.2$ & $-1.5$ & 95.2 & $-2.0$ & $-2.2$ \\
    \texttt{CDFv2} & 96.7 & $-3.7$ & $+0.9$ & 97.1 & $-3.0$ & $+0.7$ \\
    \texttt{RF} & 90.6 & $-5.0$ & $-3.8$ & 90.9 & $-3.1$ & $-5.4$ \\
    \texttt{FAVC} & 99.9 & $-0.8$ & $+1.2$ & 99.9 & $-0.6$ & $+1.2$ \\
    \texttt{DFDM} & 99.8 & $-2.0$ & $+2.3$ & 99.8 & $-1.5$ & $+2.0$ \\
    \texttt{IDF} & 99.2 & $-2.1$ & $+1.9$ & 99.1 & $-1.9$ & $+1.6$ \\
    \texttt{FFIW} & 98.9 & $-1.8$ & $+1.2$ & 98.6 & $-1.7$ & $+1.0$ \\
    \texttt{THB} & 98.7 & $-1.8$ & $+1.0$ & 98.0 & $-1.4$ & $+0.1$ \\
    \texttt{OF} & 97.6 & $-1.6$ & $-0.3$ & 94.3 & $-2.3$ & $-2.8$ \\
    \midrule
    \textbf{Mean} (19) & 94.3 & $-4.4$ & $-0.7$ & 96.1 & $-2.4$ & $-0.9$ \\
    \bottomrule
  \end{tabular}
\end{table}

\subsection{What the split does and does not show}

The split has four useful properties. Every AUROC uses the scores of one detector, so the result does not change under any strictly increasing rescaling of the scores of either detector. It needs no threshold, calibration or normalization, and its only choice is the reference set. Its parts are in AUROC points and sum exactly to the gain. It needs only the stored predictions on $\mathcal S$ and $\mathcal T$.

The split also has three limits.

\paragraph{It is relative to the reference} The split is exact for any reference, but \q{from reals} and \q{from fakes} always mean compared to the reals and fakes of $\mathcal S$. A small change of the AUROC on $\mathcal S$ does not show that $\mathcal S$ is a stable reference. Consider one video per group with the following scores:
\begin{center}
\small
\begin{tabular}{@{}l|cccc@{}}
\toprule
 & $\mathcal S^{r}$ & $\mathcal S^{f}$ & $\mathcal T^{r}$ & $\mathcal T^{f}$ \\
\midrule
$h_1$ & 0.1 & 0.2 & 0.4 & 0.6 \\
$h_2$ & 0.1 & 0.9 & 0.4 & 0.6 \\
\bottomrule
\end{tabular}
\end{center}
Both detectors have an AUROC of 100 on $\mathcal S$ and on $\mathcal T$, and the same scores on $\mathcal T$. Yet $h_1$ gets $\phi^{r} = -50$ and $\phi^{f} = +50$, because its real of $\mathcal T$ scores higher than its fake of $\mathcal S$, and $h_2$ gets $0$ and $0$. The comparison reads $+50$ from reals and $-50$ from fakes, although nothing changed on $\mathcal T$. So the choice of the FF++ test set as the reference rests on the training setup: both detectors are trained on FF++ reals and FF++ fakes.

\paragraph{The parts are contributions to a gap, not errors} They split a difference of AUROCs and can cancel. A fake part of zero does not mean that the fakes of the test dataset cause no errors.

\paragraph{A test class is not a training cause} The split shows on which test class the gain appears. It does not show whether the CCv2 reals or their twins cause it (\cref{sec:supp-ccv2-reals}).

\subsection{Relation to prior work}

Zhang \etal~\scite{ShiftAttribution} attribute the change in the performance of a fixed model between two environments to shifts of distributions. The players of their Shapley value are distributions, and the value of a set of players is the performance of the model when only these distributions are switched to the target. The gap split uses the same game with other players and another metric. Their players are the causal mechanisms of a known causal graph, their metric is an average of a loss per data point, and they estimate the performance under a mixed distribution by importance weighting. Here the players are the real and the fake videos of the test dataset, the metric is the AUROC, which is an average over (real, fake) pairs, and every mixed AUROC is computed directly from the stored scores. Comparing two detectors is not part of their method. It follows from the linearity of the Shapley value.

\subsection{CCv2 reals without their twins}
\label{sec:supp-ccv2-reals}

To test whether more real frames alone explain the gain, we train three detectors without FF++ fakes, so that all fakes are twins (\cref{tab:supp-ccv2-reals}). Adding the CCv2 reals without their twins does not raise the mean AUROC. It lowers it by 1.7 points. Adding the CCv2 reals and their twins raises it by 3.0 points. So in this setting the gain comes from real frames paired with fakes made from them, not from more real frames.

\begin{table}[h]
\centering
\small
\caption{\label{tab:supp-ccv2-reals}
    \textbf{CCv2 reals with and without their twins.} Video-level AUROC (\%) of single models on all 19 test datasets, mean $\pm$ standard deviation over three seeds. All three detectors are trained without FF++ fakes. $\Delta$ is the change against the first row in percentage points.
}
\begin{tabular}{@{}cccc|cr@{}}
\toprule
\multicolumn{2}{c}{FF++} & \multicolumn{2}{c|}{CCv2} & & \\
\cmidrule(r){1-2} \cmidrule(lr){3-4}
reals & twins & reals & twins & All & $\Delta$ \\
\midrule
\checkmark & \checkmark & $-$ & $-$ & 91.3$\pm$0.4 & \\
\checkmark & \checkmark & \checkmark & $-$ & 89.6$\pm$0.6 & \dn{1.7} \\
\checkmark & \checkmark & \checkmark & \checkmark & \textbf{94.3$\pm$0.3} & \up{3.0} \\
\bottomrule
\end{tabular}
\end{table}

\clearpage
\section{Results per generator}
\label{sec:supp-generators}

\Cref{tab:supp-generators} extends \cref{tab:generators} of the main paper to all 15 prior detectors. None of the seven detectors that the main paper leaves out is above \method on any generator.

\begin{table}[h]
\centering
\caption{\label{tab:supp-generators}
    \textbf{Results per generator.} Video-level AUROC (\%) on \texttt{DF26}, \texttt{OF} and \texttt{MV} for each of their generators. The row of \method is the AUROC. Every other row is the difference to \method in percentage points, green if higher and red if lower. This is \cref{tab:generators} of the main paper with all 15 prior detectors, in the order of \cref{tab:sota}. The rows under \method make the reconstruction twins with another autoencoder instead of CogVideoX. HV is HunyuanVideo, T2V is text-to-video and I2V is image-to-video. Mean is the average over the 17 columns. The best value in each column is in bold. \textsuperscript{\dag}A temporal model, which takes several frames as input.
}
\tabcolsep=3.4pt
\resizebox{\textwidth}{!}{
\begin{tabular}{@{}l|cccccccccc|cccc|ccc|c@{}}
\toprule
 & \multicolumn{10}{c|}{\texttt{DF26}~\cite{DF26}} & \multicolumn{4}{c|}{\texttt{OF}~\cite{OmniFake}} & \multicolumn{3}{c|}{\texttt{MV}~\cite{MintVid}} & \\
 & \multicolumn{4}{c}{\it closed-source} & \multicolumn{6}{c|}{\it open-source} & \multicolumn{4}{c|}{} & \multicolumn{3}{c|}{} & \\[-2ex]
 & \multicolumn{4}{@{\hskip 4pt}c@{\hskip 4pt}}{\leaders\hrule height 0.3pt\hskip 0pt plus 1filll\null} & \multicolumn{6}{@{\hskip 4pt}c@{\hskip 4pt}|}{\leaders\hrule height 0.3pt\hskip 0pt plus 1filll\null} & \multicolumn{4}{c|}{} & \multicolumn{3}{c|}{} & \\
 & \rotatebox{90}{Grok T2V} & \rotatebox{90}{Kling T2V} & \rotatebox{90}{Veo3 T2V} & \rotatebox{90}{Wan 2.6 T2V} & \rotatebox{90}{HV 1.5 I2V} & \rotatebox{90}{HV 1.5 T2V} & \rotatebox{90}{Wan 2.2 I2V} & \rotatebox{90}{Wan 2.2 T2V} & \rotatebox{90}{LTX 2.3 I2V} & \rotatebox{90}{LTX 2.3 T2V} & \rotatebox{90}{ACTalker} & \rotatebox{90}{Ditto} & \rotatebox{90}{EDTalk} & \rotatebox{90}{FLOAT} & \rotatebox{90}{FantasyTalk.} & \rotatebox{90}{OmniAvatar} & \rotatebox{90}{Phantom 14B} & \rotatebox{90}{Mean} \\
\midrule
\multicolumn{19}{@{}l}{\textit{AIGC detectors}} \\[2pt]
Effort-SDv1.4~\cite{Effort} \C{Effort-sdv14-D26CVPR} & \dn{22.9} & \dn{23.0} & \dn{22.3} & \dn{28.7} & \dn{40.1} & \dn{30.6} & \dn{50.2} & \dn{46.9} & \dn{43.6} & \dn{36.1} & \dn{17.5} & \dn{11.9} & \dn{21.1} & \dn{20.0} & \dn{68.6} & \dn{56.3} & \dn{44.3} & \dn{34.4} \\
DGS-Net~\cite{DGS-Net} \C{DGS-Net-D26CVPR} & \dn{11.8} & \dn{40.7} & \dn{10.8} & \dn{21.5} & \dn{21.8} & \dn{4.5} & \dn{33.4} & \dn{23.5} & \dn{25.5} & \dn{19.8} & \dn{40.6} & \dn{39.4} & \dn{70.3} & \dn{30.1} & \dn{24.1} & \dn{33.7} & \dn{17.7} & \dn{27.6} \\
LTD-UFD~\cite{LTD} \C{LTD-UFD-D26CVPR} & \dn{2.7} & \dn{40.7} & \dn{0.5} & \up{2.8} & \dn{9.3} & \dn{2.8} & \dn{16.4} & \up{0.7} & \dn{13.2} & \dn{4.7} & \dn{34.5} & \dn{64.3} & \dn{79.6} & \dn{60.4} & \dn{64.5} & \dn{47.1} & \dn{5.6} & \dn{26.0} \\
AlignedFor.~\cite{AlignedForensics} \C{AlignedForensics-Sync-D26CVPR} & \dn{5.4} & \textbf{\boldmath\up{3.5}} & \dn{15.0} & \dn{4.0} & \dn{40.5} & \dn{51.0} & \dn{48.6} & \dn{54.4} & \dn{21.4} & \dn{21.9} & \dn{2.1} & \dn{33.4} & \dn{30.9} & \dn{64.1} & \dn{86.5} & \dn{64.4} & \dn{13.9} & \dn{32.6} \\
DDA~\cite{DDA} \C{DDA-D26CVPR} & \dn{8.4} & \up{1.6} & \dn{9.8} & \textbf{\boldmath\up{7.6}} & \dn{14.8} & \dn{28.3} & \dn{25.9} & \dn{33.3} & \dn{24.0} & \dn{34.7} & \dn{0.2} & \dn{70.1} & \dn{66.4} & \dn{62.0} & \dn{54.1} & \dn{11.1} & \dn{6.1} & \dn{25.9} \\
B-Free~\cite{B-Free} \C{B-Free-D26CVPR} & \dn{53.0} & \dn{23.2} & \dn{41.1} & \up{6.5} & \dn{20.1} & \dn{18.3} & \dn{22.2} & \dn{33.4} & \dn{21.6} & \dn{41.6} & \dn{0.1} & \dn{48.5} & \dn{62.8} & \dn{46.0} & \dn{27.7} & \dn{9.4} & \dn{23.4} & \dn{28.6} \\
\midrule
\multicolumn{19}{@{}l}{\textit{Facial deepfake detectors}} \\[2pt]
SBI~\cite{SBI} \C{SBI-FFraw-D26CVPR} & \dn{39.5} & \dn{16.0} & \dn{28.8} & \dn{26.8} & \dn{52.1} & \dn{24.4} & \dn{47.8} & \dn{47.6} & \dn{37.5} & \dn{36.5} & \dn{23.1} & \dn{23.7} & \dn{13.4} & \dn{18.3} & \dn{56.0} & \dn{40.2} & \dn{58.9} & \dn{34.7} \\
FSFM~\cite{FSFM} \C{FSFM-D26CVPR} & \dn{61.1} & \dn{45.5} & \dn{56.9} & \dn{22.2} & \dn{48.8} & \dn{24.3} & \dn{32.6} & \dn{38.7} & \dn{32.4} & \dn{55.2} & \dn{31.6} & \dn{54.7} & \dn{31.4} & \dn{55.5} & \dn{34.0} & \dn{18.4} & \dn{44.4} & \dn{40.5} \\
ForAda~\cite{ForAda} \C{ForAda-D26CVPR} & \dn{58.6} & \dn{41.9} & \dn{55.4} & \dn{27.7} & \dn{44.3} & \dn{21.8} & \dn{32.7} & \dn{45.4} & \dn{26.8} & \dn{38.9} & \dn{10.1} & \dn{45.8} & \dn{11.6} & \dn{39.9} & \dn{37.9} & \dn{14.1} & \dn{35.9} & \dn{34.6} \\
Effort~\cite{Effort} \C{Effort-D26CVPR} & \dn{83.8} & \dn{62.8} & \dn{73.8} & \dn{62.3} & \dn{50.8} & \dn{30.8} & \dn{33.5} & \dn{63.4} & \dn{36.0} & \dn{51.1} & \dn{3.1} & \dn{29.1} & \dn{17.9} & \dn{18.2} & \dn{45.3} & \dn{28.6} & \dn{39.7} & \dn{43.0} \\
DFD-FCG\textsuperscript{\dag}~\cite{DFD-FCG} \C{DFD-FCG-x1.4-D26CVPR} & \dn{75.4} & \dn{59.2} & \dn{59.7} & \dn{28.5} & \dn{48.3} & \dn{20.0} & \dn{30.7} & \dn{36.7} & \dn{31.1} & \dn{49.0} & \dn{14.9} & \dn{20.8} & \dn{17.1} & \dn{20.3} & \dn{32.7} & \dn{33.3} & \dn{22.8} & \dn{35.3} \\
VLAForge~\cite{VLAForge} \C{VLAForge-D26CVPR} & \dn{64.0} & \dn{36.1} & \dn{64.0} & \dn{30.2} & \dn{46.1} & \dn{20.2} & \dn{33.8} & \dn{46.9} & \dn{33.7} & \dn{49.3} & \dn{5.4} & \dn{34.8} & \dn{15.0} & \dn{26.9} & \dn{48.5} & \dn{18.7} & \dn{29.2} & \dn{35.5} \\
FS-VFM~\cite{FS-VFM} \C{FS-VFM-D26CVPR} & \dn{43.7} & \dn{35.2} & \dn{40.3} & \dn{16.5} & \dn{48.0} & \dn{19.2} & \dn{32.5} & \dn{51.5} & \dn{27.0} & \dn{47.4} & \dn{4.7} & \dn{36.8} & \dn{16.8} & \dn{38.4} & \dn{44.4} & \dn{17.5} & \dn{42.7} & \dn{33.1} \\
DFD-HR~\cite{DFD-HR} \C{DFD-HR-D26CVPR} & \dn{53.5} & \dn{23.1} & \dn{45.2} & \dn{27.0} & \dn{33.8} & \dn{15.2} & \dn{22.2} & \dn{45.4} & \dn{13.8} & \dn{40.0} & \dn{2.0} & \dn{31.1} & \dn{6.1} & \dn{30.3} & \dn{22.5} & \dn{4.7} & \dn{15.3} & \dn{25.4} \\
GenD-PE~\cite{GenD} \C{GenD-PE-HF-D26CVPR} & \dn{30.9} & \dn{24.2} & \dn{18.6} & \dn{3.1} & \dn{30.3} & \dn{17.0} & \dn{29.2} & \dn{35.0} & \dn{16.2} & \dn{25.1} & \dn{2.8} & \dn{37.0} & \dn{7.2} & \dn{31.4} & \dn{27.3} & \dn{6.9} & \dn{11.4} & \dn{20.8} \\
\midrule
\textbf{\method} \C{eCVPR27-M1-rsLoRA-r32-soup-D26CVPR} & 97.4 & 95.2 & 95.1 & 91.4 & 95.1 & 99.7 & \textbf{95.3} & 93.7 & 97.6 & 96.5 & \textbf{99.9} & \textbf{88.8} & \textbf{98.9} & \textbf{95.0} & \textbf{92.9} & \textbf{92.1} & 99.9 & \textbf{95.6} \\
\quad w/ KVAE 2D \C{eCVPR27-M2-dec-kvae2d-soup-D26CVPR} & \dn{1.3} & \dn{7.2} & \dn{1.0} & \dn{2.6} & \dn{4.2} & \dn{1.4} & \dn{7.4} & \dn{18.2} & \dn{2.9} & \dn{2.0} & \dn{0.1} & \dn{10.5} & \dn{4.2} & \dn{7.3} & \dn{19.9} & \dn{15.5} & \textbf{\boldmath\up{0.1}} & \dn{6.2} \\
\quad w/ LTX-Video \C{eCVPR27-M2-dec-ltxvideo-soup-D26CVPR} & \up{0.7} & \dn{6.5} & \up{3.8} & \up{3.2} & \up{1.0} & \dn{0.2} & \dn{8.0} & \dn{10.7} & \dn{0.4} & \up{1.5} & \dn{0.2} & \dn{10.5} & \dn{4.0} & \dn{12.4} & \dn{10.4} & \dn{10.7} & \textbf{\boldmath\up{0.1}} & \dn{3.7} \\
\quad w/ HV 1.0 \C{eCVPR27-M2-dec-hunyuanvideo-soup-D26CVPR} & \up{1.1} & \dn{6.3} & \up{1.2} & \up{1.1} & \up{0.5} & \dn{0.4} & \dn{1.4} & \dn{1.3} & \dn{1.0} & \up{1.3} & \dn{1.3} & \dn{18.9} & \dn{7.6} & \dn{15.4} & \dn{4.9} & \dn{4.7} & \textbf{\boldmath\up{0.1}} & \dn{3.4} \\
\quad w/ LTX-2 \C{eCVPR27-M2-dec-ltx2-soup-D26CVPR} & \textbf{\boldmath\up{1.8}} & \dn{8.7} & \textbf{\boldmath\up{4.1}} & \textbf{\boldmath\up{7.6}} & \textbf{\boldmath\up{3.2}} & \textbf{\boldmath\up{0.2}} & \dn{0.3} & \textbf{\boldmath\up{1.6}} & \textbf{\boldmath\up{0.9}} & \textbf{\boldmath\up{2.7}} & \textbf{\boldmath\ns{0.0}} & \dn{25.9} & \dn{8.7} & \dn{27.5} & \dn{1.3} & \dn{5.8} & \textbf{\boldmath\up{0.1}} & \dn{3.3} \\
\bottomrule
\end{tabular}
} %
\end{table}

\paragraph{\texttt{THB} and \texttt{DFAV}} \Cref{tab:supp-generators-thb-dfav} gives the same comparison for the generators of \texttt{THB} and \texttt{DFAV}. For \texttt{DFAV}, we use the fakes of its FVFA subset, where both the video and the audio are fake. \method reaches at least 96.7\% on every generator of \texttt{THB} and at least 93.1\% on every generator of \texttt{DFAV}. No prior detector is above \method on any generator of \texttt{THB}. On \texttt{DFAV}, prior detectors are above it in five cells, by at most 2.4 points: DDA and B-Free on Hallo, and B-Free, FS-VFM and DFD-HR on EchoMimic. On average, DDA, the best prior detector in this table, is 6.9 points below \method. The four other autoencoders are at most 3.5 points below CogVideoX on average, and LTX-Video and LTX-2 are on par with it.

\paragraph{\texttt{HF}} \Cref{tab:supp-generators-hf} gives the same comparison for \texttt{HF}, which has four test subsets: cross-domain, cross-forgery, cross-model and in-domain. \texttt{HF} also contains fakes of FF++ and \texttt{FFIW}. We removed them from \texttt{HF}, because they duplicate datasets that this paper already uses in full: FF++ for training (\cref{sec:data}) and \texttt{FFIW} as a test dataset (\cref{tab:test-datasets}). The real videos of some \texttt{HF} generators still include FF++ test reals, which \method never sees in training. \method reaches 94.0\% on average and at least 89.4\% on 18 of the 20 generators. The two exceptions are Adobe Firefly (63.9\%) and GPT-4o (67.9\%). As in \cref{tab:sota}, the AIGC detectors B-Free and DDA are strong on \texttt{HF}. On average, B-Free is 2.0 points above \method and DDA is 0.3 points above it. Their largest margins are on Adobe Firefly, 30.8 and 29.2 points. All facial deepfake detectors are below \method on average. FS-VFM, the best of them, is 18.4 points below. Facial deepfake detectors are above \method only on Dreamina, GPT-4o, Hailuo and Adobe Firefly. The four other autoencoders are 1.7 to 3.3 points below CogVideoX on average.

\begin{table}[h]
\centering
\caption{\label{tab:supp-generators-thb-dfav}
    \textbf{Results per generator on \texttt{THB} and \texttt{DFAV}.} Video-level AUROC (\%) on \texttt{THB} and \texttt{DFAV} for each of their generators. The row of \method is the AUROC. Every other row is the difference to \method in percentage points, green if higher and red if lower. The rows are those of \cref{tab:supp-generators}. The rows under \method make the reconstruction twins with another autoencoder instead of CogVideoX. HV is HunyuanVideo. AniPortrait is driven by audio (A) or by video (V). The \texttt{DFAV} columns use the fakes of its FVFA subset (fake video, fake audio). Mean is the average over the 9 columns. The best value in each column is in bold. \textsuperscript{\dag}A temporal model, which takes several frames as input.
}
\small\tabcolsep=3.4pt
\begin{tabular}{@{}l|cccccc|ccc|c@{}}
\toprule
 & \multicolumn{6}{c|}{\texttt{THB}~\cite{TalkingHeadBench}} & \multicolumn{3}{c|}{\texttt{DFAV}~\cite{DigiFakeAV}} & \\
 & \rotatebox{90}{AniPortrait A} & \rotatebox{90}{AniPortrait V} & \rotatebox{90}{EmoPortrait} & \rotatebox{90}{Hallo} & \rotatebox{90}{Hallo2} & \rotatebox{90}{LivePortrait} & \rotatebox{90}{Sonic} & \rotatebox{90}{Hallo} & \rotatebox{90}{EchoMimic} & \rotatebox{90}{Mean} \\
\midrule
\multicolumn{11}{@{}l}{\textit{AIGC detectors}} \\[2pt]
Effort-SDv1.4~\cite{Effort} \C{Effort-sdv14-D26CVPR} & \dn{19.5} & \dn{14.0} & \dn{16.2} & \dn{15.8} & \dn{23.4} & \dn{23.8} & \dn{43.9} & \dn{39.4} & \dn{54.5} & \dn{27.8} \\
DGS-Net~\cite{DGS-Net} \C{DGS-Net-D26CVPR} & \dn{18.5} & \dn{19.1} & \dn{83.5} & \dn{15.0} & \dn{9.8} & \dn{22.4} & \dn{23.7} & \dn{21.5} & \dn{34.3} & \dn{27.5} \\
LTD-UFD~\cite{LTD} \C{LTD-UFD-D26CVPR} & \dn{15.0} & \dn{9.0} & \dn{61.6} & \dn{6.1} & \dn{6.4} & \dn{15.7} & \dn{21.3} & \dn{13.8} & \dn{16.4} & \dn{18.4} \\
AlignedFor.~\cite{AlignedForensics} \C{AlignedForensics-Sync-D26CVPR} & \dn{1.9} & \dn{3.2} & \dn{3.0} & \dn{4.7} & \dn{2.6} & \dn{12.7} & \dn{7.3} & \dn{13.9} & \dn{13.5} & \dn{7.0} \\
DDA~\cite{DDA} \C{DDA-D26CVPR} & \dn{7.0} & \ns{0.0} & \dn{36.3} & \dn{2.3} & \dn{1.6} & \dn{15.8} & \dn{1.1} & \textbf{\boldmath\up{2.0}} & \dn{0.2} & \dn{6.9} \\
B-Free~\cite{B-Free} \C{B-Free-D26CVPR} & \dn{12.2} & \dn{1.1} & \dn{7.2} & \dn{3.1} & \dn{5.8} & \dn{35.2} & \dn{5.0} & \up{1.4} & \textbf{\boldmath\up{2.4}} & \dn{7.3} \\
\midrule
\multicolumn{11}{@{}l}{\textit{Facial deepfake detectors}} \\[2pt]
SBI~\cite{SBI} \C{SBI-FFraw-D26CVPR} & \dn{36.9} & \dn{30.5} & \dn{12.7} & \dn{40.9} & \dn{34.6} & \dn{20.2} & \dn{33.3} & \dn{46.7} & \dn{29.4} & \dn{31.7} \\
FSFM~\cite{FSFM} \C{FSFM-D26CVPR} & \dn{59.6} & \dn{50.0} & \dn{46.9} & \dn{53.0} & \dn{57.8} & \dn{44.1} & \dn{19.7} & \dn{31.1} & \dn{3.7} & \dn{40.7} \\
ForAda~\cite{ForAda} \C{ForAda-D26CVPR} & \dn{53.8} & \dn{38.0} & \dn{21.6} & \dn{36.6} & \dn{37.5} & \dn{45.8} & \dn{20.6} & \dn{30.2} & \dn{7.9} & \dn{32.4} \\
Effort~\cite{Effort} \C{Effort-D26CVPR} & \dn{42.9} & \dn{24.9} & \dn{1.9} & \dn{33.9} & \dn{29.2} & \dn{29.3} & \dn{23.6} & \dn{27.3} & \dn{5.7} & \dn{24.3} \\
DFD-FCG\textsuperscript{\dag}~\cite{DFD-FCG} \C{DFD-FCG-x1.4-D26CVPR} & \dn{29.2} & \dn{23.0} & \dn{8.9} & \dn{33.8} & \dn{39.5} & \dn{17.4} & \dn{34.1} & \dn{35.9} & \dn{1.6} & \dn{24.8} \\
VLAForge~\cite{VLAForge} \C{VLAForge-D26CVPR} & \dn{40.3} & \dn{19.3} & \dn{2.4} & \dn{24.7} & \dn{19.2} & \dn{26.6} & \dn{21.8} & \dn{25.8} & \dn{5.3} & \dn{20.6} \\
FS-VFM~\cite{FS-VFM} \C{FS-VFM-D26CVPR} & \dn{45.0} & \dn{32.1} & \dn{41.2} & \dn{33.5} & \dn{36.1} & \dn{30.1} & \dn{11.1} & \dn{24.4} & \up{1.7} & \dn{28.0} \\
DFD-HR~\cite{DFD-HR} \C{DFD-HR-D26CVPR} & \dn{36.3} & \dn{18.6} & \dn{4.3} & \dn{20.9} & \dn{18.6} & \dn{12.0} & \dn{6.9} & \dn{14.5} & \up{0.4} & \dn{14.6} \\
GenD-PE~\cite{GenD} \C{GenD-PE-HF-D26CVPR} & \dn{30.8} & \dn{13.3} & \dn{9.8} & \dn{14.5} & \dn{16.7} & \dn{7.5} & \dn{6.4} & \dn{19.0} & \dn{1.4} & \dn{13.3} \\
\midrule
\textbf{\method} \C{eCVPR27-M1-rsLoRA-r32-soup-D26CVPR} & 98.5 & 98.3 & 99.8 & \textbf{99.7} & 96.7 & 97.7 & 93.8 & 93.1 & 94.0 & 96.8 \\
\quad w/ KVAE 2D \C{eCVPR27-M2-dec-kvae2d-soup-D26CVPR} & \dn{2.0} & \dn{1.3} & \dn{0.2} & \dn{0.6} & \textbf{\boldmath\up{0.8}} & \dn{6.1} & \up{0.6} & \dn{0.9} & \up{1.0} & \dn{1.0} \\
\quad w/ LTX-Video \C{eCVPR27-M2-dec-ltxvideo-soup-D26CVPR} & \ns{0.0} & \up{0.5} & \textbf{\boldmath\up{0.1}} & \textbf{\boldmath\ns{0.0}} & \dn{0.9} & \textbf{\boldmath\up{0.2}} & \up{1.8} & \dn{1.2} & \up{0.3} & \textbf{\boldmath\up{0.1}} \\
\quad w/ HV 1.0 \C{eCVPR27-M2-dec-hunyuanvideo-soup-D26CVPR} & \dn{2.9} & \dn{1.7} & \dn{1.2} & \dn{0.9} & \dn{3.8} & \dn{12.6} & \dn{1.6} & \dn{5.6} & \dn{1.1} & \dn{3.5} \\
\quad w/ LTX-2 \C{eCVPR27-M2-dec-ltx2-soup-D26CVPR} & \textbf{\boldmath\up{0.1}} & \textbf{\boldmath\up{0.9}} & \ns{0.0} & \textbf{\boldmath\ns{0.0}} & \dn{1.9} & \dn{2.9} & \textbf{\boldmath\up{3.4}} & \up{0.8} & \dn{0.4} & \ns{0.0} \\
\bottomrule
\end{tabular}
\end{table}

\begin{table}[h]
\centering
\caption{\label{tab:supp-generators-hf}
    \textbf{Results per generator on \texttt{HF}.} Video-level AUROC (\%) on \texttt{HF} for each generator of its four test subsets: cross-domain (CD), cross-forgery (CF), cross-model (CM) and in-domain (ID). The row of \method is the AUROC. Every other row is the difference to \method in percentage points, green if higher and red if lower. The rows are those of \cref{tab:supp-generators}. The rows under \method make the reconstruction twins with another autoencoder instead of CogVideoX. HV is HunyuanVideo. CD mean, CF mean, CM mean and ID mean are the averages over the generator columns of each subset. Mean is the average over all 20 generator columns. The best value in each column is in bold. \textsuperscript{\dag}A temporal model, which takes several frames as input.
}
\tabcolsep=2.5pt
\resizebox{\textwidth}{!}{
\begin{tabular}{@{}l|cccccc|ccccccc|ccccccc|cccc|c@{}}
\toprule
 & \multicolumn{6}{c|}{Cross-domain (CD)} & \multicolumn{7}{c|}{Cross-forgery (CF)} & \multicolumn{7}{c|}{Cross-model (CM)} & \multicolumn{4}{c|}{In-domain (ID)} & \\
 & \rotatebox{90}{DeepFaceLab} & \rotatebox{90}{Dreamina} & \rotatebox{90}{GPT-4o} & \rotatebox{90}{Hailuo} & \rotatebox{90}{InfiniteYou} & \rotatebox{90}{\textit{CD mean}} & \rotatebox{90}{CodeFormer} & \rotatebox{90}{FaceAdapter} & \rotatebox{90}{IC-Light} & \rotatebox{90}{InfiniteYou} & \rotatebox{90}{PuLID} & \rotatebox{90}{StarGANv2} & \rotatebox{90}{\textit{CF mean}} & \rotatebox{90}{Adobe Firefly} & \rotatebox{90}{FLUX1.1 Pro} & \rotatebox{90}{HART} & \rotatebox{90}{Infinity} & \rotatebox{90}{MAGI} & \rotatebox{90}{StarryAI} & \rotatebox{90}{\textit{CM mean}} & \rotatebox{90}{Hallo2} & \rotatebox{90}{Midjourney} & \rotatebox{90}{StyleGAN} & \rotatebox{90}{\textit{ID mean}} & \rotatebox{90}{Mean} \\
\midrule
\multicolumn{26}{@{}l}{\textit{AIGC detectors}} \\[2pt]
Effort-SDv1.4~\cite{Effort} \C{Effort-sdv14-D26CVPR} & \dn{51.6} & \dn{32.8} & \dn{18.0} & \dn{44.1} & \dn{21.7} & \dn{33.6} & \dn{29.5} & \dn{22.6} & \dn{23.2} & \dn{21.4} & \dn{25.4} & \dn{58.4} & \dn{30.1} & \up{10.4} & \dn{7.7} & \dn{8.1} & \dn{4.2} & \dn{49.8} & \dn{3.3} & \dn{10.5} & \dn{56.4} & \dn{10.6} & \dn{34.5} & \dn{33.8} & \dn{25.6} \\
DGS-Net~\cite{DGS-Net} \C{DGS-Net-D26CVPR} & \dn{68.3} & \up{0.4} & \dn{19.5} & \dn{9.4} & \dn{5.2} & \dn{20.4} & \dn{18.5} & \dn{50.5} & \dn{50.0} & \dn{41.2} & \dn{19.3} & \dn{88.0} & \dn{44.6} & \dn{51.8} & \dn{16.3} & \dn{25.2} & \dn{20.8} & \dn{36.1} & \dn{19.1} & \dn{28.2} & \dn{36.1} & \dn{26.7} & \dn{18.4} & \dn{27.1} & \dn{31.0} \\
LTD-UFD~\cite{LTD} \C{LTD-UFD-D26CVPR} & \dn{62.5} & \dn{38.9} & \dn{41.3} & \dn{26.5} & \dn{1.4} & \dn{34.1} & \dn{57.3} & \dn{35.3} & \dn{59.0} & \dn{58.7} & \dn{61.1} & \dn{88.8} & \dn{60.0} & \dn{57.8} & \dn{55.6} & \dn{22.1} & \dn{34.6} & \dn{41.6} & \dn{88.5} & \dn{50.0} & \dn{69.8} & \dn{45.6} & \dn{8.2} & \dn{41.2} & \dn{47.7} \\
AlignedFor.~\cite{AlignedForensics} \C{AlignedForensics-Sync-D26CVPR} & \dn{40.2} & \dn{2.7} & \dn{34.3} & \dn{44.2} & \dn{0.4} & \dn{24.4} & \up{1.8} & \up{0.2} & \textbf{\boldmath\up{2.3}} & \up{0.7} & \dn{0.9} & \dn{24.6} & \dn{3.4} & \textbf{\boldmath\up{32.4}} & \up{1.2} & \dn{0.6} & \up{1.2} & \dn{11.8} & \textbf{\boldmath\up{5.9}} & \up{4.7} & \dn{12.7} & \dn{11.4} & \dn{68.1} & \dn{30.7} & \dn{10.3} \\
DDA~\cite{DDA} \C{DDA-D26CVPR} & \dn{32.3} & \textbf{\boldmath\up{7.0}} & \up{17.7} & \textbf{\boldmath\up{8.7}} & \dn{0.7} & \up{0.1} & \dn{0.1} & \textbf{\boldmath\up{0.6}} & \textbf{\boldmath\up{2.3}} & \up{1.0} & \up{1.2} & \up{1.2} & \up{1.0} & \up{29.2} & \dn{0.1} & \dn{0.7} & \up{1.0} & \dn{2.1} & \textbf{\boldmath\up{5.9}} & \up{5.5} & \textbf{\boldmath\up{1.9}} & \dn{8.4} & \dn{27.1} & \dn{11.2} & \up{0.3} \\
B-Free~\cite{B-Free} \C{B-Free-D26CVPR} & \dn{21.4} & \up{5.1} & \up{8.6} & \up{5.2} & \dn{2.3} & \dn{1.0} & \textbf{\boldmath\up{2.4}} & \textbf{\boldmath\up{0.6}} & \up{2.2} & \textbf{\boldmath\up{1.2}} & \textbf{\boldmath\up{1.3}} & \textbf{\boldmath\up{1.3}} & \textbf{\boldmath\up{1.5}} & \up{30.8} & \textbf{\boldmath\up{3.2}} & \textbf{\boldmath\ns{0.0}} & \textbf{\boldmath\up{1.3}} & \dn{2.8} & \textbf{\boldmath\up{5.9}} & \textbf{\boldmath\up{6.4}} & \up{0.3} & \dn{4.8} & \textbf{\boldmath\up{1.6}} & \dn{1.0} & \textbf{\boldmath\up{2.0}} \\
\midrule
\multicolumn{26}{@{}l}{\textit{Facial deepfake detectors}} \\[2pt]
SBI~\cite{SBI} \C{SBI-FFraw-D26CVPR} & \dn{20.6} & \dn{11.5} & \dn{18.4} & \dn{17.2} & \dn{44.2} & \dn{22.4} & \dn{65.5} & \dn{35.7} & \dn{14.8} & \dn{34.2} & \dn{30.3} & \dn{35.4} & \dn{36.0} & \dn{15.8} & \dn{34.8} & \dn{32.1} & \dn{39.4} & \dn{44.8} & \dn{53.5} & \dn{36.7} & \dn{39.2} & \dn{42.4} & \dn{59.1} & \dn{46.9} & \dn{34.4} \\
FSFM~\cite{FSFM} \C{FSFM-D26CVPR} & \dn{10.2} & \up{2.5} & \up{12.2} & \up{1.5} & \dn{29.2} & \dn{4.6} & \dn{29.8} & \dn{28.8} & \dn{12.7} & \dn{37.8} & \dn{20.0} & \dn{38.8} & \dn{28.0} & \dn{8.1} & \dn{30.2} & \dn{17.5} & \dn{24.8} & \dn{45.3} & \dn{33.3} & \dn{26.5} & \dn{44.5} & \dn{29.0} & \dn{68.3} & \dn{47.3} & \dn{24.6} \\
ForAda~\cite{ForAda} \C{ForAda-D26CVPR} & \dn{3.9} & \dn{34.5} & \dn{23.3} & \dn{24.0} & \dn{34.2} & \dn{24.0} & \dn{23.4} & \dn{17.4} & \dn{14.5} & \dn{46.1} & \dn{35.2} & \dn{23.9} & \dn{26.8} & \dn{19.1} & \dn{66.3} & \dn{42.4} & \dn{49.7} & \dn{45.4} & \dn{52.2} & \dn{45.9} & \dn{29.4} & \dn{62.1} & \dn{46.8} & \dn{46.1} & \dn{34.7} \\
Effort~\cite{Effort} \C{Effort-D26CVPR} & \dn{4.3} & \dn{36.6} & \dn{26.9} & \dn{22.7} & \dn{31.8} & \dn{24.5} & \dn{6.9} & \dn{4.9} & \dn{18.7} & \dn{27.3} & \dn{40.3} & \dn{4.8} & \dn{17.2} & \dn{18.2} & \dn{64.2} & \dn{42.4} & \dn{45.2} & \dn{25.2} & \dn{62.6} & \dn{43.0} & \dn{13.0} & \dn{64.2} & \dn{6.8} & \dn{28.0} & \dn{28.4} \\
DFD-FCG\textsuperscript{\dag}~\cite{DFD-FCG} \C{DFD-FCG-x1.4-D26CVPR} & \dn{2.9} & \dn{40.4} & \dn{8.8} & \dn{38.4} & \dn{50.0} & \dn{28.1} & \dn{21.0} & \dn{14.6} & \dn{12.2} & \dn{51.3} & \dn{40.7} & \dn{18.5} & \dn{26.4} & \dn{5.5} & \dn{35.4} & \dn{19.6} & \dn{32.7} & \dn{54.2} & \dn{40.5} & \dn{31.3} & \dn{50.9} & \dn{48.8} & \dn{38.8} & \dn{46.2} & \dn{31.3} \\
VLAForge~\cite{VLAForge} \C{VLAForge-D26CVPR} & \dn{6.3} & \dn{26.4} & \dn{24.5} & \dn{23.2} & \dn{23.2} & \dn{20.7} & \dn{14.4} & \dn{8.1} & \dn{23.5} & \dn{34.2} & \dn{39.7} & \dn{27.6} & \dn{24.6} & \dn{15.5} & \dn{70.0} & \dn{42.5} & \dn{56.7} & \dn{47.6} & \dn{57.5} & \dn{48.3} & \dn{29.2} & \dn{68.7} & \dn{16.2} & \dn{38.0} & \dn{32.8} \\
FS-VFM~\cite{FS-VFM} \C{FS-VFM-D26CVPR} & \dn{4.5} & \up{6.0} & \textbf{\boldmath\up{18.8}} & \up{3.2} & \dn{18.3} & \up{1.0} & \dn{21.8} & \dn{12.2} & \dn{5.2} & \dn{30.4} & \dn{12.4} & \dn{27.3} & \dn{18.2} & \up{3.0} & \dn{35.0} & \dn{11.6} & \dn{22.5} & \dn{42.0} & \dn{37.2} & \dn{24.2} & \dn{29.8} & \dn{34.6} & \dn{55.0} & \dn{39.8} & \dn{18.4} \\
DFD-HR~\cite{DFD-HR} \C{DFD-HR-D26CVPR} & \dn{3.2} & \dn{7.8} & \dn{1.4} & \dn{6.3} & \dn{14.3} & \dn{6.6} & \dn{12.4} & \dn{5.2} & \dn{12.2} & \dn{33.5} & \dn{47.1} & \dn{13.9} & \dn{20.7} & \dn{40.2} & \dn{80.7} & \dn{43.3} & \dn{60.4} & \dn{20.9} & \dn{76.7} & \dn{53.7} & \dn{9.9} & \dn{62.7} & \dn{21.7} & \dn{31.4} & \dn{28.7} \\
GenD-PE~\cite{GenD} \C{GenD-PE-HF-D26CVPR} & \dn{3.8} & \up{6.3} & \up{10.4} & \up{0.2} & \dn{1.8} & \textbf{\boldmath\up{2.3}} & \dn{21.3} & \dn{6.6} & \dn{15.3} & \dn{14.8} & \dn{24.4} & \dn{29.8} & \dn{18.7} & \dn{34.0} & \dn{29.6} & \dn{21.2} & \dn{22.3} & \dn{22.9} & \dn{56.9} & \dn{31.2} & \dn{29.3} & \dn{34.5} & \dn{42.9} & \dn{35.6} & \dn{19.7} \\
\midrule
\textbf{\method} \C{eCVPR27-M1-rsLoRA-r32-soup-D26CVPR} & \textbf{99.6} & 89.4 & 67.9 & 90.3 & 97.6 & 89.0 & 97.6 & 99.4 & 97.7 & 98.2 & 98.2 & 98.6 & 98.3 & 63.9 & 96.2 & \textbf{99.9} & 98.7 & 99.6 & 94.1 & 92.1 & 97.9 & 98.8 & 97.0 & \textbf{97.9} & 94.0 \\
\quad w/ KVAE 2D \C{eCVPR27-M2-dec-kvae2d-soup-D26CVPR} & \dn{0.1} & \dn{2.4} & \dn{20.5} & \dn{12.0} & \up{1.0} & \dn{6.8} & \dn{2.3} & \dn{0.2} & \up{0.9} & \up{0.8} & \up{0.4} & \dn{0.2} & \dn{0.1} & \up{5.1} & \up{0.2} & \dn{0.2} & \up{0.2} & \dn{0.2} & \dn{7.3} & \dn{0.4} & \dn{0.4} & \textbf{\boldmath\up{0.6}} & \dn{3.7} & \dn{1.2} & \dn{2.0} \\
\quad w/ LTX-Video \C{eCVPR27-M2-dec-ltxvideo-soup-D26CVPR} & \dn{1.0} & \dn{6.9} & \dn{11.3} & \dn{17.6} & \dn{1.4} & \dn{7.6} & \ns{0.0} & \up{0.2} & \up{0.5} & \up{0.8} & \up{0.6} & \up{0.4} & \up{0.4} & \up{0.3} & \up{1.5} & \textbf{\boldmath\ns{0.0}} & \up{0.3} & \ns{0.0} & \dn{0.4} & \up{0.3} & \ns{0.0} & \dn{0.1} & \dn{0.2} & \dn{0.1} & \dn{1.7} \\
\quad w/ HV 1.0 \C{eCVPR27-M2-dec-hunyuanvideo-soup-D26CVPR} & \dn{0.1} & \dn{4.4} & \dn{23.5} & \dn{10.5} & \textbf{\boldmath\up{1.2}} & \dn{7.5} & \dn{3.7} & \dn{0.4} & \dn{0.6} & \up{0.8} & \up{0.1} & \dn{0.6} & \dn{0.7} & \dn{2.7} & \up{0.7} & \dn{0.1} & \up{0.1} & \dn{0.2} & \dn{16.1} & \dn{3.1} & \dn{1.8} & \up{0.3} & \dn{4.3} & \dn{1.9} & \dn{3.3} \\
\quad w/ LTX-2 \C{eCVPR27-M2-dec-ltx2-soup-D26CVPR} & \dn{1.2} & \dn{8.3} & \dn{20.5} & \dn{15.3} & \dn{0.3} & \dn{9.1} & \dn{0.7} & \up{0.3} & \up{1.0} & \textbf{\boldmath\up{1.2}} & \up{0.5} & \ns{0.0} & \up{0.4} & \dn{0.5} & \up{1.7} & \textbf{\boldmath\ns{0.0}} & \up{0.8} & \textbf{\boldmath\up{0.1}} & \dn{2.9} & \dn{0.1} & \dn{0.3} & \up{0.5} & \dn{0.4} & \dn{0.1} & \dn{2.2} \\
\bottomrule
\end{tabular}
} %
\end{table}

\subsection{What decodes the pixels in the generators of \texttt{OF} and \texttt{MV}}
\label{sec:supp-generator-decoders}

Reconstruction twins teach \method the traces of the decoder of an autoencoder (\cref{sec:rec-twins}). \Cref{tab:supp-generator-decoders} lists what produces the pixels in each generator of \texttt{OF} and \texttt{MV}, as stated in the paper or the released code of the generator. None of them uses the CogVideoX autoencoder or one of the four other autoencoders of \cref{tab:supp-generators}.

\begin{table}[h]
  \centering
  \small
  \caption{\label{tab:supp-generator-decoders}
    \textbf{What produces the pixels in the generators of \texttt{OF} and \texttt{MV}.} As stated in the paper or the released code of each generator.
  }
  \begin{tabular}{@{}lll@{}}
    \toprule
    Dataset & Generator & Pixels are produced by \\
    \midrule
    \texttt{OF} & ACTalker~\scite{ACTalker} & the autoencoder of Stable Video Diffusion~\scite{SVD} \\
     & Ditto~\scite{Ditto} & a face renderer from LivePortrait~\scite{LivePortrait}, which warps features of the source image \\
     & EDTalk~\scite{EDTalk} & the generator of LIA~\scite{LIA}, trained with an adversarial loss \\
     & FLOAT~\scite{FLOAT} & a motion autoencoder based on LIA~\scite{LIA}, trained with adversarial losses \\
    \midrule
    \texttt{MV} & FantasyTalking~\scite{FantasyTalking} & the Wan 2.1 autoencoder~\scite{WAN-2.1} \\
     & OmniAvatar~\scite{OmniAvatar} & the Wan 2.1 autoencoder~\scite{WAN-2.1} \\
     & Phantom-14B~\scite{Phantom} & the Wan 2.1 autoencoder~\scite{WAN-2.1} \\
    \bottomrule
  \end{tabular}
\end{table}

\paragraph{Generators without a latent diffusion autoencoder} Ditto, EDTalk and FLOAT do not decode the latents of a latent diffusion model. Their diffusion or flow model generates motion, and a separate network turns the source image and the motion into frames. \method still reaches 88.8\% on Ditto, 98.9\% on EDTalk and 95.0\% on FLOAT. On Ditto and FLOAT, all four other autoencoders are at least 7.3 points below CogVideoX.

\clearpage
\section{Gaussian noise and downscaling per dataset}
\label{sec:supp-degradation}

\Cref{tab:supp-degradation} gives the AUROC of \method on each dataset under the Gaussian noise and the downscaling of \cref{sec:robustness}. Both degradations cost the most on the \texttt{EFS} datasets. At the weakest noise, \texttt{DF26} drops from 95.8 to 55.8 and \texttt{OF} from 94.9 to 60.0, while eight of the ten \texttt{Blending} datasets lose less than 5 points. Downscaling to 112\,px costs the most on \texttt{MV} (95.0 to 63.5) and \texttt{HF} (96.1 to 70.2). The AUROC does not always fall as the degradation gets stronger. \texttt{HF}, \texttt{DFAV} and \texttt{MV} have a higher AUROC at 64\,px than at 112\,px.

\begin{table}[h]
\centering
\small
\caption{\label{tab:supp-degradation}
    \textbf{Gaussian noise and downscaling per dataset.} Video-level AUROC (\%) of \method under the Gaussian noise and the downscaling of \cref{sec:robustness}. Clean is the row of \method in \cref{tab:sota}.
}
\begin{tabular}{@{}l|c|ccccc|cc@{}}
\toprule
 & & \multicolumn{5}{c|}{Gaussian noise variance} & \multicolumn{2}{c}{Downscale to (px)} \\
 & Clean & 0.001 & 0.002 & 0.005 & 0.01 & 0.05 & 112 & 64 \\
\midrule
\multicolumn{9}{@{}l}{\textbf{\texttt{Blending}}} \\[2pt]
\texttt{DFD} & 98.9 & 95.7 & 93.4 & 88.7 & 84.1 & 61.7 & 97.7 & 91.5 \\
\texttt{FSh} & 96.8 & 90.0 & 88.2 & 85.0 & 82.0 & 67.2 & 86.2 & 77.1 \\
\texttt{CDFv2} & 97.1 & 84.2 & 80.7 & 75.0 & 70.5 & 64.5 & 90.0 & 85.4 \\
\texttt{FFIW} & 98.9 & 97.6 & 95.7 & 90.5 & 84.5 & 66.9 & 92.6 & 80.8 \\
\texttt{FAVC} & 99.9 & 95.4 & 92.5 & 87.6 & 82.8 & 65.8 & 97.2 & 88.2 \\
\texttt{DFDM} & 99.9 & 98.4 & 96.7 & 91.5 & 84.4 & 61.1 & 99.5 & 97.0 \\
\texttt{PGF} & 97.3 & 96.0 & 95.0 & 91.7 & 88.0 & 74.3 & 87.7 & 74.4 \\
\texttt{IDF} & 99.5 & 98.2 & 96.6 & 93.7 & 89.1 & 56.3 & 97.9 & 85.9 \\
\texttt{DSv1} & 93.5 & 91.2 & 88.7 & 84.4 & 81.3 & 68.6 & 82.8 & 81.9 \\
\texttt{HiDF} & 98.7 & 94.9 & 92.4 & 87.4 & 81.8 & 64.0 & 87.2 & 82.8 \\[3pt]
\textit{Mean} & 98.1 & 94.2 & 92.0 & 87.6 & 82.9 & 65.0 & 91.9 & 84.5 \\
\midrule
\multicolumn{9}{@{}l}{\textbf{\texttt{B+EFS}}} \\[2pt]
\texttt{CDFv3} & 90.4 & 84.1 & 82.9 & 80.5 & 77.1 & 67.5 & 76.3 & 73.9 \\
\texttt{DSv2} & 92.1 & 76.6 & 70.8 & 65.1 & 61.3 & 51.7 & 84.6 & 79.2 \\
\texttt{RF} & 92.3 & 72.6 & 64.8 & 55.0 & 52.9 & 55.7 & 88.0 & 84.0 \\
\texttt{HF} & 96.1 & 83.4 & 78.1 & 74.9 & 75.0 & 67.4 & 70.2 & 75.9 \\[3pt]
\textit{Mean} & 92.7 & 79.2 & 74.2 & 68.9 & 66.6 & 60.6 & 79.8 & 78.3 \\
\midrule
\multicolumn{9}{@{}l}{\textbf{\texttt{EFS}}} \\[2pt]
\texttt{THB} & 98.5 & 82.0 & 76.3 & 67.1 & 63.7 & 53.2 & 89.3 & 65.8 \\
\texttt{OF} & 94.9 & 60.0 & 53.9 & 48.9 & 49.9 & 53.7 & 78.7 & 57.2 \\
\texttt{DFAV} & 93.7 & 75.3 & 71.9 & 65.8 & 61.8 & 56.2 & 77.3 & 79.1 \\
\texttt{MV} & 95.0 & 74.5 & 73.8 & 77.3 & 79.9 & 73.0 & 63.5 & 65.7 \\
\texttt{DF26} & 95.8 & 55.8 & 47.9 & 42.2 & 41.3 & 43.3 & 86.5 & 73.1 \\[3pt]
\textit{Mean} & 95.6 & 69.5 & 64.8 & 60.3 & 59.3 & 55.9 & 79.1 & 68.2 \\
\midrule
All & 96.3 & 84.5 & 81.1 & 76.4 & 73.2 & 61.7 & 86.0 & 78.9 \\
\bottomrule
\end{tabular}
\end{table}

{
    \small
    \renewcommand{\bibnumfmt}[1]{[S#1]}
    \bibliographystylesupp{ieeenat_fullname}
    \bibliographysupp{main}
}

\end{document}